%% file: main.tex
\documentclass{article}
\ifdefined\pdfinfoomitdate\pdfinfoomitdate=1\fi
\ifdefined\pdftrailerid\pdftrailerid{}\fi

\makeatletter
\@namedef{ver@algorithmic.sty}{9999/99/99}
\makeatother

\usepackage{xcolor} 
\usepackage{microtype}
\usepackage{graphicx}
\usepackage{subfigure}
\usepackage{booktabs} 
\usepackage{adjustbox} 
\usepackage{amsmath,amssymb,amsthm}
\usepackage{mathtools}
\usepackage{enumitem}
\usepackage{algorithm}
\usepackage{algpseudocode}

\usepackage{hyperref}

\usepackage{trl_preprint,times}
\iclrfinalcopy
\hypersetup{
  hidelinks,
  pdftitle={Tubular Low-Rank Laplace Approximations for Bayesian Neural Networks},
  pdfsubject={Preprint},
  pdfauthor={Rodrigo Pereira David},
  pdfkeywords={}
}

\usepackage{natbib}

\usepackage{float}

\newcommand{\meanstd}[2]{#1\,$\pm$\,#2}

\newcommand{\N}{\mathcal{N}}

\newcommand{\KL}{\mathrm{KL}}

\title{Tubular Low-Rank Laplace Approximations\\
for Bayesian Neural Networks}

\author{Rodrigo Pereira David\\
{\normalfont National Institute of Metrology, Technology and Quality (Inmetro)}}

\begin{document}

\maketitle

\begin{abstract}
Laplace approximations provide post-hoc uncertainty estimates, but a single local Gaussian around a maximum-a-posteriori (MAP) solution can poorly represent curved, anisotropic low-loss regions. We introduce Tubular Low-Rank Laplace (TRL), a single-checkpoint post-hoc posterior approximation that combines a discrete predictor--corrector spine with transverse Gaussian uncertainty across the full network. Starting from one MAP checkpoint, TRL uses stochastic cross-entropy Hessian-vector products to select a low-rank positive-curvature subspace as a Fisher/GGN-motivated proxy, transports the basis along a predictor--corrector spine, and samples validation-scaled transverse perturbations around uniformly selected anchors. Predictions use the resulting nonlinear networks, avoiding full curvature matrices and repeated local decompositions. On CIFAR-100, TRL improves NLL and calibration over the evaluated last-layer and rank-matched all-weight Laplace baselines and SWAG-Diag. Across five independently trained checkpoints, mean top-label ECE is $0.013$, versus $0.096$ for MAP and $0.028$ after temperature scaling. On ImageNet/ResNet-50, the MAP-centered TRL tube reduces ECE relative to the deterministic checkpoint in all ten evaluation splits, with small NLL and accuracy costs; temperature scaling achieves lower NLL and ECE. Ablations show that transverse uncertainty accounts for most gains around deeply trained CIFAR-100 checkpoints. In small-data fine-tuning, the spine contains functionally distinct anchors; uniform mixing increases predictive diversity while retaining mean NLL and Brier comparable to a validation-selected single anchor, without an additional anchor-selection search.
\end{abstract}

\section{Introduction}
\label{sec:intro}

Laplace approximations are attractive post-hoc uncertainty methods for neural networks. Starting from a trained maximum a posteriori (MAP) solution, they fit a local Gaussian using curvature of the negative log posterior and can therefore be attached to an existing checkpoint without changing the original training procedure. Scalable variants use diagonal, Kronecker-factored, last-layer, or low-rank curvature approximations \citep{martens2015optimizing,ritter2018scalable,daxberger2021laplace}. Their main limitation is locality: a single Euclidean Gaussian around one MAP point can be poorly aligned with the posterior geometry of an overparameterized network.

That locality is especially restrictive when low-loss sets are connected and strongly anisotropic. A local ellipsoid can represent uncertainty around one point, but it has no explicit way to allocate support along a curved low-loss direction. Linearized Laplace improves how local weight uncertainty is propagated to predictions \citep{immer2021linearized,antoran2022adapting,daxberger2021laplace,roy2024reparameterization}, yet it remains tied to a local approximation around the MAP. Geometric and Riemannian posterior methods address this structure more directly through Fisher/GGN metrics, geodesic transport, or learned low-loss manifolds \citep{amari1998natural,bonnabel2013sgdriem,kristiadi2023geometry,leger2024geometric,bergamin2023riemannian,yu2024fisherlaplace,dold2024embedding,dacosta2025geometric}. Their more faithful forms, however, require operations that are difficult to scale, such as repeated local curvature fitting, geodesic construction, or MCMC over a connected manifold. This motivates a simpler question: can we preserve the useful longitudinal/transverse decomposition while keeping the inference procedure post hoc and low rank?

We introduce \emph{Tubular Low-Rank Laplace} (TRL), a posterior approximation built from two support components around one MAP checkpoint. Stochastic cross-entropy Hessian-vector products identify a low-rank, full-network transverse subspace through leading positive-curvature directions used as a Fisher/GGN-motivated proxy. A discrete predictor--corrector spine provides longitudinal support through nearby stored anchors. TRL samples validation-scaled Gaussian perturbations in the transverse subspace around stored spine anchors and evaluates the resulting nonlinear networks. The construction avoids full curvature matrices, exact geodesics, and a fresh local decomposition at every anchor.

The experiments separate the roles of these two components. On CIFAR-100, TRL improves NLL and calibration over the evaluated last-layer and rank-matched all-weight Laplace baselines and SWAG-Diag. A matched-prior orientation control favors curvature-selected support in ECE, but random support has numerically lower NLL (Appendix~\ref{app:orientation_checkpoint}). Around deeply trained CIFAR checkpoints, a MAP-centered transverse tube captures most of the gain. In small-data fine-tuning, the spine carries stronger functional variation: uniform mixing increases predictive diversity while retaining mean NLL and Brier comparable to a validation-selected single anchor without a separate anchor-selection search. Across ten ImageNet/ResNet-50 splits, the MAP-centered TRL tube reduces ECE in every split, while temperature scaling attains lower NLL and ECE; Deep Ensembles retain the best reported mean clean CIFAR-100 accuracy, NLL, and Brier.

This paper makes three contributions:
\begin{itemize}[leftmargin=*]
\item We introduce TRL, a single-checkpoint post-hoc posterior approximation that combines a curvature-selected, full-network low-rank transverse sampler with a discrete predictor--corrector spine and nonlinear posterior prediction.
\item We examine the contributions of transverse sampling, longitudinal mixing, and amortized basis transport using single-checkpoint/full-spine, random-support, fixed/transported/fresh-basis, and fine-tuning diagnostics. The results show different relative contributions across the evaluated regimes: transverse support accounts for most gains around deeply trained CIFAR checkpoints, whereas fine-tuning exhibits stronger functional variation along the spine.
\item We evaluate TRL on toy problems (Appendix~\ref{app:toy}), CIFAR-100, CIFAR-100-C, CIFAR-10H human-label alignment, architecture checks, and an ImageNet scale-check (Appendix~\ref{app:imagenet_scale_check}), including rank-matched all-weight low-rank Laplace, temperature scaling, random-subspace controls, and ablations of rank, spine extent, tube scale, curvature refresh, prior structure, and BatchNorm consistency.
\end{itemize}

Throughout, we distinguish this geometric idealization from the discrete, transported low-rank algorithm used in experiments. The ancillary archive accompanying this preprint provides executable code and compact provenance artifacts used to audit and reproduce the reported experiments.

\section{Background}
\label{sec:background}

We consider a Bayesian neural network with prior $p(\theta)$ and likelihood $p(\mathcal D\mid\theta)$. Let $\mathcal L=-\log p(\mathcal D\mid\theta)-\log p(\theta)$ and $H=\nabla_\theta^2\mathcal L(\theta_{\mathrm{MAP}})$. ELA fits $\N(\theta_{\mathrm{MAP}},H^{-1})$, whereas LLA uses $f_{\theta_{\mathrm{MAP}}+\delta}(x)\approx f_{\theta_{\mathrm{MAP}}}(x)+J_{\mathrm{MAP}}(x)\delta$, where $\delta$ is a parameter perturbation and $J_{\mathrm{MAP}}(x)$ is the output Jacobian at the MAP. With $g_\theta=\nabla_\theta\log p_\theta(y\mid x)$, the Fisher is $F(\theta)=\mathbb E[g_\theta g_\theta^\top]$. Fisher/GGN are PSD sensitivity operators; TRL instead retains the positive spectrum of stochastic cross-entropy Hessian HVPs, so they motivate rather than equal the computed operator. Appendix~\ref{app:background} gives details.
\section{Geometric motivation}
\label{sec:geometry}

Deep-network loss landscapes contain highly anisotropic low-loss regions: many directions are near-invariances of the realized function, while others correspond to genuine mode-connectivity directions that can change predictions at comparable loss. TRL uses this distinction operationally. Because posterior density depends on both the likelihood and the prior, low training loss alone does not determine posterior mass. TRL uses the constructed path as a support-allocation heuristic, while the amount of transverse spread within the selected support is controlled separately. With a fixed posterior-rank budget, locally flat symmetry directions can absorb substantial parameter variance while inducing little predictive diversity. TRL therefore allocates transverse rank to data-sensitive directions on the training distribution and represents longitudinal motion separately through the spine. Low loss is a construction goal, not an enforced acceptance constraint; Appendix~\ref{app:toy_corrector_audit} audits the released sine paths.

\begin{figure}[t]
\centering
\includegraphics[width=0.48\linewidth]{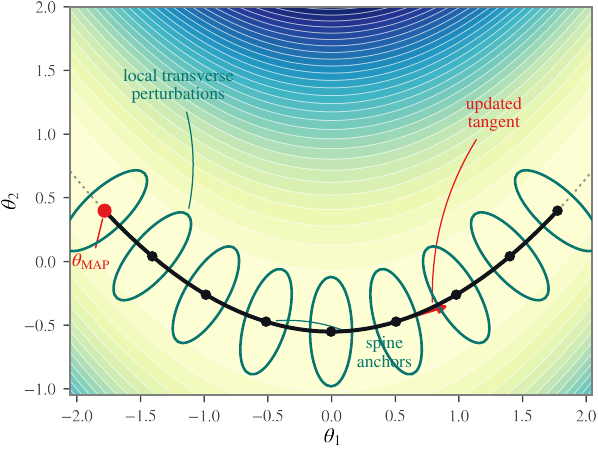}
\caption{\textbf{Practical TRL support.} Starting from the MAP, TRL stores a discrete predictor--corrector spine and samples low-rank transverse Gaussian perturbations around its anchors. The tangent is updated from the realized spine displacement, while the transverse frame is transported and re-orthogonalized without a fresh local curvature decomposition at every anchor. Path curvature and fiber sizes are exaggerated for legibility.}
\label{fig:support}
\end{figure}

Figure~\ref{fig:support} illustrates how TRL selects a low-rank set of data-sensitive transverse directions with positive-spectrum cross-entropy Hessian HVPs, used as a Fisher/GGN-motivated curvature proxy, constructs a discrete predictor--corrector spine in the complementary region, and samples a calibrated low-rank posterior around the resulting support. The relative contribution of the two support axes changes with the training setting: the spine represents longitudinal support, and our diagnostics quantify when the traversed corridor carries functional variation. Appendix~\ref{app:geometry_idealization} gives the full geometric motivation, figures, Fisher-motivated tangent/normal decomposition, and continuous tubular idealization.

\section{Tubular Low-Rank Laplace approximation}
\label{sec:trl}

TRL is a geometry-motivated, scalable post-hoc approximation to the weight-space posterior. The continuous tube is a conceptual reference; the implemented method is a discrete, transported low-rank approximation designed for scalable posterior predictive sampling from one MAP checkpoint. Its support and calibrated factors do not estimate exact posterior masses.

\subsection{Practical discrete TRL approximation}
\label{sec:trl-practical}

The implemented TRL posterior replaces the continuous tube by a discrete scalable approximation. It consists of:
\begin{enumerate}[leftmargin=*]
    \item a finite predictor--corrector spine $\{\gamma_t\}_{t=0}^{T-1}$ of $T$ stored anchors;
    \item a low-rank transverse basis initialized at the MAP and transported along the spine;
    \item local Gaussian transverse perturbations applied at discrete spine anchors;
    \item architecture-dependent batch-normalization recalibration, required when sampled networks contain BatchNorm layers.
\end{enumerate}

Starting from one trained MAP, TRL computes dominant transverse structure once, then transports, projects, and re-orthogonalizes it along the spine. This amortization assumes that, over sufficiently short steps, the stiff subspace remains informative. Appendix~\ref{app:continuous_to_discrete_trl} details the predictor--corrector update, projection--QR transport, and discrete mixture.

\paragraph{Spine construction.}
Starting from $\gamma_0=\theta_{\mathrm{MAP}}$, Lanczos/HVPs identify the stiff transverse subspace $N_0$, and the initial unit tangent satisfies $v_{\parallel,0}\in\operatorname{span}(N_0)^\perp$. For large-network CIFAR/ImageNet runs, $v_{\parallel,0}$ is a seeded random vector projected into this complement and normalized; toys instead use the smallest-eigenvalue eigenvector. The corrector subtracts the gradient component orthogonal to the tangent; norm control limits the displacement, whose normalization gives the next tangent. For the large-network classification implementation, this gradient uses the mini-batch mean cross-entropy $\ell_{\mathcal B_t}(\theta)=-|\mathcal B_t|^{-1}\sum_{(x,y)\in\mathcal B_t}\log p_\theta(y\mid x)$, without a prior-gradient term. This differs from the negative log posterior $\mathcal L$; the prior enters the transverse precision separately. The spine is not an exact geodesic or smallest-eigenvector sequence; its nominal extent is $(T-1)\Delta s$. Appendix~\ref{app:continuous_to_discrete_trl} derives the local projected-corrector rationale and its limits.

\paragraph{Transported transverse structure.}
At each spine anchor $\gamma_t$, let $N_t\in\mathbb{R}^{K\times k_\perp}$ denote the low-rank transverse basis used for sampling, where $K$ is the number of trainable parameters and $k_\perp$ is the transverse rank. The columns of $N_t$ are the parameter-space directions in which TRL applies Gaussian transverse perturbations. At the MAP, $N_0$ is computed from dominant data-sensitive curvature directions using implicit curvature-vector products. These directions define the stiff subspace avoided by longitudinal motion; the prior affects scale, not selection. Along the spine, the basis is transported rather than recomputed. After each predictor--corrector update, we refresh the tangent from the realized displacement. Here $d_t$ is the realized predictor--corrector displacement from $\gamma_t$ to $\gamma_{t+1}$, whose predictor term follows $v_{\parallel,t}$; normalizing $d_t$ gives the next tangent:
\begin{align}
d_t &= \gamma_{t+1}-\gamma_t,
&
v_{\parallel,t+1} &= \frac{d_t}{\|d_t\|}.
\label{eq:tangent-main}
\end{align}
Using the refreshed tangent from Eq.~\eqref{eq:tangent-main}, we project the previous transverse frame away from this tangent and re-orthogonalize:
\begin{align}
\widetilde N_{t+1}
&=
\left(I-v_{\parallel,t+1}v_{\parallel,t+1}^{\top}\right)N_t,
&
N_{t+1}
&=
\operatorname{qf}\!\left(\widetilde N_{t+1}\right).
\label{eq:transport-main}
\end{align}
Equation~\eqref{eq:transport-main} defines the transported-frame update; $\operatorname{qf}(\cdot)$ denotes the orthonormal $Q$ factor from a thin QR decomposition. The resulting $N_t$ should therefore be interpreted as a transported sampling frame, not as the eigenbasis of a freshly decomposed local curvature matrix at $\gamma_t$.

The practical factor $L_{\perp,t}$ is diagonal in transported coordinates. If $\widehat\lambda_i$ is the $i$th retained MAP curvature value, $n_{t,i}$ is the $i$th column of $N_t$, and $\Lambda_{\rm prior}$ is the diagonal prior-precision matrix, the implementation uses
\begin{equation}
L_{\perp,t}
=
\operatorname{diag}_{i=1}^{k_\perp}
\left[
\max\!\left\{\widehat\lambda_i+n_{t,i}^{\top}\Lambda_{\rm prior}n_{t,i},10^{-6}\right\}^{-1/2}
\right].
\label{eq:transverse-factor-main}
\end{equation}
In Eq.~\eqref{eq:transverse-factor-main}, the curvature values are fixed at the MAP, while the prior projection follows the transported frame. Here $\widehat\lambda_i$ are eigenvalues of the mini-batch-mean cross-entropy Hessian (per-example scale), whereas $\Lambda_{\rm prior}$ inherits a prior fitted by marginal likelihood from likelihood curvature accumulated over a 5,000-example subset; $\beta_\perp$ and $c$ therefore calibrate a tempered transverse Gaussian rather than a matched full-data Laplace covariance. This is an inverse-square-root precision factor, not a dense Cholesky factor of freshly recomputed local curvature; $N_tL_{\perp,t}z_\perp$ maps a transverse Gaussian draw into parameter space.

\paragraph{Discrete longitudinal sampling.}
The practical sampler represents the longitudinal coordinate by sampling $t$ uniformly from the stored anchors $\{0,\ldots,T-1\}$ and then drawing a transverse Gaussian perturbation at that anchor; it does not sample a continuous $z_\parallel$ or evaluate the smooth longitudinal map of Appendix~\ref{sec:trl-ideal}. A practical sample has the form
\begin{equation}
\theta^{(r)}
= 
\gamma_t + N_t\bigl(\beta_\perp L_{\perp,t} z^{(r)}_\perp\bigr),
\quad z^{(r)}_\perp\sim\mathcal{N}(0,I_{k_\perp}),
\label{eq:practical-trl-sample}
\end{equation}
where $t$ indexes a spine anchor and $\beta_\perp$ is a global calibration/tempering multiplier on transverse spread. Thus longitudinal uncertainty is the discrete anchor distribution, while the Gaussian precision/covariance factor is applied only in the transported transverse coordinates; Appendix~\ref{app:continuous_to_discrete_trl} derives this continuous-to-discrete mixture explicitly. The implemented posterior is best understood as a discrete mixture of local low-rank Gaussian perturbations along a discrete spine, not as a smooth global Gaussian pushforward. The MAP-centered single-checkpoint tube is the collapsed-spine special case of the same family: the anchor distribution becomes a point mass at $\gamma_0=\theta_{\mathrm{MAP}}$, while the curvature-selected transverse Gaussian is retained.

\paragraph{Empirical effect of amortization.}
Appendix~\ref{app:transport} compares transported, fixed-MAP, single-checkpoint, and fresh-refresh variants. On CIFAR-100/ResNet-18, the stiff eigenspace rotates, but fresh refresh (a sampling-frame check; Appendix~\ref{app:additional_orthogonality}) changes predictions only modestly and adds substantial cost; a MAP-centered tube already captures most gains. In this setting, computing the transverse structure once at the MAP provides the better cost--fidelity trade-off, although fresh local factors may matter in other regimes.

\paragraph{Batch-normalization recalibration.}
For architectures with batch normalization, each sampled network can have shifted activation statistics. Evaluating sampled weights with the MAP running statistics would mismatch the sampled parameters and BatchNorm state. We therefore use FixBN as an inference-time validity step: after sampling weights, batch-normalization statistics are recalibrated by forward passes on a training subset before evaluating predictions. FixBN is executed inside the posterior-prediction stages. Appendix~\ref{app:runtime_measurement} explains the timing scopes and overlapping FixBN sub-timers used in Table~\ref{tab:app_runtime_memory}, while Appendix~\ref{app:ablations} reports the rolling-mode batch-count sensitivity.

\subsection{Algorithmic summary}
\label{sec:trl-algorithm}

Algorithm~\ref{alg:trl_inference} summarizes the large-network classification pipeline. The cross-entropy corrector constructs the spine, while MAP eigenvalues and projected prior precision define the transverse factors. Predictions average samples around one trained solution; no independently trained predictive models are introduced.

\begin{algorithm}[H]
\caption{Practical TRL inference with transported transverse structure}
\label{alg:trl_inference}
\begin{algorithmic}[1]
\Require MAP $\theta_{\mathrm{MAP}}$, query input $x_*$, step $\Delta s$, stored-anchor count $T$, rank $k_\perp$, scale $\beta_\perp$, correction strength $\eta_{\mathrm{corr}}$, sample count $S$.
\State \textbf{Initialize:} $\gamma_0\gets\theta_{\mathrm{MAP}}$.
\State Compute dominant transverse basis $N_0$ at the MAP via Lanczos/HVPs.
\State Draw a seeded random $v_{\parallel,0}$, project it into $\operatorname{span}(N_0)^\perp$, and normalize.
\For{$t=0$ to $T-1$}
    \State Form $L_{\perp,t}$ from MAP eigenvalues and the prior projected onto $N_t$.
    \State Store $\mathcal{T}_{t}=\{\gamma_t,N_t,L_{\perp,t}\}$.
    \If{$t<T-1$}
        \State $d_t\gets\Delta s\,v_{\parallel,t}-\eta_{\mathrm{corr}}(I-v_{\parallel,t}v_{\parallel,t}^{\top})\nabla_\theta\ell_{\mathcal B_t}(\gamma_t)$.
        \State Apply norm control to $d_t$; set $\gamma_{t+1}\gets\gamma_t+d_t$.
        \State Update $v_{\parallel,t+1}$ from the realized displacement; transport $N_t$ by tangent projection and QR.
    \EndIf
\EndFor
\For{$r=1$ to $S$}
    \State Sample $t$ uniformly from $\{0,\ldots,T-1\}$.
    \State Sample transverse coordinate $z_\perp^{(r)}\sim\mathcal{N}(0,I_{k_\perp})$.
    \State $\theta^{(r)}\gets \gamma_t+N_t(\beta_\perp L_{\perp,t}z_\perp^{(r)})$.
    \If{the network uses batch normalization}
        \State Recalibrate BN statistics for $\theta^{(r)}$ on a training subset.
    \EndIf
    \State Compute prediction $y^{(r)}\gets f(x_*;\theta^{(r)})$.
\EndFor
\State \Return predictive Monte Carlo mean and posterior functional variance from $\{y^{(r)}\}_{r=1}^S$.
\end{algorithmic}
\end{algorithm}

The sampler represents one longitudinal direction and $k_\perp$ transported transverse directions, not a complete posterior over all locally flat directions. It provides within-basin averaging and does not capture the across-basin diversity of Deep Ensembles.

\subsection{Predictive moments and interpretation}
\label{sec:trl-prediction}

Given posterior samples $\theta^{(r)}\sim q_{\mathrm{TRL}}$, Eqs.~\eqref{eq:pred_mean}--\eqref{eq:pred_var} evaluate the nonlinear network directly, without linearizing it around the MAP. For classification, $f_\theta(x)$ denotes the post-softmax class-probability vector. We report the Monte Carlo predictive mean and posterior functional variance:
\begin{align}
\hat{\mu}_*(x_*) &= \frac{1}{S}\sum_{r=1}^S f_{\theta^{(r)}}(x_*),\label{eq:pred_mean}\\
\widehat{\mathcal V}_f(x_*)
&=\frac{1}{S}\sum_{r=1}^S\left\|f_{\theta^{(r)}}(x_*)-\hat{\mu}_*(x_*)\right\|_2^2.\label{eq:pred_var}
\end{align}
For vector outputs, $\widehat{\mathcal V}_f$ is the trace of the empirical covariance of the predictive functions under the $1/S$ Monte Carlo normalization used in Eq.~\eqref{eq:pred_var}; for scalar regression it is the empirical posterior variance of the network output. It is the epistemic component of predictive variance, not the total $\operatorname{Var}(y_*\mid x_*,\mathcal D)$, which also includes conditional observation or class uncertainty.

\subsection{Scope and limitations of the approximation}
\label{sec:trl-scope}

The practical approximation makes several explicit choices. It uses a one-dimensional constructed spine, truncates transverse uncertainty to a low-rank subspace, transports transverse structure without a fresh local curvature factorization at every anchor, and uses a global scale parameter to calibrate transverse spread. Higher-dimensional low-loss manifolds are outside the present approximation. These choices are not incidental implementation details. They define the scalable approximation studied in this paper and should be evaluated as fidelity--cost trade-offs. A richer method could refresh curvature more frequently, use higher-dimensional spines, or combine within-basin geometry with cross-basin diversity, but such extensions would move the method to a different computational regime. Appendix~\ref{app:limitations} discusses these limitations and possible extensions in more detail.

\section{Relation to existing methods}
\label{sec:relation}

TRL is a post-hoc Laplace-family approximation, but it makes a different scalability cut from the Euclidean and linearized Laplace baselines used in our experiments. Here ``Laplace-family'' refers to the local Gaussian transverse fibers and curvature-informed construction; practical TRL is not a single second-order expansion of one globally specified posterior around an exact mode. Last-layer ELA/LLA restrict uncertainty to the classifier and factorize its curvature; TRL instead keeps full-network access through stochastic HVPs and restricts the posterior to a low-rank transverse subspace selected in the full parameter space, coupled to a discrete predictor--corrector spine whose contribution varies across training settings.

TRL is also related to geometric and Riemannian posterior approximations, low-rank subspace and subnetwork inference, mode connectivity, loss-tunnel samplers, and geometric Gaussian approximations~\citep{bergamin2023riemannian,yu2024fisherlaplace,izmailov2020subspace,daxberger2021subnetwork,garipov2018loss,draxler2018essentially,dold2024embedding,dacosta2025geometric,faller2026lowrank}. The computational distinction is that TRL amortizes support construction around one MAP solution and then samples closed-form low-rank Gaussian fibers, without claiming exact metric transport, repeated local covariance refits, or MCMC over a learned tunnel. Appendix~\ref{app:extended_related} gives the detailed method-by-method positioning, including the different approximation target of Faller and Martin, the random/PCA/curve subspaces of Izmailov et al., and the endpoint assumptions of mode-connectivity methods.

To keep the comparison sets fixed, we distinguish four computational regimes. \emph{Single-checkpoint post-hoc} methods start from one already-trained deterministic model and do not continue the original task training or fit additional independently optimized predictive models. Their post-hoc construction may still use training data, gradients, and nearby parameter-space updates; this computational regime includes MAP+TS, last-layer ELA/LLA, FM-LR-LLA ValScale, and TRL, with MAP serving as the deterministic reference. ELA/LLA, FM-LR-LLA, and TRL return posterior approximations, whereas MAP+TS is a scalar non-posterior calibration control. SWAG-Diag is \emph{checkpoint-initialized} but follows a short additional optimization trajectory to collect diagonal weight statistics. MC~Dropout is a \emph{single-training, training-time stochastic} method whose predictive dispersion is introduced during training. Deep Ensembles require multiple independent trainings. Thus, ``single-training'' is broader than ``single-checkpoint post-hoc,'' and TRL specifically asks how much posterior averaging can be recovered from one deterministic checkpoint by selecting a useful full-network low-rank support.


\section{Experiments}
\label{sec:experiments}

\textbf{Protocol.}
We compare TRL with MAP, last-layer ELA/LLA, FM-LR-LLA ValScale \citep{faller2026lowrank}, SWAG-Diag, MC Dropout, and Deep Ensembles. TRL, MAP+TS, and FM-LR-LLA ValScale use held-out clean-validation NLL for scale/temperature selection; headline ELA/LLA instead use marginal-likelihood prior precision, while a validation-NLL prior grid sends both to $\lambda=10^4$, effectively MAP (Appendix~\ref{app:calibration_sanity}). OOD AUROC uses predictive entropy. The main CIFAR-100 TRL setting is $k_\perp=30$, $T=40$, $\beta_\perp=4.0$, $S=25$, with 25-batch FixBN. Appendix~\ref{app:implementation} gives the complete split, seed, tuning, and baseline protocols, including the rolling-versus-reset FixBN sensitivity audit.

\textbf{Experimental roles.} Headline tables aggregate independently trained checkpoints; fixed-MAP ablations vary posterior draws, while single-checkpoint/fresh-refresh and FixBN/replay studies are mechanism or provenance diagnostics rather than benchmark replicates.

At CIFAR scale, ELA/LLA use last-layer K-FAC/KRON curvature; FM-LR-LLA ValScale is a checkpoint-, split-, and rank-matched matrix-free all-weight Kronecker reference with a validation-scaled rank-$30$ predictive-covariance subspace. The prior families differ: FM uses one scalar precision, whereas TRL uses its validation-selected two-block head/backbone prior. Appendix~\ref{app:faller_martin_comparison} gives the FM diagnostics; Appendices~\ref{app:trl_vs_ela_support} and~\ref{app:transport} give random-support and transport controls.

\subsection{CIFAR-100 clean test performance}
Table~\ref{tab:cifar100_clean_5seed} reports the five-checkpoint clean comparison. Deep Ensembles retain the best reported mean accuracy, NLL, and Brier at five-training cost; TRL has the lowest reported mean clean 15-bin equal-width top-label ECE. MAP+TS preserves the checkpoint argmax and improves NLL/ECE/Brier to $0.960/0.028/0.355$; TRL remains lower at $0.951/0.013/0.353$. All five checkpoint-matched ECE differences favor TRL (mean difference $-0.0147$; TRL mean about $48\%$ of MAP+TS). OOD ranking is distinct: last-layer LLA has the highest SVHN AUROC despite poor clean likelihood and calibration.

The low-rank controls narrow the comparison. FM-LR-LLA improves over MAP but remains above TRL in NLL, ECE, and Brier. Historical random-support results depend on the scale range; the separate matched-prior diagnostic reports lower ECE but numerically higher NLL for curvature-selected support (Appendix~\ref{app:orientation_checkpoint}). The remaining differences include curvature-selected support, TRL's two-block prior, and nonlinear prediction. FixBN is the BatchNorm-consistency step required to evaluate sampled networks and does not define posterior support. MAP+TS shows that scalar recalibration explains most of the NLL/Brier correction, while TRL retains lower top-label ECE and sample-based functional diagnostics. Appendix~\ref{app:calibration_sanity} confirms the top-label ordering under equal-width, equal-mass, and KS checks; classwise ECE slightly favors MAP+TS, and Deep Ensembles are numerically best under some estimators.

\begin{table*}[t]
\centering
\caption{CIFAR-100 clean test performance (five independently trained MAP checkpoints; mean $\pm$ sample SD; one TRL posterior per checkpoint). ECE is 15-bin equal-width top-label ECE. MAP+TS fits one positive temperature per checkpoint by clean-validation NLL and preserves the checkpoint argmax. FM-LR-LLA ValScale is the rank-matched all-weight Kronecker comparison; Appendix~\ref{app:faller_martin_comparison} gives protocol details. Text comparisons use unrounded means; see Appendix~\ref{app:calibration_metrics}.}
\label{tab:cifar100_clean_5seed}
\setlength{\tabcolsep}{2pt}
\begin{tabular}{lccccc}
\toprule
Method & Acc $\uparrow$ & NLL $\downarrow$ & ECE $\downarrow$ & Brier $\downarrow$ & AUROC $\uparrow$ \\
\midrule
Deep Ensemble & \meanstd{0.779}{0.001} & \meanstd{0.805}{0.007} & \meanstd{0.017}{0.002} & \meanstd{0.308}{0.003} & \meanstd{0.862}{0.008} \\
MAP & \meanstd{0.743}{0.003} & \meanstd{1.040}{0.008} & \meanstd{0.096}{0.002} & \meanstd{0.370}{0.003} & \meanstd{0.849}{0.021} \\
MAP+TS & \meanstd{0.743}{0.003} & \meanstd{0.960}{0.006} & \meanstd{0.028}{0.004} & \meanstd{0.355}{0.003} & \meanstd{0.870}{0.022} \\
SWAG-Diag & \meanstd{0.744}{0.003} & \meanstd{1.040}{0.007} & \meanstd{0.096}{0.003} & \meanstd{0.369}{0.003} & \meanstd{0.852}{0.020} \\
MC Dropout & \meanstd{0.742}{0.007} & \meanstd{0.953}{0.021} & \meanstd{0.048}{0.005} & \meanstd{0.357}{0.007} & \meanstd{0.780}{0.027} \\
ELA (last-layer) & \meanstd{0.724}{0.002} & \meanstd{1.356}{0.013} & \meanstd{0.262}{0.008} & \meanstd{0.468}{0.004} & \meanstd{0.874}{0.021} \\
LLA (last-layer) & \meanstd{0.739}{0.002} & \meanstd{1.492}{0.017} & \meanstd{0.368}{0.008} & \meanstd{0.521}{0.006} & \meanstd{0.903}{0.008} \\
FM-LR-LLA ValScale & \meanstd{0.743}{0.002} & \meanstd{0.985}{0.006} & \meanstd{0.041}{0.010} & \meanstd{0.358}{0.003} & \meanstd{0.843}{0.026} \\
TRL (ours) & \meanstd{0.744}{0.003} & \meanstd{0.951}{0.006} & \meanstd{0.013}{0.002} & \meanstd{0.353}{0.003} & \meanstd{0.867}{0.021} \\
\bottomrule
\end{tabular}
\end{table*}

\subsection{CIFAR-100-C robustness and calibration}

Table~\ref{tab:cifar100c_main_3seed} averages 19 corruptions and five severities over three independently trained, seed-matched checkpoints. Deep Ensembles lead every reported mean. Among posterior approximations, TRL has the lowest NLL, ECE, and Brier at MAP-like accuracy. The frozen clean-validation MAP+TS control has lower NLL than TRL in all three paired checkpoints, essentially the same aggregate ECE, and a slightly higher Brier; neither method is tuned on corruption. The comparison therefore separates two effects: scalar recalibration remains a strong control under shift, while TRL additionally provides a sampled full-network posterior and support diagnostics.

\begin{table*}[t]
\centering
\caption{CIFAR-100-C over 19 corruptions and five severities (mean $\pm$ sample SD over three seed-matched checkpoints). TRL uses the matched MAP checkpoint; MAP+TS fits one clean-validation temperature and freezes it for all corruptions.}
\label{tab:cifar100c_main_3seed}
\begin{tabular}{lcccc}
\toprule
Method & Acc $\uparrow$ & NLL $\downarrow$ & ECE $\downarrow$ & Brier $\downarrow$ \\
\midrule
Deep Ensemble & \meanstd{0.509}{0.001} & \meanstd{2.292}{0.012} & \meanstd{0.114}{0.002} & \meanstd{0.650}{0.003} \\
MAP & \meanstd{0.471}{0.004} & \meanstd{2.776}{0.033} & \meanstd{0.234}{0.003} & \meanstd{0.747}{0.005} \\
MAP+TS & \meanstd{0.471}{0.004} & \meanstd{2.390}{0.041} & \meanstd{0.118}{0.008} & \meanstd{0.691}{0.007} \\
SWAG-Diag & \meanstd{0.473}{0.005} & \meanstd{2.774}{0.062} & \meanstd{0.234}{0.006} & \meanstd{0.745}{0.008} \\
MC Dropout & \meanstd{0.475}{0.006} & \meanstd{2.554}{0.060} & \meanstd{0.172}{0.008} & \meanstd{0.707}{0.010} \\
TRL (ours) & \meanstd{0.473}{0.004} & \meanstd{2.464}{0.047} & \meanstd{0.118}{0.006} & \meanstd{0.689}{0.005} \\
\bottomrule
\end{tabular}
\end{table*}

\subsection{Ablations and regime diagnostics}

ECE varies non-monotonically with transverse rank (Table~\ref{tab:app_ablation_k}), with $k_\perp=30$ giving the best ECE among the five evaluated ranks. Transported and fixed-MAP bases are nearly tied, fresh refresh changes little, and a single-checkpoint tube captures most gains in the main CIFAR regime.

\begin{figure}[t]
\centering
\includegraphics[width=0.64\linewidth]{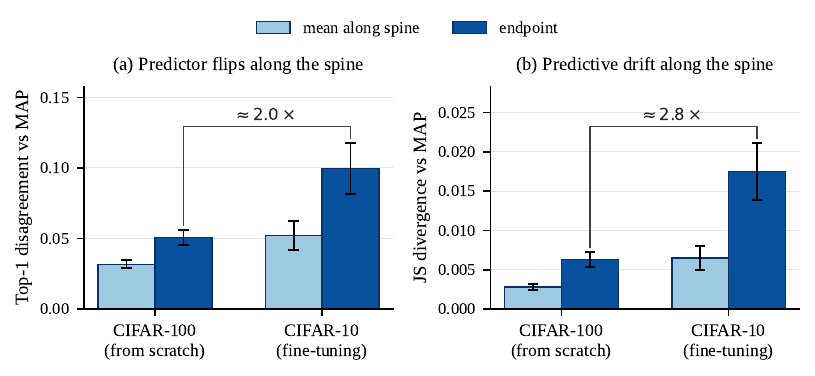}
\caption{\textbf{Longitudinal variation differs across regimes.} Endpoint top-1 disagreement and JS drift are about $2\times$ and $2.8\times$ larger under small-data fine-tuning (sample SD over five and ten runs).}
\label{fig:regime_spine_dependence}
\end{figure}

Figure~\ref{fig:regime_spine_dependence} shows greater longitudinal variation under small-data adaptation. Endpoint top-1 disagreement/JS are $5.08\%\pm0.54\%$/$0.0063\pm0.0009$ on CIFAR-100 from scratch and $9.95\%\pm1.80\%$/$0.0175\pm0.0036$ after CIFAR-100$\rightarrow$CIFAR-10 fine-tuning. In fine-tuning, full-spine mixing increases functional diversity while retaining mean NLL and Brier comparable to the validation-selected single-anchor control. The mixture keeps support across the discovered corridor without a separate anchor-selection rule. Appendices~\ref{app:transport} and~\ref{app:finetune_spine_diagnostic} give the full comparisons; Appendices~\ref{app:ablations} and~\ref{app:imagenet_scale_check} give the remaining ablations and ImageNet checks.

Transported--fresh top-30 eigenspace overlap falls to $0.198\pm0.021$ at the CIFAR endpoint, with mean principal angle $70.8^\circ\pm1.8^\circ$. Nevertheless, fixed and transported frames give nearly identical predictive scores; recomputing eigenspaces adds $89.2\pm7.7$ minutes per checkpoint (Appendix~\ref{app:transport}). Fresh/fixed controls also change tangent orthogonality, so these are sampling-frame checks rather than one-factor refresh ablations. Appendix~\ref{app:additional_diagnostics} adds paired full-precision/centering tests and explicit tangent diagnostics.

\begin{table}[H]
\centering
\caption{Compact regime comparison. CIFAR-100 rows summarize Tables~\ref{tab:app_basis_ablation} and~\ref{tab:app_fresh_refresh_single} over three checkpoint-specific runs; fine-tuning rows reproduce Table~\ref{tab:app_finetune_spine_diagnostic} over ten seeds. Comparisons should be made within each block.}
\label{tab:compact_regime_comparison}
\footnotesize
\setlength{\tabcolsep}{3.2pt}
\begin{tabular}{llcccc}
\toprule
Regime & Variant & NLL $\downarrow$ & ECE $\downarrow$ & Brier $\downarrow$ & Avg. func. var. \\
\midrule
CIFAR-100 & single checkpoint & $0.958\pm0.005$ & $0.015\pm0.003$ & $0.357\pm0.002$ & -- \\
CIFAR-100 & transported full spine & $0.956\pm0.004$ & $0.016\pm0.003$ & $0.355\pm0.001$ & -- \\
CIFAR-100 & fresh refresh & $0.955\pm0.008$ & $0.013\pm0.003$ & $0.356\pm0.002$ & -- \\
Fine-tuning & single checkpoint & $0.649\pm0.014$ & $0.019\pm0.003$ & $0.315\pm0.005$ & $0.00078\pm0.00010$ \\
Fine-tuning & best single (val.) & $0.647\pm0.015$ & $0.019\pm0.004$ & $0.314\pm0.006$ & $0.00094\pm0.00008$ \\
Fine-tuning & endpoint single & $0.647\pm0.014$ & $0.019\pm0.004$ & $0.314\pm0.006$ & $0.00096\pm0.00011$ \\
Fine-tuning & full spine & $0.647\pm0.015$ & $0.017\pm0.003$ & $0.314\pm0.006$ & $0.00230\pm0.00043$ \\
\bottomrule
\end{tabular}
\end{table}
As Table~\ref{tab:compact_regime_comparison} shows, in the fine-tuning diagnostic center relocation accounts for most of the mean NLL/Brier change: both the validation-selected and endpoint-centered tubes are close to the full-spine mixture. A material NLL or Brier improvement over the endpoint-centered Gaussian is not established. Mixing primarily increases functional diversity, raising average functional variance to $2.30\times10^{-3}$ from roughly $0.9\times10^{-3}$ for the single-anchor controls.

The endpoint is a strong ex-post control, but its location is fixed by path length and step schedule rather than by a predictive criterion. Validation selection provides such a criterion by evaluating six candidate anchors; uniform mixing instead retains support across the discovered corridor without selecting one anchor. The larger fine-tuning disagreement also coincides with more loss drift (endpoint validation cross-entropy change $0.0594\pm0.0239$, versus about $0.010$ from scratch), so functional variation and low loss should be read separately.

The orientation diagnostic separates transverse support from longitudinal mixing: both K and RB are centered at the MAP. K has lower ECE in all five checkpoints, with larger differences in checkpoints 0--2; RB has numerically lower NLL in all five (Table~\ref{tab:orientation_checkpoint}). This supports a metric-specific distinction between transverse subspaces, not an additional benefit from the spine.

Across the two evaluated regimes, the balance between transverse and longitudinal support changes. A MAP-centered tube is sufficient when longitudinal variation is weak, whereas the full spine preserves additional diversity when the corridor contains distinct predictors of comparable quality. The ordering is regime-specific: representative counterexamples are LLA on OOD AUROC; MAP+TS on CIFAR-100-C NLL and ImageNet NLL/ECE; MC~Dropout on WideResNet-16-4 accuracy/NLL/Brier and VGG-11-BN accuracy/NLL/ECE/Brier; and Deep Ensembles on clean CIFAR-100 accuracy/NLL/Brier and ambiguous-CIFAR-10H soft Brier/JS (Appendix~\ref{app:cifar10h}).

\section{Conclusion}
\label{sec:conclusion}

TRL combines a curvature-selected full-network transverse subspace with a discrete predictor--corrector spine in a single-checkpoint posterior approximation. On CIFAR-100, it improves NLL and top-label calibration over the evaluated Laplace baselines, with most gains already captured by the MAP-centered tube. On ImageNet, that tube lowers checkpoint ECE, but temperature scaling is lower in NLL and ECE. The longitudinal contribution is regime-dependent: in small-data fine-tuning, spine mixing increases functional diversity while retaining mean NLL and Brier comparable to the evaluated single-anchor controls. On sine, mixing has lower NLL than globally development-selected single-center controls, while tuned LLA remains better; tuned classification scores are similar (Appendix~\ref{app:toy}). The audited sine paths do not preserve near-MAP training loss; their mixture scores do not establish a low-loss geometric mechanism.

\section*{Reproducibility Statement}
The ancillary archive includes code, checksums, artifacts and reproduction commands. The manuscript source is available from arXiv. Appendices~\ref{app:implementation} and~\ref{app:additional_diagnostics} document protocols, rolling-FixBN dependencies and replay checks. Selection excludes reporting-set, corruption and CIFAR-10H labels; toy scale controls use separate development seeds.

\newpage
\bibliography{references}
\bibliographystyle{iclr2027_conference}

\newpage
\appendix
\onecolumn

\section{Background on Laplace, Fisher/GGN, and Calibration Metrics}
\label{app:background}

This appendix provides a short self-contained background for readers less familiar with Bayesian neural networks, Laplace approximations, and the geometric language used in the main text. The goal is not to survey the full literature, but to define the objects that appear in TRL and in the experimental baselines.

\subsection{Bayesian prediction and MAP checkpoints}
A Bayesian neural network places a prior $p(\theta)$ over network parameters and predicts by marginalizing over the posterior \citep{gal2016thesis,kendall2017uncertainties,rasmussen2006gp},
\[
 p(y_*\mid x_*,\mathcal D)
 = \int p(y_*\mid x_*,\theta)\,p(\theta\mid\mathcal D)\,d\theta .
\]
For modern deep networks this integral is intractable. Practical Bayesian deep learning therefore replaces $p(\theta\mid\mathcal D)$ by a tractable approximation built around a trained solution. The maximum a posteriori (MAP) parameter $\theta_{\rm MAP}$ minimizes the negative log posterior,
\[
\mathcal L(\theta)
= -\log p(\mathcal D\mid\theta)-\log p(\theta),
\]
and serves as the deterministic checkpoint to which post-hoc uncertainty methods attach posterior mass. In the experiments, MAP denotes this deterministic model and also the center from which Laplace, SWAG-Diag, and TRL variants start.

A useful distinction is between \emph{parameter uncertainty} and \emph{predictive uncertainty}. A posterior over weights is only useful through the distribution over functions it induces. Two parameter perturbations can be large in Euclidean norm yet nearly identical functionally because of symmetries, normalization effects, or flat directions; conversely, a small movement in a data-sensitive direction can change predictions substantially. This is why the paper repeatedly separates geometric movement in parameter space from functional diagnostics such as disagreement, Jensen--Shannon divergence, and functional variance.

\subsection{Euclidean and linearized Laplace}
\label{app:laplace_background}
The Euclidean Laplace approximation (ELA) fits a Gaussian around the MAP by using the local quadratic approximation
\begin{equation}
\mathcal L(\theta_{\rm MAP}+\delta)
\approx \mathcal L(\theta_{\rm MAP}) + \tfrac12 \delta^\top G_0 \delta,
\qquad
q_{\rm ELA}(\theta)=\mathcal N(\theta_{\rm MAP},G_0^{-1}).
\label{eq:ela-background}
\end{equation}
In Eq.~\eqref{eq:ela-background}, $\delta\in\mathbb R^K$ is a parameter displacement and $G_0$ is a curvature matrix such as a Hessian, Fisher, or generalized Gauss--Newton (GGN) matrix plus prior precision \citep{mackay1992practical,daxberger2021laplace}. If $u_i$ is an eigenvector of the likelihood-curvature component with eigenvalue $\lambda_i$ and an isotropic prior contributes precision $\lambda$, the Laplace variance along $u_i$ is approximately $(\lambda_i+\lambda)^{-1}$. Thus high-curvature directions receive small posterior variance, while low-curvature directions receive larger variance. In a neural network, however, the full $K\times K$ curvature is too large to materialize at CIFAR scale, so practical Laplace baselines use restrictions or approximations such as last-layer curvature, Kronecker factorizations, diagonal structure, or low-rank subspaces.

For classification, let $a_\theta(x)$ denote the pre-link logit vector and $p_\theta(x)$ the post-link class-probability vector. Linearized Laplace approximation (LLA) uses the same local Gaussian weight posterior but propagates it through the first-order Taylor expansion of the logits,
\begin{equation}
a_{\theta_{\rm MAP}+\delta}(x)
\approx a_{\theta_{\rm MAP}}(x) + J^a_{\theta_{\rm MAP}}(x)\delta .
\label{eq:lla-linearization}
\end{equation}
Here $J^a_{\theta_{\rm MAP}}(x)=\left.\partial a_\theta(x)/\partial\theta\right|_{\theta=\theta_{\rm MAP}}$ is the logit Jacobian at the MAP. For regression, the analogous expansion applies directly to the model output. Equation~\eqref{eq:lla-linearization} therefore yields a Gaussian approximation in logit space for classification, or function space for regression, before the likelihood/link approximation is applied \citep{kristiadi2020being,daxberger2021laplace}. Thus ELA and LLA differ mainly in how predictions are computed from the local Gaussian: ELA evaluates sampled nonlinear networks, whereas LLA evaluates the linearized model, often with a probit or moment-matching approximation for classification. Section~\ref{sec:trl-prediction}, by contrast, uses $f_\theta(x)=p_\theta(x)$ for classification when defining TRL predictive moments. In the toy experiments, full-network curvature is small enough that ELA and LLA can be run as full-parameter baselines. In the CIFAR-scale experiments, the reported ELA/LLA baselines use last-layer Kronecker-factored Laplace for tractability, as specified in Appendix~\ref{app:implementation}.

\subsection{Fisher/GGN geometry and data-sensitive directions}
Information geometry equips parameter space with a local metric derived from predictive sensitivity rather than Euclidean distance \citep{amari1998natural,bonnabel2013sgdriem,kristiadi2023geometry}. In the Fisher/GGN view, the quadratic form
\[
 v^\top F(\theta)v
\]
measures how strongly a small parameter displacement $v$ changes the model's predictions on the data, at least locally. Large values indicate data-sensitive or stiff directions; small values indicate directions that are approximately functionally invariant under the local metric. For a generalized Gauss--Newton matrix, this statement is especially transparent because $F$ has the form of a Jacobian-transpose--curvature--Jacobian operator, so $Fv\approx0$ is associated with $Jv\approx0$ in the relevant data-weighted sense.

TRL uses this sensitivity picture to motivate its tangent/transverse decomposition. A low-loss region may contain directions along which the network can move while leaving training loss and predictions nearly unchanged, together with transverse directions where predictions change rapidly. The continuous tube in the main text idealizes these roles as a longitudinal spine with transverse Gaussian uncertainty. The implementation uses a lower-cost approximation: Hessian-vector products select a low-rank set of dominant data-sensitive directions, a finite spine is constructed in a low-curvature complement, and the transverse representation is transported along that spine. The geometric language organizes posterior support into components that can be approximated at scale; it is not a claim of exact geodesic sampling.

\subsection{Continuous TRL and its discrete approximation}
\label{app:continuous_to_discrete_trl}

The continuous tubular idealization writes the TRL posterior as a longitudinal mixture of transverse Gaussian fibers,
\begin{align}
q_{\mathrm{TRL}}(d\theta)
=
\int
\left(\exp_{\gamma(s)}\right)_{\#}
\mathcal N_{\perp}\!\left(0,\Sigma_{\perp}(s)\right)
\,\rho(ds).
\label{eq:continuous-trl-appendix}
\end{align}
The objects in Eq.~\eqref{eq:continuous-trl-appendix} are as follows:
\begin{itemize}
    \item $\gamma(s)$ is the point on the ideal low-loss spine at longitudinal coordinate $s$.
    \item $\mathcal N_{\perp}(0,\Sigma_{\perp}(s))$ is a Gaussian supported only in the normal/transverse space at that point.
    \item For a measurable map $\phi$ and measure $\mu$, $\phi_{\#}\mu$ denotes the pushforward measure, $(\phi_{\#}\mu)(A)=\mu(\phi^{-1}(A))$; thus $(\exp_{\gamma(s)})_{\#}$ maps the transverse Gaussian back into parameter space.
    \item $\rho(ds)$ is the longitudinal mixing measure, i.e., the probability weight assigned to choosing position $s$ along the spine.
\end{itemize}
Equation~\eqref{eq:continuous-trl-appendix} is a geometric reference object: it separates longitudinal support from transverse uncertainty, but the implementation does not evaluate it by continuous quadrature. In Eq.~\eqref{eq:trl-general} (Appendix~\ref{sec:trl-ideal}), the tubular map is the flat, first-order normal-chart approximation to the exponential-map idealization. Under a locally Euclidean chart, the pushforward by $\exp_{\gamma(s)}$ is approximated by the affine map $\gamma(s)+N(s)L_\perp(s)z_\perp$. The released implementation uses this affine chart at each spine anchor, written below as $\psi_t$.

The implemented sampler replaces this object by an empirical mixture over discrete spine anchors. With discrete spine anchors $\{\gamma_t\}_{t=0}^{T-1}$ and transported frames $N_t$, the longitudinal measure is approximated by the uniform empirical measure,
\begin{align}
\rho(ds)
\approx
\frac{1}{T}\sum_{t=0}^{T-1}\delta_{s_t}(ds),
\label{eq:discrete-longitudinal-measure}
\end{align}
where $s_t$ is the longitudinal coordinate associated with anchor $\gamma_t$ in the continuous idealization and $\delta_{s_t}$ is the Dirac probability measure concentrated at $s_t$. Substituting the empirical measure in Eq.~\eqref{eq:discrete-longitudinal-measure} gives the practical posterior
\begin{align}
q_{\mathrm{TRL}}^{(T)}(d\theta)
=
\frac{1}{T}
\sum_{t=0}^{T-1}
\left(\psi_t\right)_{\#}
\mathcal N(0,I_{k_\perp}),
\qquad
\psi_t(\xi)
=
\gamma_t+N_t\!\left(\beta_\perp L_{\perp,t}\xi\right).
\label{eq:discrete-trl-mixture}
\end{align}
In Eq.~\eqref{eq:discrete-trl-mixture}, $\xi\sim\mathcal N(0,I_{k_\perp})$ is the transverse coordinate and $L_{\perp,t}$ is the diagonal inverse-square-root precision factor of Eq.~\eqref{eq:transverse-factor-main}. The equation is the implemented discrete mixture: each posterior draw first samples an anchor index $t$ uniformly, then draws a rank-$k_\perp$ transverse Gaussian perturbation in the transported coordinates. Uniform anchor weighting is a practical heuristic; it is not claimed to represent equal local evidence, equal posterior volume, or an exact quadrature rule along the corridor. The main CIFAR-100 configuration uses $T=40$ discrete spine anchors to define this discrete mixture and $S=25$ Monte Carlo draws from $q_{\mathrm{TRL}}^{(T)}$ to estimate posterior predictive probabilities.

The large-network classification spine uses mini-batch mean cross-entropy $\ell_{\mathcal B_t}$, defined in Section~\ref{sec:trl-practical}, rather than the negative log posterior $\mathcal L$. No prior gradient is added to this corrector; prior precision enters the transverse factors. If $v_{\parallel,t}$ is the current unit tangent, the un-clipped update has the form
\begin{align}
\gamma_{t+1}
\leftarrow
\gamma_t
+
\Delta s\,v_{\parallel,t}
-
\eta_{\mathrm{corr}}
\left(I-v_{\parallel,t}v_{\parallel,t}^{\top}\right)
\nabla_\theta \ell_{\mathcal B_t}(\gamma_t).
\label{eq:spine-step}
\end{align}
For the unclipped displacement $d_t=\Delta s\,v_{\parallel,t}-\eta_{\rm corr}\Pi_t g_t$, with $\Pi_t=I-v_{\parallel,t}v_{\parallel,t}^\top$ and $g_t=\nabla\ell_{\mathcal B_t}(\gamma_t)$, the same minibatch loss has first-order change
\[
g_t^\top d_t=\Delta s\,g_t^\top v_{\parallel,t}-\eta_{\rm corr}\lVert\Pi_tg_t\rVert_2^2.
\]
The corrector therefore contributes a nonpositive term without directly canceling the prescribed longitudinal step; the total change can remain positive. This is a local identity for the minibatch objective, not a monotonicity statement for validation loss or the posterior objective.

Equation~\eqref{eq:spine-step} is implemented with additional norm control in the released code. Here $\Delta s$ is the prescribed longitudinal step size along the current tangent direction, and $\eta_{\mathrm{corr}}$ controls the strength of the projected gradient correction. The first term advances along the current longitudinal direction, while the projected corrector subtracts the component of the cross-entropy gradient orthogonal to the tangent in order to keep the trajectory in a low-loss corridor. This is not an exact geodesic update; it is a practical low-curvature-complement trajectory constructed from stochastic gradients and HVP-selected transverse directions.

After the realized update, the tangent and transported frame are refreshed using the actual displacement,
\begin{align}
d_t &= \gamma_{t+1}-\gamma_t,
&
v_{\parallel,t+1} &= \frac{d_t}{\|d_t\|},
\label{eq:tangent-refresh}
\end{align}
and the transverse basis is updated by projection and QR re-orthogonalization,
\begin{align}
\widetilde N_{t+1}
&=
\left(I-v_{\parallel,t+1}v_{\parallel,t+1}^{\top}\right)N_t,
&
N_{t+1}
&=
\operatorname{qf}\!\left(\widetilde N_{t+1}\right).
\label{eq:frame-transport}
\end{align}
Here $\operatorname{qf}(A)$ denotes the orthonormal $Q$ factor of a thin QR decomposition. Equations~\eqref{eq:tangent-refresh} and~\eqref{eq:frame-transport} define the projection--QR step that preserves orthogonality to the refreshed tangent, but it does not recompute the local Fisher/GGN eigenspace at $\gamma_{t+1}$. The diagnostics in Appendix~\ref{app:transport} explicitly measure the consequences of this amortization by comparing transported, fixed-MAP, and freshly recomputed variants.

Finally, the posterior predictive distribution is estimated by Monte Carlo averaging,
\begin{align}
\widehat p(y\mid x)
=
\frac{1}{S}
\sum_{r=1}^{S}
p(y\mid x,\theta^{(r)}),
\qquad
\theta^{(r)}\sim q_{\mathrm{TRL}}^{(T)}.
\label{eq:trl-predictive-estimator}
\end{align}
Equation~\eqref{eq:trl-predictive-estimator} is the Monte Carlo estimator used after discretization. Thus the passage from the continuous object in Eq.~\eqref{eq:continuous-trl-appendix} to the released TRL sampler involves three explicit approximations: the longitudinal mixing measure $\rho(ds)$ is replaced by the empirical uniform measure over the $T$ discrete spine anchors; the normal exponential map is replaced by the local affine chart $\psi_t$; and the position-dependent transverse covariance is represented by transported frames and amortized diagonal inverse-square-root precision factors rather than freshly recomputed dense local Cholesky factors. The scalar $\beta_\perp$ enters at this discrete sampling stage as a validation-selected transverse temperature, not as an additional invariant of the continuous idealization. Finally, posterior prediction is estimated with $S$ Monte Carlo draws from this discrete tube mixture. $T$ controls the number of stored anchors and, at fixed $\Delta s$, the nominal longitudinal extent; $S$ controls the Monte Carlo accuracy of the predictive estimate.

\subsection{Calibration and probabilistic metrics}
\label{app:calibration_metrics}
The experiments report accuracy, negative log likelihood (NLL), expected calibration error (ECE), Brier score, and, for OOD detection, predictive-entropy AUROC. Accuracy measures only the most likely class. NLL evaluates the probability assigned to the true label and strongly penalizes confident mistakes. The Brier score is the mean squared error between the predictive probability vector and the one-hot label, so it measures both calibration and sharpness of the predictive distribution.

ECE summarizes whether top-label confidence agrees with empirical top-label accuracy. Unless a table states otherwise, we use 15 equal-width confidence bins; within each bin, top-label ECE compares average maximum-class probability with average correctness and weights the absolute gap by bin frequency. ECE is easy to interpret but depends on binning and on whether calibration is summarized top-label-wise or classwise, so it should be read together with NLL and Brier score. Appendix~\ref{app:calibration_sanity} reports equal-width, equal-mass, classwise, and Kolmogorov--Smirnov sensitivity checks and limits the headline claim to top-label calibration. OOD AUROC is different: it measures how well predictive entropy ranks in-distribution and out-of-distribution examples. A method can have strong OOD ranking but poor in-distribution calibration, or vice versa, which is why the tables report both probabilistic metrics and predictive-entropy AUROC when applicable. Unless explicitly marked as diagnostic experiments, tables report mean $\pm$ standard deviation over independent seeds. Tables generally use three decimal places; additional digits are retained where a study's reported comparisons require them, for example to prevent a nonzero standard deviation or a diagnostically relevant small difference from rounding to zero. Unless a table caption states otherwise, boldface marks the best unrounded mean among the rows listed in that table; values that tie after display rounding need not all be bold. Some tables intentionally use no boldface; absence of the mark is not a claim of ties. Numerical differences, percentages, and ratios stated in prose are likewise computed from the unrounded aggregates unless explicitly identified as display-level arithmetic. Where a table omits a method class, or where the surrounding text makes a narrower claim, the mark carries no scope beyond the rows shown.

\input{appendix_geometry_idealization}

\section{Implementation and Reproducibility Details}
\label{app:implementation}

This appendix specifies the experimental protocol and the practical implementation used in the consolidated ResNet experiments: concrete data splits, hyperparameter choices, baseline protocols, and BatchNorm recalibration procedure.

\subsection{Dataset splits and validation protocol}
CIFAR-100 experiments use ResNet-18-CIFAR with the standard CIFAR stem. We split the original training set into 45,000 training images and 5,000 held-out validation images; the standard 10,000-image test set is used only for final reporting. OOD AUROC uses SVHN as the out-of-distribution set for CIFAR-100 evaluations unless otherwise noted. Hyperparameters are selected exclusively on the clean validation split by minimizing validation NLL. In the consolidated CIFAR-100 experiments, all five seeds selected $\beta_\perp=4.0$ from the grid $\{2,3,4,6,10,20\}$. The same clean-validation-selected value is then reused for CIFAR-100-C; no corrupted validation data, corrupted test images, or test labels are used for tuning. Consolidated clean CIFAR-100 results aggregate five independently trained MAP checkpoints. CIFAR-100-C likewise aggregates three independently trained, seed-matched MAP checkpoints, whereas most fixed-MAP ablations use three posterior-sampling replicates on one checkpoint; the low-loss and functional-disagreement spine diagnostics use five independently trained checkpoint-specific runs, and the small-data fine-tuning study reports ten seeds. This protocol separates model selection from both clean test evaluation and corruption robustness evaluation. ``Single-checkpoint post-hoc'' means that TRL starts from one deterministically trained checkpoint, does not continue the original task training, and does not train additional independent backbones; it does not mean data-free post-processing. The construction uses training inputs for stochastic HVPs and, for BatchNorm networks, FixBN, while the clean held-out validation split is used for tube-scale and prior-boost selection.

\subsection{Practical TRL algorithm}
\label{app:practical_algorithm}
The practical sampler implements the discrete mixture of Appendix~\ref{app:continuous_to_discrete_trl}. From a trained MAP checkpoint it estimates a leading positive-curvature rank-$k_\perp$ basis with stochastic cross-entropy HVPs, constructs a predictor--corrector spine in the complementary low-curvature region, transports and re-orthogonalizes the transverse basis, and samples a discrete anchor followed by a low-rank Gaussian perturbation. We treat this stochastic-HVP rank-$k_\perp$ basis as a scalable subsampled curvature proxy; it is not an exact full-data eigenspace: mini-batch quadratics can systematically distort extreme curvature directions relative to the full-data quadratic \citep{tatzel2025debiasing}. A seed-0 exact-GGN selector replacement changed downstream Acc/NLL/ECE/Brier by only $-0.0014/-0.0025/+0.0008/-0.0009$, supporting predictive robustness without establishing eigenspace equivalence.

The main ResNet-18-CIFAR-100 setting uses $k_\perp=30$, $T=40$ anchors, $\Delta s=0.01$, $S=25$, $\beta_\perp=4.0$, and 25 FixBN batches. Appendix~\ref{app:ablations} varies rank, anchor count, tube scale, and FixBN budget.

\subsection{FixBN protocol}
\label{app:fixbn_protocol}
For architectures with BatchNorm, sampled weights can shift internal activation statistics. The reported CIFAR TRL tables use 25 FixBN mini-batches per posterior sample in \emph{rolling} mode, in which running statistics are carried from one posterior draw into the next. The main ResNet-18 and WideResNet-16-4 SWAG-Diag rows use 20 posterior samples and 20 rolling-mode FixBN batches per sample. The VGG-11-BN SWAG-Diag row uses 25 posterior samples and 20 rolling-mode FixBN batches per sample. All reported SWAG-Diag runs are initialized from their seed-matched MAP checkpoints. In rolling mode, BatchNorm running-statistics buffers are not reset between sampled weight states. The validation sweep and subsequent final evaluation are therefore stateful: the running-statistics state reached during the selected sweep can carry into final prediction, creating the sweep-to-final dependency referenced in the Reproducibility Statement. MC~Dropout recalibrates BatchNorm once before each predictive evaluation call and then performs its stochastic dropout passes without per-draw recalibration. In contrast, TRL and SWAG-Diag perform FixBN for each sampled weight state. The batch-count ablations in Appendix~\ref{app:ablations} use rolling mode and are labeled accordingly.

The released code defaults to independent \emph{reset} recalibration. For every posterior draw, it calls \texttt{reset\_running\_stats()}, uses cumulative averaging with \texttt{momentum=None} during the recalibration passes, and restores the original momentum afterward. The ImageNet/ResNet-50 scale-check uses this reset protocol independently for every posterior sample; the reported CIFAR tables and their batch-count ablations use the rolling protocol described above. Architectures without running normalization statistics, including LayerNorm-based Vision Transformers, do not require FixBN; this is an architectural consequence, not an unreported ViT experiment. Inputs are non-corrupted training inputs drawn from the training pipeline used for BN recalibration, including its standard preprocessing and augmentation. FixBN does not use labels, update weights, or access test/CIFAR-100-C images. It is an inference-time consistency correction after a weight sample has been generated; it does not define posterior support.

A paired seed-0 rolling-versus-reset audit found only small downstream changes for both TRL and SWAG-Diag and did not trigger a multi-seed rerun. A separate seed-0 end-to-end reconstruction matched the reported metric regime, whereas replay from the stored spine alone did not reproduce the stateful rolling-FixBN realization bitwise. These are implementation/provenance audits, not benchmark replicates; exact audit rows are retained in the release documentation.

\subsection{Hyperparameter search spaces and selected values}
\label{sec:b4}
For CIFAR-100 we use $k_\perp=30$, $T=40$, $\beta_\perp=4.0$, $S=25$, and 25 FixBN batches; $\beta_\perp$ is selected from $\{2,3,4,6,10,20\}$ by held-out clean-validation NLL. The two-block isotropic prior assigns $\lambda_{\rm base}$ to the head and $\max(c\lambda_{\rm base},5)$ to the backbone. We treat this as pragmatic regularization of the approximate posterior, not as a claim that the weight prior intrinsically splits into head and backbone components. The single boost coefficient $c$ is likewise validation-selected; $c=50$ is transferred unchanged to the architecture checks. Table~\ref{tab:boost} ablates $c$, showing the no-boost collapse and that $c$ is not interchangeable with $\beta_\perp$.

\subsection{Baseline configurations}
\label{app:baseline_configurations}
This subsection distinguishes the baseline protocols used in the large-network CIFAR experiments from the full-network toy diagnostics. The method names are shared across settings, but the curvature approximation differs with scale.

The ResNet-18 CIFAR-100 benchmark defines the baseline protocol used throughout the main large-network experiments. MAP, ELA, LLA, SWAG-Diag, and TRL are attached to the same deterministic MAP checkpoints when applicable, whereas MC Dropout and Deep Ensembles require their own training-time stochastic or multi-seed training protocols. The WideResNet-16-4 and VGG-11-BN checks reuse these same baseline roles with architecture-specific checkpoints and validation-selected TRL tube scales; the VGG-11-BN SWAG-Diag row uses 25 posterior samples, compared with 20 in the ResNet-18 protocol, as documented above. The CIFAR-100$\rightarrow$CIFAR-10 fine-tuning experiment is a separate diagnostic of spine behavior under task adaptation and is not part of the baseline suite.

\paragraph{Main ResNet-18 CIFAR-100 benchmark.}
The consolidated CIFAR-100 benchmark uses ResNet-18-CIFAR with the standard CIFAR stem. MAP is the deterministic model trained for 50 epochs with SGD and cosine scheduling. ELA and LLA use last-layer Laplace approximations with Kronecker-factored curvature and marginal-likelihood prior precision \citep{daxberger2021laplace,immer2021scalable}. The reported last-layer ELA/LLA fits use the same seeded 5,000-example subset used for the marginal-likelihood fit supplying $\lambda_{\rm base}$; their curvature is accumulated on that subset without rescaling to the 45,000-example training set. These are the scalable ELA/LLA baselines reported in the main CIFAR tables: the posterior uncertainty is restricted to the classifier layer, and the difference between ELA and LLA is the predictive approximation used at test time. Deep Ensembles use $M=5$ independently trained models and average their predictions \citep{lakshminarayanan2017simple}. MC Dropout trains a dropout-modified ResNet-18 with dropout probability $p=0.2$ and uses $S=25$ stochastic dropout forward passes at inference \citep{gal2016dropout}. The evaluated SWAG-Diag baseline is initialized from the MAP checkpoint and collects only the running per-parameter mean and second moment along a short additional fine-tuning trajectory; it does not store the low-rank snapshot-deviation matrix of full SWAG \citep{maddox2019simple}. The published row uses 20 posterior samples and 20 rolling-mode FixBN batches per sample. A separate 20-versus-25 posterior-sample sensitivity check, holding FixBN fixed at 20 batches per draw, changed accuracy, NLL, and ECE by at most $0.0003$, $0.000201$, and $0.000326$, respectively, across seeds 0--2; it does not replace the published 20/20 protocol.

\paragraph{Rank-matched all-weight Faller--Martin comparison.}
FM-LR-LLA ValScale is a checkpoint-matched integration of the predictive-covariance projector of \citet{faller2026lowrank}. For each of the five ResNet-18 checkpoints, we fit an all-weight Kronecker Laplace reference on 5,000 training examples with Laplace batch size 1, excluding BatchNorm affine parameters as in the reference configuration. We build the rank-$30$ projector from 256 examples and use 25 Monte Carlo samples from the induced linearized Gaussian logits. A scalar multiplier $\alpha_{\rm FM}\in\{0,0.01,0.03,0.1,0.3,1\}$ rescales the projected covariance and is selected independently for each seed by clean-validation NLL; test and SVHN data are not used for selection. The $\alpha_{\rm FM}=0$ sanity check reproduces the MAP probabilities exactly in every seed. Because covariance tempering and Monte Carlo linearized prediction strengthen the predictive baseline relative to the native unscaled/probit evaluation, we label the row ValScale to distinguish it from an unmodified reproduction. Appendix~\ref{app:faller_martin_comparison} reports the per-seed results and projector diagnostics.

The random-subspace control of Appendix~\ref{app:trl_vs_ela_support} replaces the HVP-selected transverse basis $N_0$ by a random rank-$30$ orthonormal basis (i.i.d.\ Gaussian draws orthonormalized by two modified Gram--Schmidt reorthogonalization passes in parameter space), keeping the block-isotropic prior, the validation grid for $\beta_\perp$, $S=25$, and the FixBN protocol unchanged; we evaluate five independent bases on one MAP checkpoint and one basis on each of two further checkpoints.

\paragraph{Architecture-sensitivity checks.}
The WideResNet-16-4 and VGG-11-BN checks reuse the same CIFAR-100 data protocol, evaluation metrics, baseline roles, and clean-validation model selection rule, while retaining the stochastic budgets used by the architecture-specific runners. In particular, WideResNet-16-4 SWAG-Diag uses 20 posterior samples and 20 rolling-mode FixBN batches per sample, whereas the VGG-11-BN SWAG-Diag row uses 25 posterior samples and 20 rolling-mode FixBN batches per sample. They are intended to test whether the calibration behavior of TRL transfers across architectures, not to introduce new baseline definitions. Accordingly, MAP, SWAG-Diag, MC Dropout with $p=0.2$, ELA, LLA, and TRL are implemented with the same roles as in the ResNet-18 benchmark, with architecture-specific checkpoint directories and classifier heads. The ELA/LLA baselines remain scalable last-layer Kronecker-factored Laplace baselines in these CIFAR-scale checks. Deep Ensembles are omitted from the supplementary WideResNet and VGG tables to focus on post-hoc and single-training methods under architecture changes.

\paragraph{Toy full-network diagnostics.}
The toy sine-regression and two-moons models are small enough that all-parameter, full-matrix GGN Laplace baselines are feasible. The released toy code uses \texttt{subset\_of\_weights="all"} and \texttt{hessian\_structure="full"}; with the pinned \texttt{laplace-torch==0.2.2.2} default \texttt{CurvlinopsGGN} backend, ``full'' denotes a dense all-parameter curvature representation, not the exact Hessian of the training loss. In those tables, ELA and LLA use this dense GGN curvature, but their predictive construction, support rank and covariance scale differ from TRL; this is not a single-factor geometry comparison. The CIFAR-scale ELA/LLA baselines use last-layer curvature. The toy results probe mechanism; the CIFAR-scale experiments carry the main evidence for scalable performance.

\subsection{Runtime and memory measurement}
\label{app:runtime_measurement}
\label{app:complexity}
Runtime in Table~\ref{tab:app_runtime_memory} is the sum of recorded timing keys, so scopes differ and nested keys can overlap. The MAP, ELA/LLA, and TRL rows come from the same five core executions; in each, the shared MAP stage took the training branch, so the $504.6\pm7.2$~s MAP time is already included in ELA/LLA and TRL. TRL additionally logs spine construction, the six-point validation sweep over $\beta_\perp$, and in-distribution/OOD posterior prediction. It excludes the preceding last-layer Laplace fit used to obtain $\lambda_{\mathrm{base}}$ and the separate validation sweep used to select $c=50$. The ELA/LLA total exceeds MAP by $93.7$~s, and that increment also contains prior optimization and four ELA/LLA prediction passes; hence the omitted $\lambda_{\mathrm{base}}$ fit alone is below $93.7$~s, or less than $4.8\%$ of the reported TRL total. FixBN is also logged both inside the enclosing prediction timers and as nested diagnostic sub-timers, inflating the summed TRL value. Thus $1985.8$~s is an instrumented logged total with overlapping scopes, not a non-overlapping wall-clock measurement, and MAP must not be added to it again.

The Deep Ensemble row comes from separate recorded executions in which MAP was loaded and all five ensemble members were trained and evaluated; its $2444.1$~s total therefore contains five member trainings, not a sixth MAP training. SWAG-Diag likewise loaded existing MAP checkpoints, explaining why its logged time can be below the standalone MAP-training row. MC~Dropout mixes checkpoint hits and misses, which produces its large runtime variance. Peak VRAM is the maximum of the instrumented stages when available. These implementation- and hardware-dependent totals characterize the observed pipelines and do not establish an absolute wall-clock ranking. The asymptotic comparison to ODE-based Riemannian Laplace samplers is given in Appendix~\ref{app:complexity_discussion}.

\section{Toy Experiments with Full-Network Laplace}
\label{app:toy}

The toys use tractable dense all-parameter GGN curvature. ELA evaluates the nonlinear network at Gaussian weight draws; LLA propagates the same Gaussian through a MAP-linearized model; TRL restricts transverse sampling to rank 30 and adds discrete spine anchors. Differences in rank, scale and predictive construction preclude a single-factor interpretation of cross-method rankings.

\subsection{Noisy sine regression}
\label{app:toy_sine}
For sine regression, $80$ training inputs are equally spaced over $[-3,3]$, and the fixed $200$-point evaluation grid spans $[-6,6]$. Training and evaluation targets use independent observation noise,
\begin{equation}
y_{\rm tr}=\sin(x_{\rm tr})+\epsilon_{\rm tr},\qquad y_{\rm te}^{\rm obs}=\sin(x_{\rm te})+\epsilon_{\rm te},\qquad
\epsilon_{\rm tr},\epsilon_{\rm te}\overset{\rm ind}{\sim}\mathcal N(0,0.15^2).
\label{eq:toy-sine-data}
\end{equation}
RMSE uses latent $\sin(x_{\rm te})$; NLL, CRPS and predictive coverage use observed $y_{\rm te}^{\rm obs}$. All methods retain the known observation-noise standard deviation $\sigma_{\rm obs}=0.15$, including MAP. Prior precision is fitted independently per training seed by marginal likelihood. We score the predictive Gaussian-observation mixture directly:
\begin{equation}
\mathrm{NLL}_{\rm mix}=-\frac{1}{N}\sum_{i=1}^{N}\log\!\left[\frac{1}{S}\sum_{r=1}^{S}
\mathcal N\!\left(y_i^{\rm obs};f_{\theta^{(r)}}(x_i),0.15^2\right)\right],\qquad N=S=200.
\label{eq:toy-sine-nll}
\end{equation}
The log sum is evaluated stably. Central 68\%/95\% intervals use quantiles of this mixture, obtained by numerical CDF inversion, rather than mean-plus-standard-deviation bands. CRPS is computed by the Gaussian-mixture pairwise formula. The original moment-matched Gaussian score is distinct: for the original TRL-full configuration it is $0.242509$, versus mixture NLL $0.312885$. Moment matching is no longer used as the predictive likelihood score in the revised comparison.

The original longitudinal extent was selected using the moment-matched score: a 30-anchor candidate path with $\Delta s=0.04$ was compared at global prefixes $10$, $15$ and $20$ on development seeds $50$--$69$. The selected 15-anchor prefix, rank 30 and original $\beta_\perp=0.005$ were fixed before reporting seeds $100$--$199$ were evaluated. The replay in Table~\ref{tab:toy_sine} changes scoring, not these fitted models or sampling streams. All 500 original method/seed score checks reproduced the old moment-matched results to maximum absolute discrepancy $8.9\times10^{-7}$.

At the original scales, TRL-full has finite-mixture NLL $0.3129$, versus closed-form Gaussian NLL $0.3914$ for unit-scale LLA. Against this analytic LLA reference, the paired difference is $-0.0785$ (95\% t interval $[-0.1441,-0.0129]$), with 67/100 seeds favoring TRL. The 200-sample LLA estimate is $0.4052$; its difference from $0.3914$ reflects finite-sample scoring, not a deterioration of the fitted LLA model (Appendix~\ref{app:toy_scale_controls}). TRL retains finite-sample mixture scoring. After independent scale adjustment, tuned LLA is better. RMSE and calibration answer different questions; original TRL-full covers 60.5\%/85.9\% at nominal 68\%/95\%, indicating undercoverage at both levels.

\begin{table}[H]
\centering
\caption{Sine replay at original scales: mean $\pm$ sample SD over 100 reporting seeds. NLL scores the 200-sample mixture in Eq.~\eqref{eq:toy-sine-nll}; unit-scale LLA has closed-form Gaussian NLL $0.3914\pm0.0781$, versus its sampled $0.405\pm0.090$ shown below. Coverage uses mixture quantiles; observation noise stays at 0.15. Functional variance excludes that noise. Scale controls are in Table~\ref{tab:additional_sine_scales}.}
\label{tab:toy_sine}
\footnotesize
\begin{adjustbox}{max width=\textwidth}
\begin{tabular}{lcccccc}
\toprule
Method & RMSE $\downarrow$ & NLL$_{\rm mix}$ $\downarrow$ & CRPS $\downarrow$ & Cov. 68\% & Cov. 95\% & Func. var. \\
\midrule
MAP & \meanstd{0.266}{0.047} & \meanstd{1.138}{0.660} & \meanstd{0.172}{0.022} & \meanstd{0.478}{0.050} & \meanstd{0.741}{0.052} & \meanstd{0.0000}{0.0000} \\
ELA, unit scale & \meanstd{1.269}{0.351} & \meanstd{4.013}{0.197} & \meanstd{3.739}{0.182} & \meanstd{1.000}{0.000} & \meanstd{1.000}{0.000} & \meanstd{246.6581}{22.0101} \\
LLA, unit scale & \meanstd{0.281}{0.070} & \meanstd{0.405}{0.090} & \meanstd{0.255}{0.033} & \meanstd{0.822}{0.037} & \meanstd{0.974}{0.012} & \meanstd{1.7025}{0.6456} \\
TRL, original scale & \meanstd{0.271}{0.038} & \meanstd{0.313}{0.318} & \meanstd{0.169}{0.021} & \meanstd{0.605}{0.092} & \meanstd{0.859}{0.077} & \meanstd{0.0373}{0.0249} \\
\bottomrule
\end{tabular}
\end{adjustbox}
\end{table}

\begin{figure}[t]
\centering
\includegraphics[width=0.98\textwidth]{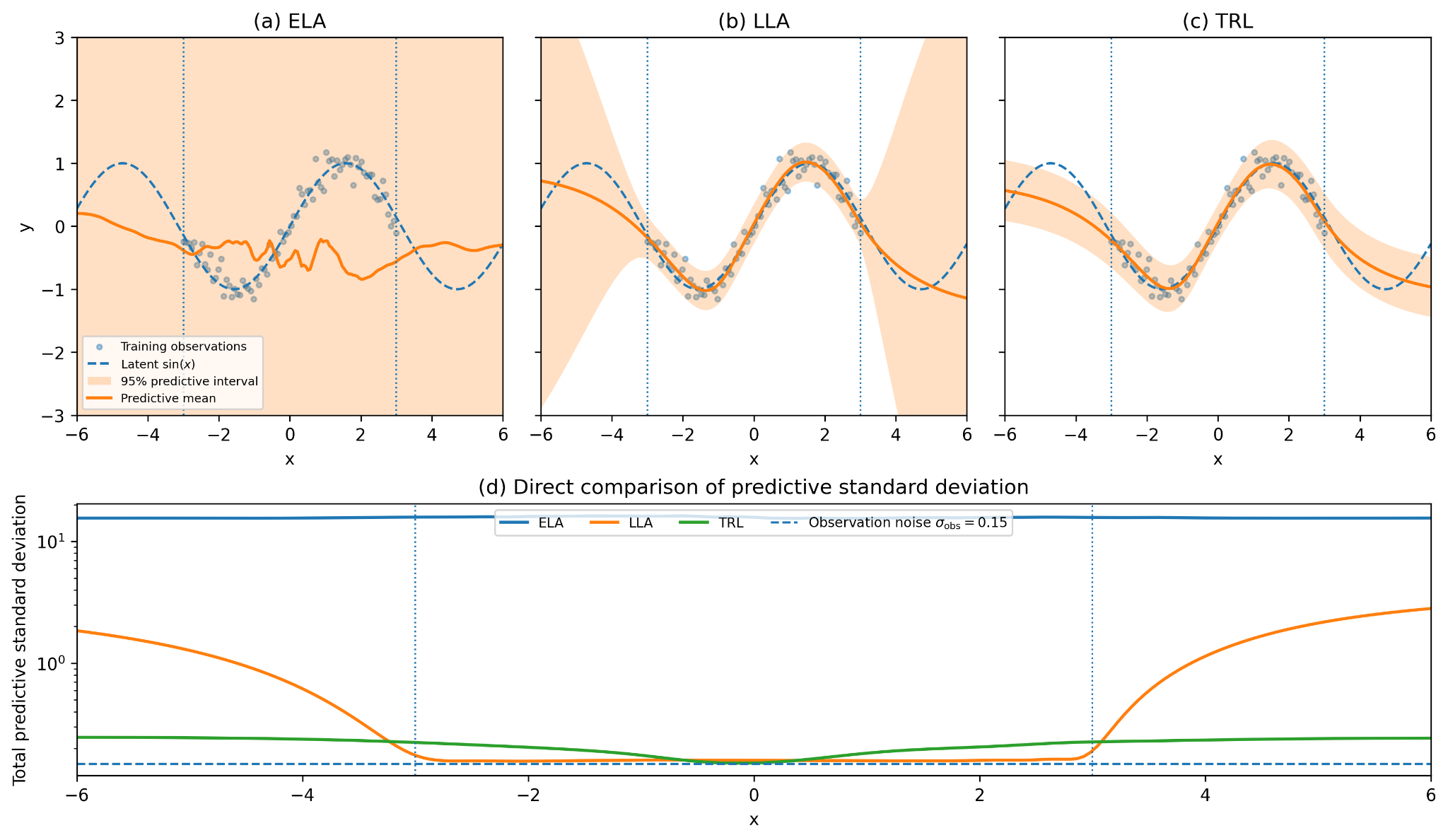}
\caption{\textbf{Original sine configurations, with moment-matched display bands.} Top: predictive means and moment-based display bands; these are not mixture-quantile intervals. The representative seed was selected by the original median moment-matched TRL-NLL rule, not by the new mixture score. Bottom: total predictive standard deviation; the horizontal line is the observation-noise floor. Vertical lines delimit training. The 15-anchor prefix is fixed. Scale-controlled conclusions rely on Table~\ref{tab:additional_sine_scales}.}
\label{fig:toy_sine}
\end{figure}
Figure~\ref{fig:toy_sine} describes the original scales: ELA is broadly over-dispersed and LLA has large extrapolation variance. It is not evidence that these baselines remain inferior after calibration. The separately selected LLA scale substantially reduces extrapolation variance and improves NLL/CRPS.

\subsection{Toy classification}
\label{app:toy_moons_setup}
Two-moons uses input-coordinate noise $0.30$, 500 training examples, 1,000 independently generated test examples, and a two-hidden-layer width-16 MLP trained for 3,000 epochs. The likelihood is Bernoulli; input noise is not an extra observation-variance term. With scalar logits $a_\theta(x)$ and logistic sigmoid $\sigma$, prediction uses
\begin{equation}
p_i^{\rm MAP}=\sigma(a_{\theta_{\rm MAP}}(x_i)),\qquad
p_i=\frac1S\sum_{r=1}^{S}\sigma(a_{\theta^{(r)}}(x_i)),\qquad S=250,
\label{eq:toy-moons-probabilities}
\end{equation}
and predictive NLL is evaluated after averaging probabilities:
\begin{equation}
\mathrm{NLL}=-\frac1N\sum_{i=1}^{N}\left[y_i\log p_i+(1-y_i)\log(1-p_i)\right],\qquad N=1000.
\label{eq:toy-moons-nll}
\end{equation}
We preserve the original $10^{-6}$ probability clipping for NLL. Brier is the scalar Bernoulli score $N^{-1}\sum_i(p_i-y_i)^2$. ECE is 15-bin equal-width top-label ECE, with confidence $\max(p_i,1-p_i)$. Functional variance is across sampled scalar probabilities; it is not a monotone quality metric.

Table~\ref{tab:toy_moons} is a post-hoc extension of the original 10-seed experiment to 100 new reporting seeds $100$--$199$. All 10 historical seeds were replayed exactly: the maximum discrepancy across 250 method/seed/metric checks was zero. Scales were selected only on development seeds $50$--$69$, with the same grid as sine and predictive NLL as criterion. ELA/LLA select standard-deviation multipliers $0.02/0.05$; TRL-single/full select transverse $\beta_\perp=0.10/0.05$. ECE is reported at these NLL-selected settings, not tuned separately.

ELA/LLA NLL falls from $0.7038/0.4336$ at unit scale to $0.2296/0.2292$, near MAP ($0.2300$). Tuned TRL-single/full gives $0.2282/0.2287$. The paired full-minus-tuned-single NLL difference is $+0.000486$ (95\% interval $[-0.000066,0.001038]$). Tuned LLA minus full is $+0.000487$ ($[-0.000102,0.001075]$); ECE differs by $+0.000032$ ($[-0.000786,0.000850]$), with LLA winning 50/100 ECE comparisons. These intervals do not establish equivalence, but show no clear full-spine advantage over the tuned alternatives in this experiment. Tuned single has slightly lower NLL than tuned LLA: single minus LLA $-0.000973$ ($[-0.001492,-0.000454]$). The original baseline gap is strongly scale-dependent.

\begin{table}[H]
\centering
\caption{Two-moons: mean $\pm$ sample SD over 100 new paired reporting seeds, 250 predictive samples. Original and development-tuned scales are shown together. TRL-full retains its original $\beta=0.05$ after tuning. Historical 10-seed rows are preserved in the ancillary archive and are not pooled with these 100 seeds.}
\label{tab:toy_moons}
\footnotesize
\begin{adjustbox}{max width=\textwidth}
\begin{tabular}{lccccc}
\toprule
Method & Acc. $\uparrow$ & NLL $\downarrow$ & ECE $\downarrow$ & Brier $\downarrow$ & Func. var. \\
\midrule
MAP & \meanstd{0.9098}{0.0086} & \meanstd{0.2300}{0.0196} & \meanstd{0.0220}{0.0068} & \meanstd{0.0671}{0.0056} & \meanstd{0.00000}{0.00000} \\
ELA, unit scale & \meanstd{0.4572}{0.1295} & \meanstd{0.7038}{0.0258} & \meanstd{0.1466}{0.0671} & \meanstd{0.2553}{0.0128} & \meanstd{0.20115}{0.00442} \\
LLA, unit scale & \meanstd{0.9048}{0.0131} & \meanstd{0.4336}{0.0301} & \meanstd{0.2317}{0.0218} & \meanstd{0.1303}{0.0125} & \meanstd{0.17008}{0.01346} \\
TRL-single, $\beta=0.05$ & \meanstd{0.9100}{0.0086} & \meanstd{0.2292}{0.0190} & \meanstd{0.0211}{0.0065} & \meanstd{0.0670}{0.0055} & \meanstd{0.00099}{0.00055} \\
TRL-full, $\beta=0.05$ & \meanstd{0.9100}{0.0086} & \meanstd{0.2287}{0.0181} & \meanstd{0.0207}{0.0064} & \meanstd{0.0670}{0.0054} & \meanstd{0.00122}{0.00073} \\
ELA, $\alpha=0.02$ & \meanstd{0.9098}{0.0086} & \meanstd{0.2296}{0.0193} & \meanstd{0.0217}{0.0063} & \meanstd{0.0670}{0.0055} & \meanstd{0.00030}{0.00013} \\
LLA, $\alpha=0.05$ & \meanstd{0.9099}{0.0088} & \meanstd{0.2292}{0.0189} & \meanstd{0.0208}{0.0065} & \meanstd{0.0670}{0.0055} & \meanstd{0.00184}{0.00081} \\
TRL-single, $\beta=0.10$ & \meanstd{0.9099}{0.0087} & \meanstd{0.2282}{0.0176} & \meanstd{0.0205}{0.0067} & \meanstd{0.0669}{0.0053} & \meanstd{0.00399}{0.00220} \\
\bottomrule
\end{tabular}
\end{adjustbox}
\end{table}

\begin{figure}[H]
\centering
\includegraphics[width=0.95\textwidth]{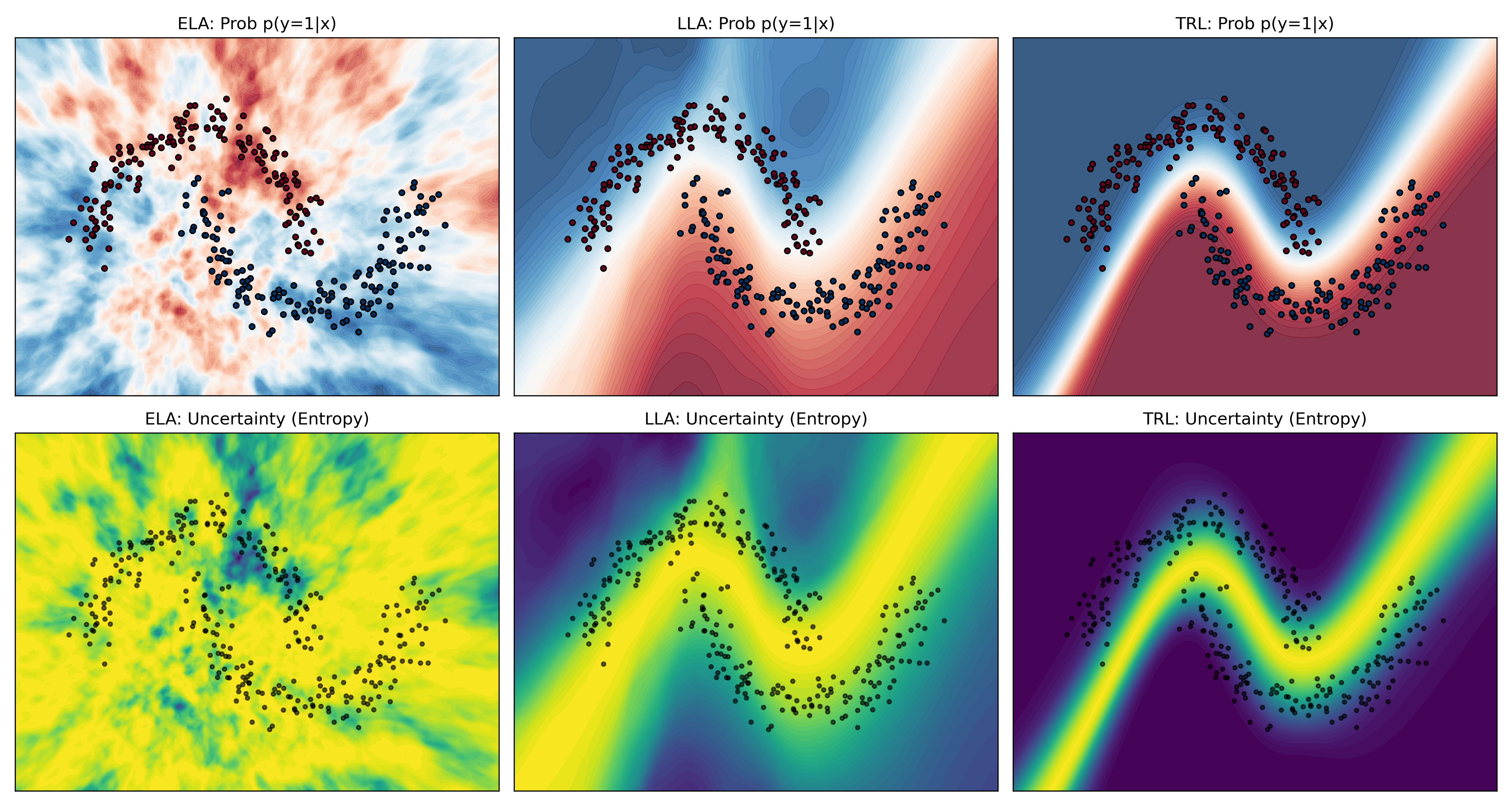}
\caption{Original, untuned two-moons configurations: predictive probability and entropy maps for ELA, LLA and TRL. This retained illustration is not a scale-controlled comparison; Table~\ref{tab:toy_moons} reports the new 100-seed scale controls.}
\label{fig:toy_moons}
\end{figure}
The likelihood enters regression and classification differently. Gaussian regression mixes observation densities; moment matching can alter its likelihood score. Bernoulli classification mixes probabilities before taking the log. Neither scoring distinction establishes a geometric advantage over independently calibrated baselines.

\subsection{Toy hyperparameters and selection boundary}
\label{app:toy_hyperparameters}
Both toys fit dense all-parameter GGN Laplace with a per-seed marginal-likelihood prior. Sine uses known observation noise 0.15, 600 training epochs, a 30-anchor candidate path with step 0.04 and correction 0.10, then the frozen 15-anchor prefix; the original transverse scale is 0.005. Two-moons uses 3,000 training epochs, batch size 128, 50 anchors, step 0.08, correction 0.10 and original scale 0.05. Both use rank 30: 30 of 2,701 sine parameters and 30 of 337 classification parameters. The 50-anchor classification spine is retained without a length search. New scale controls leave rank and path fixed; matching their scale-selection protocol does not match the baseline families or isolate tubular geometry. Appendix~\ref{app:toy_scale_controls} gives the complete scale grid and the distinction between previously examined sine seeds and new classification seeds.

With 80 scalar observations, the sine GGN has rank at most 80. Under the isotropic prior, this gives a minimum-precision eigenspace of dimension at least 2,621 in exact arithmetic, so the smallest-eigenvalue prescription does not uniquely determine the initial tangent. The toy eigensolver and the seeded random-complement rule used at CIFAR scale select directions differently; no equivalence between their induced distributions is assumed.

\paragraph{Audit of the released sine corrector.}
\label{app:toy_corrector_audit}
The released implementation stores the current anchor, takes a predictor step of $0.04$, and applies one projected MSE-gradient step with rate $0.10$. It has no loss-acceptance test or backtracking. A read-only audit of the same 100 saved 15-anchor prefixes found an anchor exceeding $1.5$ times the MAP training MSE in 99 tasks; the largest per-task ratio was $10.86$. The realized prefix length averaged $2.42$ times the nominal $14\times0.04$. The saved displacements also exhibit oscillation along a stiff direction; this is consistent with numerical amplification contributing to diversity, without identifying a causal fraction of the predictive gain. These trajectories therefore do not demonstrate preservation of near-MAP loss or geodesic behavior. The finite predictive mixtures remain normalized, and their reported scores are unchanged. They should be interpreted separately from the low-loss construction motivation.

\subsection{Isolating longitudinal support}
\label{app:toy_spine_isolation}
Table~\ref{tab:toy_spine_isolation} compares the original MAP-centered and full-spine samplers at the same transverse scale within each task. Sine uses the frozen 15-anchor prefix; two-moons retains all 50 anchors. Mixture scoring lowers sine NLL from $1.129$ to $0.313$; full minus single is $-0.8164$ (95\% interval $[-0.9449,-0.6880]$), favoring full in all 100 seeds. Against the separately scale-tuned single ($0.5548$), original full still wins 84/100 seeds, with difference $-0.2419$ ($[-0.2943,-0.1896]$). These 84/100 wins compare original-scale full with the tuned MAP-centered tube.

A further post-hoc control jointly selects a global anchor index and its own transverse scale on development seeds $50$--$69$, then fixes both across the 100 reporting seeds (Appendix~\ref{app:toy_scale_controls}; Table~\ref{tab:additional_sine_scales}). The unrestricted search selects the MAP ($t=0$, $\beta=0.1$). Restricting the same development grid to $t>0$ selects $t=2$, $\beta=0.1$, giving NLL $0.5517$; the endpoint $t=14$ independently selects $\beta=0.5$ and gives $1.2849$. Scale-tuned full ($\beta=0$, NLL $0.3126$) minus displaced-single NLL is $-0.2390$ (95\% paired interval $[-0.2903,-0.1878]$), favoring full in 86/100 seeds. Original-scale full also wins 86/100, with difference $-0.2388$ ($[-0.2899,-0.1876]$). Mixing thus improves over these globally development-selected single-center alternatives; the experiment does not test per-task anchor selection. The full-spine boundary $\beta=0$ retains the gain without transverse perturbations. Tuned LLA remains better on sine, and these comparisons do not establish posterior fidelity.

At original two-moons scale 0.05, full minus single NLL is $-0.000504$ ($[-0.001075,0.000067]$), with 49/100 seeds favoring full; after separate tuning, full is also not clearly better than single. No additional longitudinal score gain is established in this toy. The spine's incremental value is regime-dependent.

\begin{table}[H]
\centering
\caption{Original-scale spine comparison over 100 reporting seeds in each task. Sine NLL is the finite mixture score; two-moons rows are the new post-hoc 100-seed extension. Scales match within each pair (sine 0.005, classification 0.05); the path is 15 or 50 anchors, respectively. Functional variance excludes observation noise.}
\label{tab:toy_spine_isolation}
\footnotesize
\begin{adjustbox}{max width=\textwidth}
\begin{tabular}{llccccc}
\toprule
Task & Method & NLL $\downarrow$ & RMSE $\downarrow$ & Acc. $\uparrow$ & Brier $\downarrow$ & Func. var. \\
\midrule
Sine & TRL-single & \meanstd{1.129}{0.651} & \meanstd{0.266}{0.047} & -- & -- & \meanstd{0.00003}{0.00001} \\
Sine & TRL-full & \meanstd{0.313}{0.318} & \meanstd{0.271}{0.038} & -- & -- & \meanstd{0.03727}{0.02490} \\
Two-moons & TRL-single & \meanstd{0.22923}{0.01896} & -- & \meanstd{0.9100}{0.0086} & \meanstd{0.06698}{0.00548} & \meanstd{0.00099}{0.00055} \\
Two-moons & TRL-full & \meanstd{0.22872}{0.01805} & -- & \meanstd{0.9100}{0.0086} & \meanstd{0.06701}{0.00542} & \meanstd{0.00122}{0.00073} \\
\bottomrule
\end{tabular}
\end{adjustbox}
\end{table}
The revised toy conclusions distinguish within-TRL support changes from cross-method ranking. Original-scale TRL is better than the original ELA/LLA configurations, but tuned LLA is better on sine and tuned local alternatives nearly match TRL-full on classification. These experiments do not establish general superiority over calibrated ELA/LLA or single-factor evidence for tubular geometry.

\section{Full CIFAR-100 Results}
\label{app:cifar100_full}

Table~\ref{tab:cifar100_clean_5seed} in the main text is the canonical five-checkpoint clean comparison. This appendix adds the matched all-weight low-rank, calibration-estimator, architecture-sensitivity, and runtime diagnostics needed to interpret that table.

\subsection{Rank-matched all-weight Faller--Martin comparison}
\label{app:faller_martin_comparison}

The comparison in Table~\ref{tab:app_faller_martin} targets the main potential confound of the last-layer Laplace rows: both FM-LR-LLA ValScale and TRL have access to the full network and use rank $30$. The Faller--Martin projector is optimized to preserve the predictive covariance of an all-weight Kronecker Laplace reference on its construction set, whereas TRL selects parameter-space support with the positive-spectrum curvature proxy and evaluates nonlinear sampled networks. Table~\ref{tab:app_faller_martin} matches checkpoint, architecture, split, and posterior rank, but not the prior family: FM uses a single scalar prior precision fitted for the Kronecker reference, while TRL uses the validation-selected two-block head/backbone precision described in Appendix~\ref{app:implementation}. It should therefore be read as a matched low-rank support-budget comparison, not as a one-factor ablation of the subspace criterion. Writing $\Psi$ for the all-weight Kronecker-Laplace weight covariance and $J_X$ for the stacked logit Jacobian on the 256-example projector set $X$, the predictive-covariance operator is $\Sigma_X=J_X\Psi J_X^\top$.

\begin{table}[H]
\centering
\caption{FM-LR-LLA ValScale on CIFAR-100. The scalar covariance multiplier $\alpha_{\rm FM}$ is selected independently per seed by clean-validation NLL from $\{0,0.01,0.03,0.1,0.3,1\}$. Predictive metrics are test-set values. Trace fraction is the sum of the leading rank-$30$ eigenvalues divided by an eight-probe Hutchinson estimate of $\operatorname{tr}(\Sigma_X)$, where $\Sigma_X=J_X\Psi J_X^\top$ is the all-weight Kronecker predictive-covariance operator on the 256-example projector set. FM uses a scalar prior precision, whereas TRL uses a two-block head/backbone prior.}
\label{tab:app_faller_martin}
\footnotesize
\begin{adjustbox}{max width=\textwidth}
\begin{tabular}{cccccccc}
\toprule
Seed & $\alpha_{\rm FM}$ & Acc $\uparrow$ & NLL $\downarrow$ & ECE $\downarrow$ & Brier $\downarrow$ & AUROC $\uparrow$ & Trace frac. \\
\midrule
0 & 0.03 & 0.7410 & 0.9913 & 0.0463 & 0.3582 & 0.8669 & 6.282\% \\
1 & 0.10 & 0.7411 & 0.9857 & 0.0235 & 0.3603 & 0.8078 & 5.806\% \\
2 & 0.03 & 0.7419 & 0.9883 & 0.0453 & 0.3620 & 0.8674 & 6.023\% \\
3 & 0.03 & 0.7463 & 0.9845 & 0.0440 & 0.3555 & 0.8267 & 5.914\% \\
4 & 0.03 & 0.7432 & 0.9744 & 0.0452 & 0.3535 & 0.8452 & 6.158\% \\
\midrule
Mean $\pm$ std. & -- & \meanstd{0.7427}{0.0022} & \meanstd{0.9848}{0.0064} & \meanstd{0.0409}{0.0098} & \meanstd{0.3579}{0.0034} & \meanstd{0.8428}{0.0259} & \meanstd{6.037\%}{0.190\%} \\
\bottomrule
\end{tabular}
\end{adjustbox}
\end{table}

Four of five seeds select $\alpha_{\rm FM}=0.03$ and one selects $0.1$; none selects the native-scale $\alpha_{\rm FM}=1$, while $\alpha_{\rm FM}=0$ exactly reproduces MAP probabilities. Thus the rank-$30$ all-weight covariance is predictively over-dispersed at native scale in this protocol; the $\alpha_{\rm FM}=0$ sanity check rules out a mean-prediction or pipeline error. The retained projector captures $6.04\%\pm0.19\%$ of the estimated predictive-covariance trace on its construction set. Relative to MAP, validation-scaled FM improves NLL/ECE/Brier; TRL remains lower by $0.0336/0.0274/0.0045$ in the unrounded five-seed means and has $0.0243$ higher mean predictive-entropy AUROC.

We report rank $30$ as a matched support budget, not as an optimum for either method. FM rank truncates a specified predictive-covariance reference, whereas TRL rank is a support-design choice and its own rank ablation is non-monotone. Because a larger FM rank with covariance scale reselected at every rank is a distinct experiment that we did not run, the captured-trace fraction should not be interpreted as predicting the outcome of such a sweep.

\subsection{Calibration sanity checks}
\label{app:calibration_sanity}

Table~\ref{tab:app_calibration_sanity} compares the headline five-checkpoint results with the current audited five-checkpoint MAP+TS control. The canonical checkpoint-matched top-label ECE comparison favors TRL in all five training seeds, with mean difference $-0.0147$ (TRL minus MAP+TS); a directly checkable sign certificate is given below. The three-seed clean MAP+TS result is retained only as the corruption-matched companion to the three-checkpoint CIFAR-100-C transfer artifact in Table~\ref{tab:app_mapts_c100c}; the five-checkpoint result is used for the clean headline comparison. The validation-NLL ELA/LLA grid-search results are reported separately below.

\begin{table}[H]
\centering
\caption{CIFAR-100 calibration sanity checks. Values are mean $\pm$ sample standard deviation over five independently trained MAP checkpoints. ECE is 15-bin equal-width top-label ECE.}
\label{tab:app_calibration_sanity}
\footnotesize
\begin{adjustbox}{max width=\textwidth}
\begin{tabular}{lccccc}
\toprule
Method & Acc $\uparrow$ & NLL $\downarrow$ & ECE $\downarrow$ & Brier $\downarrow$ & prior / temp. \\
\midrule
MAP & \meanstd{0.743}{0.003} & \meanstd{1.040}{0.008} & \meanstd{0.096}{0.002} & \meanstd{0.370}{0.003} & -- \\
MAP+TS & \meanstd{0.743}{0.003} & \meanstd{0.960}{0.006} & \meanstd{0.028}{0.004} & \meanstd{0.355}{0.003} & T=\meanstd{1.378}{0.026} \\
ELA-marglik & \meanstd{0.724}{0.002} & \meanstd{1.356}{0.013} & \meanstd{0.262}{0.008} & \meanstd{0.468}{0.004} & $\lambda\approx4.9$ \\
LLA-marglik & \meanstd{0.739}{0.002} & \meanstd{1.492}{0.017} & \meanstd{0.368}{0.008} & \meanstd{0.521}{0.006} & $\lambda\approx4.9$ \\
TRL & \meanstd{0.744}{0.003} & \meanstd{0.951}{0.006} & \meanstd{0.013}{0.002} & \meanstd{0.353}{0.003} & $\beta_\perp=4,\ c=50$ \\
\bottomrule
\end{tabular}
\end{adjustbox}
\end{table}

\noindent\textit{Aggregate-only grid-search diagnostic.}
Using the same held-out validation split and a 100-point logarithmic grid $\lambda\in[10^{-4},10^4]$, the three-seed aggregate means were ELA-grid: Acc.~$0.742$, NLL $1.043$, ECE $0.0956$, Brier $0.3715$; and LLA-grid: Acc.~$0.742$, NLL $1.042$, ECE $0.0958$, Brier $0.3716$. All three seeds selected the upper boundary $\lambda=10^4$ for both variants. Because the released artifact does not retain the corresponding per-seed metric rows, these aggregate-only values are reported in text and omitted from the uncertainty-aware rows in Table~\ref{tab:app_calibration_sanity}.

\textbf{Validation-NLL tuning of last-layer Laplace.}
Under marginal-likelihood tuning, ELA/LLA select a small prior precision ($\lambda\approx4.9$) and are strongly overdispersed, giving worse NLL and ECE than the MAP predictor itself. Re-tuning the prior precision by validation-NLL grid search--the same selection criterion used for TRL's tube scale-removes this overdispersion, but the selected value is the upper boundary of the search grid ($\lambda=10^4$) for every seed and for both predictive variants. In other words, validation-NLL tuning does not recover a stronger non-degenerate last-layer posterior; it drives the posterior toward the MAP predictor. This is why ELA-grid and LLA-grid are nearly identical despite using different predictive approximations (probit GLM versus Monte Carlo nonlinear prediction): once the posterior collapses onto the MAP point, there is negligible weight-space variance to propagate, so the link approximation has little effect. Because this sensitivity varies only the isotropic prior precision, it does not emulate refitting the likelihood curvature on all 45,000 training examples. This control therefore addresses prior-selection sensitivity within the reported subset-fit last-layer Laplace protocol while showing that, in this CIFAR-100 regime, the tested last-layer Laplace family does not provide a useful calibration improvement over MAP unless it degenerates toward the deterministic predictor.

\textbf{Temperature scaling.}
MAP temperature scaling uses one positive scalar temperature per seed, selected by NLL on the same clean validation split and frozen before test evaluation. The current five-checkpoint rerun selects $\tau=1.378\pm0.026$. Writing $a$ for the logit vector, the map $a\mapsto a/\tau$ is order-preserving for $\tau>0$ and therefore preserves the checkpoint-matched MAP argmax exactly. It reduces mean NLL from about $1.040$ to $0.960$, 15-bin equal-width top-label ECE from $0.096$ to $0.028$, and Brier score from $0.370$ to $0.355$. TRL remains lower at $0.951/0.013/0.353$; all five paired checkpoint-wise ECE differences are negative, with mean difference $-0.0147$ (TRL minus MAP+TS), while the NLL and Brier mean gaps are $0.0092$ and $0.0018$. We report temperature scaling as a calibration-only control, not as a posterior method: it returns one deterministic rescaled predictive distribution, whereas TRL returns parameter-space posterior samples and functional-variation diagnostics. MAP+TS clean predictive-entropy AUROC against SVHN is $0.870\pm0.022$.

\paragraph{Seedwise sign certificate.}
The canonical TRL top-label ECEs for seeds $0$--$4$ are $0.014815$, $0.010418$, $0.015262$, $0.013676$, and $0.013025$. For any $n=5$ observations with mean $\mu$ and sample standard deviation $s$, each observation obeys $|x_i-\mu|\leq\sqrt{n-1}\,s$. The preserved MAP+TS aggregate gives $\mu=0.028096$ and $s=0.003533$; even allowing for their six-decimal display rounding, every MAP+TS seed therefore satisfies $x_i\geq0.0210285$. This exceeds the largest canonical TRL seed ECE, $0.015262$, so all five checkpoint-matched TRL-minus-MAP+TS differences are negative. This certificate establishes the signs from the preserved aggregate and canonical TRL rows; it does not reconstruct the unpreserved per-seed MAP+TS magnitudes. The machine-readable certificate is included as \texttt{ece\_sign\_certificate.csv} in the MAP+TS provenance directory; its schema separates per-seed upper bounds from the aggregate mean difference.

\begin{table}[H]
\centering
\caption{Sensitivity of clean calibration conclusions to the estimator (mean $\pm$ sample standard deviation over three checkpoint-specific evaluations, training seeds 0--2). TRL here is the fresh-refresh mechanism diagnostic because that artifact preserves per-example probabilities; the headline five-checkpoint row remains canonical. Equal-mass and equal-width columns are top-label ECE with 15 bins. Lower is better.}
\label{tab:app_ece_estimators}
\scriptsize
\setlength{\tabcolsep}{3.5pt}
\begin{adjustbox}{max width=\textwidth}
\begin{tabular}{lcccc}
\toprule
Method & Classwise ECE & Equal-mass-15 & Equal-width-15 & KS calibration distance \\
\midrule
Deep Ensemble & \meanstd{0.001743}{0.000030} & \meanstd{0.01705}{0.00221} & \meanstd{0.01775}{0.00156} & \meanstd{0.01266}{0.00161} \\
MAP & \meanstd{0.002619}{0.000015} & \meanstd{0.09641}{0.00269} & \meanstd{0.09641}{0.00269} & \meanstd{0.09641}{0.00269} \\
MAP+TS & \meanstd{0.001921}{0.000004} & \meanstd{0.02834}{0.00450} & \meanstd{0.02946}{0.00402} & \meanstd{0.01703}{0.00678} \\
TRL fresh-refresh & \meanstd{0.001960}{0.000024} & \meanstd{0.01283}{0.00384} & \meanstd{0.01268}{0.00329} & \meanstd{0.00545}{0.00178} \\
\bottomrule
\end{tabular}
\end{adjustbox}
\end{table}

Table~\ref{tab:app_ece_estimators} supports a narrow but stable conclusion. Under both top-label binning rules and the bin-free KS distance, the fresh-refresh TRL diagnostic is lower than MAP+TS; the three paired equal-width differences (TRL minus MAP+TS) are all negative, with mean $-0.0168\pm0.0070$. The conclusion is not universal calibration dominance: classwise ECE is marginally lower for MAP+TS, and Deep Ensembles are numerically best in the classwise column. The fresh-refresh seed-0 top-label ECE differs from the canonical seed-0 value by only about $0.00045$, so the variant label is important for provenance but not material to the clean top-label conclusion. We therefore use ``top-label ECE'' in headline claims and retain NLL and Brier as the primary proper scoring rules.

\paragraph{Frozen temperature scaling under corruption.}
We additionally fit one scalar temperature per seed by minimizing NLL on the clean 5,000-example validation split, freeze it, and reuse it for the clean test set and all $19\times5$ CIFAR-100-C corruption/severity cells. No corruption data are used for calibration or model selection. MAP+TS and TRL use the same matched MAP checkpoints; the checkpoint identities were verified by SHA-256 and by exact equality of MAP and MAP+TS argmax accuracies in all 285 corruption cells. Both methods are selected by clean-validation NLL and transferred unchanged to CIFAR-100-C, although MAP+TS performs continuous optimization of one scalar whereas TRL selects its transverse scale from a discrete grid with Monte Carlo predictions. The anonymous supplementary result file contains 3 clean rows and 285 corruption rows with no missing or duplicate cells. Its clean MAP+TS row is the three-checkpoint companion to this frozen-corruption evaluation and should not be confused with the current five-checkpoint clean headline row in Tables~\ref{tab:cifar100_clean_5seed} and~\ref{tab:app_calibration_sanity}.

\begin{table}[H]
\centering
\caption{Frozen clean-validation temperature scaling for the three-checkpoint corruption transfer. CIFAR-100-C values macro-average the 95 corruption/severity cells within each of three independently trained, seed-matched checkpoints, then report mean $\pm$ sample standard deviation. The clean MAP+TS row is the corruption-matched three-checkpoint companion; the current five-checkpoint clean headline row is reported elsewhere. The clean TRL row reproduces the five-checkpoint headline result, while the CIFAR-100-C TRL row uses the three matched checkpoints.}
\label{tab:app_mapts_c100c}
\footnotesize
\begin{adjustbox}{max width=\textwidth}
\begin{tabular}{llcccc}
\toprule
Dataset & Method & Acc $\uparrow$ & NLL $\downarrow$ & ECE $\downarrow$ & Brier $\downarrow$ \\
\midrule
Clean (3-checkpoint companion) & MAP+TS & \meanstd{0.7415}{0.0014} & \meanstd{0.9635}{0.0042} & \meanstd{0.0296}{0.0042} & \meanstd{0.3570}{0.0017} \\
Clean & TRL & \meanstd{0.744}{0.003} & \meanstd{0.951}{0.006} & \meanstd{0.013}{0.002} & \meanstd{0.353}{0.003} \\
\midrule
CIFAR-100-C & MAP+TS & \meanstd{0.4711}{0.0044} & \meanstd{2.3895}{0.0411} & \meanstd{0.1180}{0.0078} & \meanstd{0.6907}{0.0070} \\
CIFAR-100-C & TRL & \meanstd{0.473}{0.004} & \meanstd{2.464}{0.047} & \meanstd{0.118}{0.006} & \meanstd{0.689}{0.005} \\
\bottomrule
\end{tabular}
\end{adjustbox}
\end{table}

\begin{table}[H]
\centering
\caption{CIFAR-100-C calibration-only control by severity. MAP+TS is macro-averaged over 19 corruptions within each of three independently trained, seed-matched checkpoints, followed by mean $\pm$ sample standard deviation. TRL values are reproduced from Table~\ref{tab:app_cifar100c_severity}.}
\label{tab:app_mapts_c100c_severity}
\footnotesize
\begin{adjustbox}{max width=\textwidth}
\begin{tabular}{llcccc}
\toprule
Severity & Method & Acc $\uparrow$ & NLL $\downarrow$ & ECE $\downarrow$ & Brier $\downarrow$ \\
\midrule
1 & MAP+TS & \meanstd{0.6210}{0.0048} & \meanstd{1.5340}{0.0332} & \meanstd{0.0570}{0.0068} & \meanstd{0.5087}{0.0071} \\
  & TRL & \meanstd{0.623}{0.005} & \meanstd{1.540}{0.030} & \meanstd{0.047}{0.005} & \meanstd{0.506}{0.005} \\
2 & MAP+TS & \meanstd{0.5330}{0.0053} & \meanstd{1.9717}{0.0439} & \meanstd{0.0836}{0.0084} & \meanstd{0.6142}{0.0080} \\
  & TRL & \meanstd{0.535}{0.004} & \meanstd{1.998}{0.041} & \meanstd{0.079}{0.007} & \meanstd{0.611}{0.006} \\
3 & MAP+TS & \meanstd{0.4774}{0.0045} & \meanstd{2.3039}{0.0434} & \meanstd{0.1094}{0.0074} & \meanstd{0.6827}{0.0067} \\
  & TRL & \meanstd{0.480}{0.004} & \meanstd{2.352}{0.050} & \meanstd{0.108}{0.005} & \meanstd{0.679}{0.005} \\
4 & MAP+TS & \meanstd{0.4104}{0.0036} & \meanstd{2.7385}{0.0438} & \meanstd{0.1437}{0.0081} & \meanstd{0.7650}{0.0068} \\
  & TRL & \meanstd{0.412}{0.003} & \meanstd{2.841}{0.059} & \meanstd{0.148}{0.003} & \meanstd{0.763}{0.003} \\
5 & MAP+TS & \meanstd{0.3136}{0.0042} & \meanstd{3.3994}{0.0482} & \meanstd{0.1964}{0.0107} & \meanstd{0.8832}{0.0089} \\
  & TRL & \meanstd{0.315}{0.004} & \meanstd{3.589}{0.066} & \meanstd{0.209}{0.009} & \meanstd{0.886}{0.009} \\
\bottomrule
\end{tabular}
\end{adjustbox}
\end{table}

The corruption NLL advantage is consistent across the three paired seeds. Defining $\Delta=\mathrm{NLL}_{\mathrm{TRL}}-\mathrm{NLL}_{\mathrm{MAP+TS}}$, the differences are $0.086415$, $0.075855$, and $0.061099$, giving $0.074456\pm0.012716$. With only three independent seeds, we treat this as consistent descriptive evidence; three independent seeds are insufficient for a strong statistical claim. The result delimits the calibration claim: scalar recalibration is a strong corruption-NLL control, but it does not define a posterior over parameters or functions, generate posterior samples, or provide the support and functional-variation diagnostics available from TRL.

\subsection{Architecture-sensitivity check: WideResNet-16-4}
\label{app:wrn_architecture}

To test whether the calibration behavior is tied to a single architecture, we also run a supplementary CIFAR-100 check with WideResNet-16-4-CIFAR. This is not intended as a full architecture-scaling study. We reuse the main TRL structural configuration and select only the tube scale on the clean validation split using the same validation-NLL criterion and grid as in the ResNet-18 experiments. The selected value is $\beta_\perp=3.0$ for all three WideResNet seeds, whereas the ResNet-18 experiments select $\beta_\perp=4.0$. This difference is expected because the effective posterior scale depends on architecture-specific curvature and normalization behavior.

\begin{table}[H]
\centering
\caption{Supplementary architecture-sensitivity check on CIFAR-100 with WideResNet-16-4-CIFAR. Values are mean $\pm$ standard deviation over three independent training seeds per method. This table compares single-training baselines; Deep Ensembles are omitted from this supplementary check to focus on the architecture sensitivity of post-hoc and single-training methods.}
\label{tab:app_wrn_architecture}
\footnotesize
\begin{adjustbox}{max width=\textwidth}
\begin{tabular}{lccccc}
\toprule
Method & Acc $\uparrow$ & NLL $\downarrow$ & ECE $\downarrow$ & Brier $\downarrow$ & AUROC $\uparrow$ \\
\midrule
MAP & \meanstd{0.739}{0.002} & \meanstd{1.026}{0.017} & \meanstd{0.095}{0.002} & \meanstd{0.373}{0.004} & \meanstd{0.798}{0.021} \\
SWAG-Diag & \meanstd{0.740}{0.001} & \meanstd{1.032}{0.014} & \meanstd{0.097}{0.001} & \meanstd{0.373}{0.004} & \meanstd{0.798}{0.018} \\
MC Dropout & \meanstd{0.747}{0.002} & \meanstd{0.910}{0.005} & \meanstd{0.053}{0.002} & \meanstd{0.350}{0.001} & \meanstd{0.839}{0.020} \\
ELA (last-layer) & \meanstd{0.728}{0.002} & \meanstd{1.124}{0.010} & \meanstd{0.150}{0.003} & \meanstd{0.407}{0.003} & \meanstd{0.852}{0.021} \\
LLA (last-layer) & \meanstd{0.729}{0.002} & \meanstd{1.172}{0.010} & \meanstd{0.207}{0.006} & \meanstd{0.428}{0.003} & \meanstd{0.888}{0.017} \\
TRL & \meanstd{0.740}{0.001} & \meanstd{0.944}{0.007} & \meanstd{0.028}{0.004} & \meanstd{0.359}{0.002} & \meanstd{0.817}{0.029} \\
\bottomrule
\end{tabular}
\end{adjustbox}
\end{table}

Table~\ref{tab:app_wrn_architecture} shows a mixed WideResNet-16-4 result. TRL substantially improves over MAP, SWAG-Diag, ELA, and LLA in NLL, ECE, and Brier score, and has the best ECE among the evaluated methods; MC~Dropout is stronger in accuracy, NLL, and Brier. The calibration benefit therefore transfers beyond ResNet-18, but the ranking among single-training methods changes with architecture and regularization. Appendix~\ref{app:vgg_architecture} reports a second check on a non-residual network.

\paragraph{Seed-count note.}
The clean headline table aggregates five independently trained checkpoints; CIFAR-100-C and the architecture checks use three checkpoint-specific runs; fixed-MAP ablations use posterior-sampling replicates unless their captions state otherwise; the boost-prior ablation uses three independently trained checkpoints. Transport and fine-tuning diagnostics use the seed counts stated locally. These error bars therefore quantify different sources of variability and should be compared only within the corresponding study.

\subsection{Architecture-sensitivity check: VGG-11-BN}
\label{app:vgg_architecture}

As a second architecture check, we run CIFAR-100 with VGG-11-BN-CIFAR. VGG-11-BN has no residual skip connections. Because residual structure contributes some of the near-exact functional invariances that shape low-loss valleys, this non-residual topology provides a useful stress test of whether TRL's calibration behavior transfers beyond residual networks. We reuse the main TRL structural configuration ($k_\perp=30$, $T=40$, $S=25$, 25 FixBN batches) and select only the tube scale on the clean validation split by validation NLL. All three seeds select $\beta_\perp=4.0$, the same value selected by the ResNet-18 experiments, with a clear interior optimum on the validation-NLL curve; this matches the WideResNet selection ($\beta_\perp=3.0$) to within one grid step and indicates that the transverse tempering scale is not strongly architecture-sensitive across the three architectures studied.

\begin{table}[H]
\centering
\caption{Supplementary architecture-sensitivity check on CIFAR-100 with VGG-11-BN-CIFAR. Values are mean $\pm$ standard deviation over three independent training seeds per method. As in the WideResNet-16-4 check, this table compares single-training baselines and omits Deep Ensembles. Among single-checkpoint post-hoc methods (MAP, ELA, LLA, and TRL), TRL gives the best ECE, NLL, and Brier. SWAG-Diag starts from MAP but collects statistics along a short additional fine-tuning trajectory. MC~Dropout, which acquires its uncertainty during training rather than post-hoc (Section~\ref{sec:relation}), is the strongest single-training baseline on this architecture.}
\label{tab:app_vgg}
\vskip 0.1in
\begin{adjustbox}{max width=\textwidth}
\begin{tabular}{lccccc}
\toprule
Method & Acc $\uparrow$ & NLL $\downarrow$ & ECE $\downarrow$ & Brier $\downarrow$ & AUROC $\uparrow$ \\
\midrule
MAP           & $0.678 \pm 0.002$ & $1.377 \pm 0.011$ & $0.134 \pm 0.002$ & $0.462 \pm 0.003$ & $0.780 \pm 0.010$ \\
SWAG-Diag     & \meanstd{0.679}{0.002} & \meanstd{1.377}{0.010} & \meanstd{0.135}{0.001} & \meanstd{0.461}{0.003} & \meanstd{0.778}{0.007} \\
MC Dropout    & \meanstd{0.686}{0.001} & \meanstd{1.135}{0.005} & \meanstd{0.023}{0.001} & \meanstd{0.419}{0.003} & \meanstd{0.808}{0.029} \\
ELA (last-layer) & $0.656 \pm 0.003$ & $1.587 \pm 0.008$ & $0.220 \pm 0.007$ & $0.528 \pm 0.001$ & $0.815 \pm 0.009$ \\
LLA (last-layer) & \meanstd{0.674}{0.002} & \meanstd{1.670}{0.012} & \meanstd{0.324}{0.007} & \meanstd{0.568}{0.003} & \meanstd{0.852}{0.009} \\
TRL           & $0.676 \pm 0.003$ & $1.248 \pm 0.014$ & $0.028 \pm 0.003$ & $0.438 \pm 0.005$ & $0.803 \pm 0.015$ \\
\bottomrule
\end{tabular}
\end{adjustbox}
\vskip -0.05in
\end{table}

Table~\ref{tab:app_vgg} shows that the VGG-11-BN result is also mixed. Among the single-checkpoint post-hoc methods (MAP, ELA, LLA, and TRL), TRL is the strongest: it reduces ECE from $0.134\pm0.002$ (MAP) to $0.028\pm0.003$, roughly a five-fold improvement, and similarly improves NLL and Brier over MAP, ELA, and LLA. It also improves these three metrics over MAP-initialized SWAG-Diag, whose three-seed result is NLL $1.377\pm0.010$, ECE $0.135\pm0.001$, and Brier $0.461\pm0.003$. The SWAG-Diag predictive-entropy AUROC is $0.778\pm0.007$, below TRL's $0.803\pm0.015$, while LLA retains the highest AUROC in the table. On this non-residual network, TRL again yields a markedly better-calibrated posterior predictive than the static local-Gaussian and diagonal checkpoint-initialized baselines.

The comparison with MC~Dropout is the one point where this architecture departs from the ResNet-18 ordering. On VGG-11-BN, MC~Dropout is better than TRL in accuracy, NLL, ECE, and Brier, with non-overlapping three-seed intervals. Because dropout is introduced during training, this comparison does not isolate post-hoc inference alone. Nevertheless, the result exposes a limitation of TRL on this architecture: the deterministic VGG checkpoint does not yield as effective a post-hoc uncertainty model as training-time dropout. Both methods use $S=25$ and follow the method-specific normalization procedures documented in Appendix~\ref{app:implementation}. TRL still improves over the evaluated post-hoc baselines, but the ordering against training-time stochastic methods is architecture-dependent.

\subsection{Runtime and memory breakdown}
\begin{table}[H]
\centering
\caption{Instrumented CIFAR-100 logged-time and peak-VRAM summary. Mean $\pm$ standard deviation over five seeds. The time column is the sum of recorded timing keys; scopes differ by method, and the TRL row contains nested FixBN sub-timers already included in its prediction timers. It is therefore not a normalized, non-overlapping wall-clock benchmark.}
\label{tab:app_runtime_memory}
\footnotesize
\setlength{\tabcolsep}{8pt}
\begin{tabular}{lcc}
\toprule
Method & Logged time (s) & Peak VRAM (GB) $\downarrow$ \\
\midrule
MAP & \meanstd{504.6}{7.2} & \meanstd{0.70}{0.00} \\
ELA / LLA (shared MAP train + fit + pred) & \meanstd{598.3}{7.3} & \meanstd{9.92}{0.00} \\
SWAG-Diag & \meanstd{192.3}{108.5} & \meanstd{0.66}{0.29} \\
MC Dropout & \meanstd{277.5}{292.1} & \meanstd{0.51}{0.34} \\
Deep Ensemble & \meanstd{2444.1}{50.4} & \meanstd{0.79}{0.00} \\
TRL & \meanstd{1985.8}{31.4} & \meanstd{6.42}{0.00} \\
\bottomrule
\end{tabular}
\end{table}

Table~\ref{tab:app_runtime_memory} documents the observed pipeline instrumentation; it is not a normalized wall-clock benchmark. TRL uses one independently optimized backbone but adds substantial post-hoc construction, validation-scale selection, and sampled-network inference. Deep Ensembles use five independently optimized backbones. Because the recorded totals differ in scope and the TRL total contains overlapping nested timers, we make no absolute wall-clock ranking between them. The supported computational distinction is the number and role of the optimization runs, not a claim that the displayed TRL total is intrinsically lower. The runtime columns in Appendix~\ref{app:ablations} are separate study-local measurements and must also be compared only within their own tables.

\section{CIFAR-100-C and Human-Ambiguity Diagnostics}
\label{app:cifar100c}

CIFAR-100-C evaluates corruption robustness without changing the trained model or tuning hyperparameters on corrupted data. This setting is useful for separating deterministic accuracy from probabilistic reliability: under stronger corruption, all methods lose accuracy, but posterior averaging and scalar recalibration can affect NLL, ECE, and Brier differently. The clean-validation-selected TRL configuration is used unchanged. Table~\ref{tab:cifar100c_main_3seed} in the main text retains the posterior/ensemble comparison; the separately audited MAP+TS calibration control is reported in Tables~\ref{tab:app_mapts_c100c}--\ref{tab:app_mapts_c100c_severity}.

\begin{table}[H]
\centering
\caption{CIFAR-100-C by severity for Deep Ensembles, MC Dropout, and TRL. Mean $\pm$ sample standard deviation over three independently trained, seed-matched checkpoints; each TRL posterior is constructed from its corresponding MAP checkpoint.}
\label{tab:app_cifar100c_severity}
\footnotesize
\begin{adjustbox}{max width=\textwidth}
\begin{tabular}{llcccc}
\toprule
Severity & Method & Acc $\uparrow$ & NLL $\downarrow$ & ECE $\downarrow$ & Brier $\downarrow$ \\
\midrule
1 & Deep Ensemble & \meanstd{0.663}{0.001} & \meanstd{1.358}{0.004} & \meanstd{0.045}{0.001} & \meanstd{0.4565}{0.0005} \\
 & MC Dropout & \meanstd{0.624}{0.003} & \meanstd{1.554}{0.012} & \meanstd{0.094}{0.004} & \meanstd{0.511}{0.004} \\
 & TRL & \meanstd{0.623}{0.005} & \meanstd{1.540}{0.030} & \meanstd{0.047}{0.005} & \meanstd{0.506}{0.005} \\
\midrule
2 & Deep Ensemble & \meanstd{0.575}{0.001} & \meanstd{1.811}{0.004} & \meanstd{0.075}{0.001} & \meanstd{0.565}{0.001} \\
 & MC Dropout & \meanstd{0.538}{0.005} & \meanstd{2.035}{0.039} & \meanstd{0.131}{0.007} & \meanstd{0.621}{0.008} \\
 & TRL & \meanstd{0.535}{0.004} & \meanstd{1.998}{0.041} & \meanstd{0.079}{0.007} & \meanstd{0.611}{0.006} \\
\midrule
3 & Deep Ensemble & \meanstd{0.517}{0.002} & \meanstd{2.183}{0.012} & \meanstd{0.105}{0.001} & \meanstd{0.639}{0.003} \\
 & MC Dropout & \meanstd{0.482}{0.006} & \meanstd{2.423}{0.074} & \meanstd{0.162}{0.010} & \meanstd{0.696}{0.012} \\
 & TRL & \meanstd{0.480}{0.004} & \meanstd{2.352}{0.050} & \meanstd{0.108}{0.005} & \meanstd{0.679}{0.005} \\
\midrule
4 & Deep Ensemble & \meanstd{0.447}{0.002} & \meanstd{2.677}{0.018} & \meanstd{0.143}{0.003} & \meanstd{0.729}{0.005} \\
 & MC Dropout & \meanstd{0.413}{0.006} & \meanstd{2.974}{0.087} & \meanstd{0.206}{0.010} & \meanstd{0.790}{0.013} \\
 & TRL & \meanstd{0.412}{0.003} & \meanstd{2.841}{0.059} & \meanstd{0.148}{0.003} & \meanstd{0.763}{0.003} \\
\midrule
5 & Deep Ensemble & \meanstd{0.344}{0.002} & \meanstd{3.429}{0.026} & \meanstd{0.201}{0.004} & \meanstd{0.859}{0.005} \\
 & MC Dropout & \meanstd{0.317}{0.008} & \meanstd{3.785}{0.086} & \meanstd{0.268}{0.010} & \meanstd{0.920}{0.014} \\
 & TRL & \meanstd{0.315}{0.004} & \meanstd{3.589}{0.066} & \meanstd{0.209}{0.009} & \meanstd{0.886}{0.009} \\
\bottomrule
\end{tabular}
\end{adjustbox}
\end{table}

Table~\ref{tab:cifar100c_main_3seed} in the main text reports the aggregate CIFAR-100-C summary, while Table~\ref{tab:app_cifar100c_severity} captures the degradation trend across corruption strengths. Deep Ensembles dominate the posterior/ensemble table at every severity, while TRL improves the probabilistic metrics over MC Dropout and remains close to the ensemble ECE at mild and moderate corruptions. The calibration-only MAP+TS control has lower NLL than TRL at each severity; ECE and Brier are close, with ordering varying by metric and severity. At severity 5, the gap between within-basin methods and Deep Ensembles widens, reflecting the limitation of a single-basin approximation under strong distribution shift. TRL is therefore best interpreted as a posterior approximation that remains competitive with scalar recalibration while providing posterior samples and functional diagnostics, not as a universal replacement for either temperature scaling or cross-basin ensemble diversity.

\subsection{CIFAR-10H human-label alignment}
\label{app:cifar10h}
CIFAR-10H provides an empirical human label distribution $q_H(\cdot\mid x)$ for each of the 10,000 CIFAR-10 test images, with roughly 51 human judgments per image on average \citep{peterson2019human}. We use these distributions only for evaluation: no CIFAR-10H labels enter training, posterior construction, FixBN, sample-budget choice, or hyperparameter selection. The official row order was independently checked against the raw CIFAR-10H test indices and true labels; all 10,000 rows match the CIFAR-10 test ordering used by our saved predictions. We evaluate three independently trained ResNet-18 seeds from the stored CIFAR-10 runs. Each deterministic MAP checkpoint was trained from scratch for 50 epochs with SGD (learning rate $0.1$, momentum $0.9$, weight decay $5\times10^{-4}$), cosine annealing, batch size 128, a 45,000/5,000 train/validation split, and standard random-crop plus horizontal-flip augmentation. Deep-Ensemble members use the same 50-epoch training recipe; MC Dropout uses the corresponding dropout-trained model.

For the CIFAR-10H evaluation, MAP is deterministic and the Deep Ensemble averages five members within each seed-group. MC Dropout uses 25 stochastic predictive passes after the one-time 20-batch BatchNorm recalibration. TRL replays each frozen posterior/spine with the validation-selected $\beta_\perp$ stored in that payload (\texttt{best\_tube\_scale}); the associated validation sweep selected $\beta_\perp=2.0$ in each of seeds 0, 1, and 2. No HVP, spine reconstruction, or tube-scale re-selection is performed. We draw 25 fresh posterior samples with a fixed evaluation seed and apply the 25-batch \emph{rolling} FixBN procedure to each draw (running statistics are not reset between draws). SWAG-Diag is collected from the corresponding MAP checkpoint with 10 fine-tuning epochs at learning rate $10^{-3}$, unbiased arithmetic diagonal moments, sample scale 1.0, 20 posterior draws, and independent 20-batch reset FixBN. CIFAR-10 SWAG caches initialized from an ensemble member are not used. These method-specific FixBN semantics reflect artifact provenance, not CIFAR-10H tuning: the frozen TRL replay retains its rolling protocol, whereas SWAG-Diag uses the reset protocol. A separate paired seed-0 CIFAR-100 audit (Appendix~\ref{app:fixbn_protocol}) found only small rolling-versus-reset changes for both methods and did not trigger reruns; this is a sensitivity/provenance audit, not a matched CIFAR-10H recalibration experiment. Predictions are evaluated separately for each training seed and only the resulting metrics are aggregated across seeds; predictions are never averaged across the three training seeds.

For each image we compare the model predictive distribution $p$ with $q_H$ using $\KL(q_H\|p)$, soft-target Brier score, Jensen--Shannon (JS) divergence, signed entropy bias $H(p)-H(q_H)$, and Pearson correlation between predictive and human entropy across images. Table~\ref{tab:cifar10h_alignment} reports the full 10,000-image test set and the 5,607-image subset with nonzero human-label entropy. The latter isolates images on which the human annotations exhibit perceptual ambiguity.

\begin{table}[H]
\centering
\caption{CIFAR-10H alignment (mean $\pm$ sample SD over three independently trained seeds). Lower is better for KL, Brier, and JS; entropy bias is signed with zero ideal; higher entropy correlation is better. ``Ambiguous'' means $H(q_H)>0$ (5,607/10,000 images). No CIFAR-10H labels are used for training or tuning.}
\label{tab:cifar10h_alignment}
\scriptsize
\setlength{\tabcolsep}{3.2pt}
\begin{tabular}{llccccc}
\toprule
Subset & Method & KL $\downarrow$ & Brier $\downarrow$ & JS $\downarrow$ & $H$ bias ($0$ ideal) & $H$ corr. $\uparrow$ \\
\midrule
All & MAP & \meanstd{0.3819}{0.0180} & \meanstd{0.0857}{0.0049} & \meanstd{0.0422}{0.0017} & \meanstd{-0.0687}{0.0026} & \meanstd{0.2827}{0.0136} \\
 & SWAG-Diag & \meanstd{0.3894}{0.0190} & \meanstd{0.0853}{0.0054} & \meanstd{0.0420}{0.0019} & \meanstd{-0.0722}{0.0023} & \meanstd{0.2796}{0.0113} \\
 & MC Dropout & \meanstd{0.3026}{0.0051} & \meanstd{0.0769}{0.0029} & \meanstd{0.0406}{0.0011} & \meanstd{-0.0260}{0.0028} & \meanstd{0.3290}{0.0064} \\
 & TRL (ours) & \meanstd{0.2939}{0.0110} & \meanstd{0.0776}{0.0044} & \meanstd{0.0434}{0.0018} & \meanstd{+0.0243}{0.0042} & \meanstd{0.3508}{0.0113} \\
 & Deep Ensemble & \meanstd{0.2838}{0.0040} & \meanstd{0.0630}{0.0010} & \meanstd{0.0361}{0.0003} & \meanstd{-0.0254}{0.0005} & \meanstd{0.3484}{0.0033} \\
\midrule
Ambiguous & MAP & \meanstd{0.6175}{0.0254} & \meanstd{0.1249}{0.0059} & \meanstd{0.0634}{0.0021} & \meanstd{-0.1559}{0.0031} & \meanstd{0.2562}{0.0132} \\
 & SWAG-Diag & \meanstd{0.6311}{0.0271} & \meanstd{0.1244}{0.0070} & \meanstd{0.0632}{0.0024} & \meanstd{-0.1608}{0.0028} & \meanstd{0.2541}{0.0090} \\
 & MC Dropout & \meanstd{0.4851}{0.0082} & \meanstd{0.1113}{0.0044} & \meanstd{0.0598}{0.0017} & \meanstd{-0.0987}{0.0043} & \meanstd{0.3011}{0.0113} \\
 & TRL (ours) & \meanstd{0.4628}{0.0133} & \meanstd{0.1098}{0.0047} & \meanstd{0.0618}{0.0018} & \meanstd{-0.0351}{0.0037} & \meanstd{0.3209}{0.0118} \\
 & Deep Ensemble & \meanstd{0.4622}{0.0060} & \meanstd{0.0923}{0.0014} & \meanstd{0.0540}{0.0005} & \meanstd{-0.0943}{0.0006} & \meanstd{0.3165}{0.0049} \\
\bottomrule
\end{tabular}
\end{table}

Deep Ensembles provide the strongest overall match to the human label distribution under KL, soft Brier, and JS on the full set. TRL is second in mean KL and has the highest mean entropy correlation, but its larger soft-target Brier and JS rule out a claim of uniformly better human-distribution matching. On the human-ambiguous subset, TRL nearly matches the five-member ensemble in mean KL ($0.4628$ versus $0.4622$), yields a comparable mean entropy correlation ($0.3209$ versus $0.3165$), and substantially reduces systematic underestimation of average human ambiguity (entropy bias $-0.0351$ versus $-0.0943$ for the ensemble and $-0.0987$ for MC Dropout). Deep Ensembles nevertheless remain clearly stronger under soft Brier and JS. CIFAR-10H thus serves as a complementary diagnostic: TRL tracks relative variation in human perceptual ambiguity and its average level relatively well from a single trained MAP solution, but does not reproduce the full human label distribution as closely as cross-training ensemble diversity.

\section{Ablations}
\label{app:ablations}

Unless stated otherwise, the CIFAR-100/ResNet-18 ablations use three posterior-sampling replicates at one fixed MAP checkpoint and the reported default $k_\perp=30$, $T=40$, $\beta_\perp=4.0$, $S=25$, with 25 rolling-mode FixBN batches. The prior-boost study instead aggregates three independently trained checkpoints. Each sweep is a separate stochastic evaluation, so repeated default rows and runtime values should be compared within, not across, tables. The released code defaults to reset FixBN as described in Appendix~\ref{app:implementation}.

\subsection{Transverse rank $k_\perp$}
\begin{table}[H]
\centering
\caption{Transverse-rank ablation on CIFAR-100. Mean $\pm$ standard deviation over three posterior-sampling replicates at one fixed MAP checkpoint.}
\label{tab:app_ablation_k}
\footnotesize
\begin{adjustbox}{max width=\textwidth}
\begin{tabular}{lcccccc}
\toprule
$k_\perp$ & Acc $\uparrow$ & NLL $\downarrow$ & ECE $\downarrow$ & Brier $\downarrow$ & AUROC $\uparrow$ & Runtime (s) $\downarrow$ \\
\midrule
5 & \meanstd{0.743}{0.002} & \meanstd{1.004}{0.004} & \meanstd{0.076}{0.004} & \meanstd{0.364}{0.002} & \meanstd{0.864}{0.016} & \meanstd{335.585}{13.889} \\
10 & \meanstd{0.743}{0.001} & \meanstd{0.983}{0.005} & \meanstd{0.058}{0.006} & \meanstd{0.359}{0.002} & \meanstd{0.865}{0.016} & \meanstd{401.157}{3.204} \\
20 & \meanstd{0.744}{0.002} & \meanstd{0.963}{0.006} & \meanstd{0.031}{0.006} & \meanstd{0.357}{0.001} & \meanstd{0.868}{0.018} & \meanstd{560.737}{19.745} \\
30 & \meanstd{0.743}{0.003} & \meanstd{0.954}{0.005} & \meanstd{0.0147}{0.0003} & \meanstd{0.356}{0.001} & \meanstd{0.870}{0.022} & \meanstd{706.998}{38.901} \\
50 & \meanstd{0.741}{0.001} & \meanstd{0.959}{0.004} & \meanstd{0.032}{0.002} & \meanstd{0.360}{0.001} & \meanstd{0.862}{0.017} & \meanstd{1036.529}{59.444} \\
\bottomrule
\end{tabular}
\end{adjustbox}
\end{table}

Rank materially affects predictive performance. With $k_\perp=5$, NLL and ECE remain substantially above the default configuration. Increasing the rank to 20 captures most of the likelihood improvement, and $k_\perp=30$ gives the best calibration and Brier score. Moving to $k_\perp=50$ does not improve the predictive metrics and substantially increases runtime, suggesting that the useful stiff subspace is low-rank but not extremely small. We therefore use $k_\perp=30$ as a fidelity--cost compromise.

\subsection{Number of stored spine anchors $T$}
\begin{table}[H]
\centering
\caption{Stored-spine-anchor ablation on CIFAR-100. Mean $\pm$ standard deviation over three posterior-sampling replicates at one fixed MAP checkpoint.}
\label{tab:app_ablation_T}
\footnotesize
\begin{adjustbox}{max width=\textwidth}
\begin{tabular}{lcccccc}
\toprule
$T$ & Acc $\uparrow$ & NLL $\downarrow$ & ECE $\downarrow$ & Brier $\downarrow$ & AUROC $\uparrow$ & Runtime (s) $\downarrow$ \\
\midrule
5 & \meanstd{0.743}{0.004} & \meanstd{0.956}{0.005} & \meanstd{0.015}{0.004} & \meanstd{0.356}{0.001} & \meanstd{0.861}{0.020} & \meanstd{417.377}{5.312} \\
10 & \meanstd{0.743}{0.002} & \meanstd{0.955}{0.004} & \meanstd{0.014}{0.003} & \meanstd{0.356}{0.001} & \meanstd{0.870}{0.019} & \meanstd{469.773}{21.452} \\
20 & \meanstd{0.743}{0.003} & \meanstd{0.955}{0.004} & \meanstd{0.014}{0.004} & \meanstd{0.356}{0.001} & \meanstd{0.869}{0.021} & \meanstd{538.676}{22.166} \\
40 & \meanstd{0.743}{0.003} & \meanstd{0.955}{0.006} & \meanstd{0.015}{0.002} & \meanstd{0.356}{0.001} & \meanstd{0.870}{0.022} & \meanstd{694.303}{40.407} \\
\bottomrule
\end{tabular}
\end{adjustbox}
\end{table}

Table~\ref{tab:app_ablation_T} varies the number of stored anchors $T$. Because the CIFAR setting keeps $\Delta s=0.01$ fixed, this also changes the nominal longitudinal extent $(T-1)\Delta s$; the sweep therefore probes anchor count and extent jointly; it does not isolate discretization density. Performance changes very little from $T=10$ to $T=40$, and even $T=5$ remains competitive. Boldface identifies the best unrounded mean separately for each metric; because the differences among $T=10$, $20$, and $40$ are small relative to run-to-run variation and often disappear at the reported precision, the scattered bold entries do not indicate a uniquely preferred anchor count. Thus extending the retained spine beyond $T=10$--$20$ adds little clean-test predictive benefit in this setting. The choice $T=40$ is therefore conservative: it increases runtime but does not materially change clean-test predictive quality relative to $T=10$--$20$.

\subsection{Tube scale $\beta_\perp$}
\begin{table}[H]
\centering
\caption{Tube-scale sensitivity on CIFAR-100. Mean $\pm$ standard deviation over three posterior-sampling replicates at one fixed MAP checkpoint.}
\label{tab:app_ablation_tube}
\footnotesize
\begin{adjustbox}{max width=\textwidth}
\begin{tabular}{lccccc}
\toprule
$\beta_\perp$ & Acc $\uparrow$ & NLL $\downarrow$ & ECE $\downarrow$ & Brier $\downarrow$ & AUROC $\uparrow$ \\
\midrule
2.0 & \meanstd{0.744}{0.002} & \meanstd{0.988}{0.007} & \meanstd{0.067}{0.003} & \meanstd{0.361}{0.002} & \meanstd{0.861}{0.020} \\
3.0 & \meanstd{0.744}{0.002} & \meanstd{0.965}{0.007} & \meanstd{0.041}{0.004} & \meanstd{0.356}{0.001} & \meanstd{0.865}{0.021} \\
4.0 & \meanstd{0.743}{0.003} & \meanstd{0.955}{0.006} & \meanstd{0.014}{0.002} & \meanstd{0.356}{0.001} & \meanstd{0.870}{0.022} \\
6.0 & \meanstd{0.738}{0.004} & \meanstd{0.989}{0.005} & \meanstd{0.063}{0.009} & \meanstd{0.370}{0.001} & \meanstd{0.876}{0.022} \\
10.0 & \meanstd{0.703}{0.004} & \meanstd{1.257}{0.015} & \meanstd{0.206}{0.014} & \meanstd{0.464}{0.006} & \meanstd{0.862}{0.021} \\
20.0 & \meanstd{0.387}{0.022} & \meanstd{2.551}{0.056} & \meanstd{0.195}{0.016} & \meanstd{0.823}{0.012} & \meanstd{0.581}{0.095} \\
\bottomrule
\end{tabular}
\end{adjustbox}
\end{table}

Table~\ref{tab:app_ablation_tube} shows how tube scale controls the posterior width in the transported transverse coordinates. In the implemented sampler it is best interpreted as a validation-selected calibration or tempering parameter for the approximate tube: the continuous idealization motivates a transverse covariance structure, while $\beta_\perp$ sets the effective predictive spread after low-rank truncation and stochastic curvature estimation. The sensitivity curve has a clear optimum for probabilistic prediction: $\beta_\perp=4.0$ gives the best NLL and ECE and the lowest unrounded Brier score, although Brier rounds to a tie with $\beta_\perp=3.0$ at three decimals. The $\beta_\perp=2.0$ and $\beta_\perp=3.0$ settings tie at the displayed precision in accuracy, while both are worse calibrated than $\beta_\perp=4.0$. Larger values over-disperse the posterior: at $\beta_\perp=10$ and $20$, samples move into regions that substantially degrade both accuracy and likelihood. The predictive-entropy AUROC is relatively flat for moderate tube scales and peaks at $\beta_\perp=6.0$ in this rerun, again showing that the scale that optimizes OOD ranking need not optimize calibrated in-distribution prediction. We select $\beta_\perp=4.0$ because the main objective is probabilistic predictive quality.

\subsection{FixBN batches}
\begin{table}[H]
\centering
\caption{Rolling-mode FixBN batch-count ablation for TRL. Mean $\pm$ standard deviation over three posterior-sampling replicates at one fixed MAP checkpoint.}
\label{tab:app_ablation_fixbn}
\footnotesize
\begin{adjustbox}{max width=\textwidth}
\begin{tabular}{lcccccc}
\toprule
FixBN batches & Acc $\uparrow$ & NLL $\downarrow$ & ECE $\downarrow$ & Brier $\downarrow$ & AUROC $\uparrow$ & Runtime (s) $\downarrow$ \\
\midrule
1 & \meanstd{0.741}{0.001} & \meanstd{1.001}{0.009} & \meanstd{0.084}{0.008} & \meanstd{0.372}{0.002} & \meanstd{0.841}{0.025} & \meanstd{603.361}{10.327} \\
5 & \meanstd{0.743}{0.003} & \meanstd{0.965}{0.005} & \meanstd{0.043}{0.007} & \meanstd{0.360}{0.001} & \meanstd{0.853}{0.019} & \meanstd{618.030}{28.308} \\
10 & \meanstd{0.744}{0.002} & \meanstd{0.955}{0.004} & \meanstd{0.020}{0.007} & \meanstd{0.3569}{0.0003} & \meanstd{0.861}{0.020} & \meanstd{636.221}{28.379} \\
25 & \meanstd{0.743}{0.003} & \meanstd{0.955}{0.005} & \meanstd{0.014}{0.002} & \meanstd{0.356}{0.001} & \meanstd{0.870}{0.022} & \meanstd{691.244}{23.798} \\
\bottomrule
\end{tabular}
\end{adjustbox}
\end{table}

Because Table~\ref{tab:app_ablation_fixbn} uses the rolling protocol, increasing the batch count changes both the recalibration budget and the amount of running-statistics carry-over from the preceding posterior draw. It should therefore be read as a joint rolling-state/batch-count sensitivity study rather than a pure recalibration-budget ablation. Top-1 accuracy changes little; NLL is similar at 10 and 25 batches, while ECE and Brier continue to change. FixBN uses unlabeled clean training inputs and does not update weights, so it is an inference-time consistency correction, not posterior support. Its necessity does not explain TRL's gain: SWAG-Diag uses its own documented FixBN protocol yet remains near MAP in clean ECE, and the paired rolling-versus-reset audit changes SWAG-Diag only negligibly.

\subsection{Backbone prior boost}
\label{sec:boost-ablation}

The large-network TRL prior is a two-block, block-isotropic approximation. The classifier head receives the base precision $\lambda_{\mathrm{base}}$ inherited from the last-layer Laplace fit, while non-head parameters receive
\[
\lambda_{\mathrm{backbone}}=\max(c\,\lambda_{\mathrm{base}},5).
\]
The floor value $5$ is a minimum non-head precision that prevents the no-boost setting from becoming an effectively unregularized backbone posterior. Across the three checkpoints used in this ablation, $\lambda_{\mathrm{base}}$ is close to $5$, ranging from $4.76$ to $4.95$; therefore $c=0$ corresponds to a head/backbone prior of comparable strength, while the floor is inactive for the adopted setting $c=50$.

Table~\ref{tab:boost} reports two complementary sensitivity checks on three independently trained ResNet-18 MAP checkpoints in the main CIFAR-100 setting. Panel~A fixes the validation-selected tube scale $\beta_\perp=4$ and varies $c$, measuring NLL and ECE on both the held-out clean validation split and the test split. On the validation split, the standard validation-NLL selection rule returns $c=50$, which also attains the best validation ECE. Panel~B varies $c$ jointly with $\beta_\perp$, testing on the test split whether the two quantities merely form a single effective scale. The selected value $c=50$ is then used unchanged in the WideResNet-16-4 and VGG-11-BN architecture-sensitivity checks.

\begin{table}[H]
\centering
\caption{Backbone-prior boost ablations on CIFAR-100/ResNet-18. Values are mean $\pm$ standard deviation over three MAP checkpoints. Panel A reports validation and test NLL/ECE at the validation-selected tube scale. The value $c=50$ is selected exclusively by validation NLL; test metrics are reported only after selection. Panel B reports the joint test sweep and shows that the boost $c$ and tube scale $\beta_\perp$ do not reduce to a single effective product.}
\label{tab:boost}
\scriptsize
\begin{adjustbox}{max width=\textwidth}
\begin{tabular}{cccrrrr}
\toprule
\multicolumn{7}{l}{Panel A: boost sweep at validation-selected $\beta_\perp=4$} \\
$c$ & $\beta_\perp$ & $c\beta_\perp$ & NLL (val) $\downarrow$ & NLL (test) $\downarrow$ & ECE (val) $\downarrow$ & ECE (test) $\downarrow$ \\
\midrule
0   & 4 &   0 & $2.000{\pm}0.058$ & $1.992{\pm}0.058$ & $0.280{\pm}0.004$ & $0.283{\pm}0.006$ \\
10  & 4 &  40 & $1.145{\pm}0.010$ & $1.135{\pm}0.010$ & $0.160{\pm}0.003$ & $0.161{\pm}0.006$ \\
50  & 4 & 200 & $0.964{\pm}0.004$ & $0.954{\pm}0.004$ & $0.019{\pm}0.002$ & $0.015{\pm}0.003$ \\
100 & 4 & 400 & $0.972{\pm}0.007$ & $0.963{\pm}0.007$ & $0.041{\pm}0.002$ & $0.041{\pm}0.004$ \\
\midrule
\multicolumn{7}{l}{Panel B: joint boost--tube-scale test sweep} \\
$c$ & $\beta_\perp$ & $c\beta_\perp$ & Acc $\uparrow$ & NLL $\downarrow$ & ECE $\downarrow$ & Brier $\downarrow$ \\
\midrule
25  & 2 &  50 & $0.7459{\pm}0.0007$ & $0.9652{\pm}0.0066$ & $0.0444{\pm}0.0036$ & $0.3540{\pm}0.0027$ \\
25  & 4 & 100 & $0.7438{\pm}0.0022$ & $0.9814{\pm}0.0026$ & $0.0568{\pm}0.0082$ & $0.3652{\pm}0.0031$ \\
25  & 8 & 200 & $0.6820{\pm}0.0066$ & $1.4140{\pm}0.0135$ & $0.2520{\pm}0.0074$ & $0.5176{\pm}0.0064$ \\
50  & 2 & 100 & $0.7461{\pm}0.0002$ & $0.9834{\pm}0.0073$ & $0.0647{\pm}0.0023$ & $0.3572{\pm}0.0026$ \\
50  & 4 & 200 & $0.7461{\pm}0.0027$ & $0.9541{\pm}0.0041$ & $0.0155{\pm}0.0028$ & $0.3537{\pm}0.0028$ \\
50  & 8 & 400 & $0.7285{\pm}0.0033$ & $1.1186{\pm}0.0094$ & $0.1547{\pm}0.0097$ & $0.4138{\pm}0.0044$ \\
100 & 2 & 200 & $0.7452{\pm}0.0005$ & $1.0006{\pm}0.0075$ & $0.0780{\pm}0.0020$ & $0.3607{\pm}0.0025$ \\
100 & 4 & 400 & $0.7462{\pm}0.0009$ & $0.9632{\pm}0.0065$ & $0.0410{\pm}0.0038$ & $0.3537{\pm}0.0028$ \\
100 & 8 & 800 & $0.7430{\pm}0.0024$ & $0.9885{\pm}0.0028$ & $0.0647{\pm}0.0087$ & $0.3678{\pm}0.0029$ \\
\bottomrule
\end{tabular}
\end{adjustbox}
\end{table}

The two lowest displayed Brier means in Panel~B, at $(c,\beta_\perp)=(50,4)$ and $(100,4)$, are indistinguishable at the reported precision; no unique Brier optimum is designated without the corresponding unrounded aggregate.

Panel~A shows that the two-block prior magnitude is material for calibrated sampling. Validation NLL selects $c=50$, which also gives the lowest validation ECE. The no-boost setting $c=0$ collapses, $c=10$ remains poorly calibrated, and $c=100$ over-regularizes, producing a U-shaped response in the backbone-to-head precision ratio. Because $c=10$ already separates backbone and head precision but still performs poorly, the result points to the magnitude of the boost rather than merely the existence of a two-block distinction. Test metrics, evaluated only after validation selection, preserve the same ordering.

Panel~B shows joint sensitivity to the boost and tube scale. Equal values of $c\beta_\perp$ do not imply equivalent behavior: the three configurations with $c\beta_\perp=200$ have test ECE values $0.2520$, $0.0155$, and $0.0780$. We therefore select $c$ and $\beta_\perp$ separately. This sweep rules out $c\beta_\perp$ itself as a sufficient one-dimensional summary; it does not identify a unique effective scalar scale for the joint behavior.

\section{Transported Eigenspace Diagnostics}
\label{app:transport}

\paragraph{Source of variability.}
Unless a caption states otherwise, seed counts in this appendix index checkpoint-specific runs; they do not denote repeated posterior draws from one fixed checkpoint. The low-loss and functional-disagreement studies use five independently trained MAP checkpoints and their corresponding TRL constructions. The more expensive stale-eigenspace, basis-transport, single-checkpoint, and fresh-refresh studies use the three independently trained checkpoint-specific runs for seeds 0--2.

This appendix measures the consequences of transporting the transverse eigenspace instead of recomputing it locally. We first verify that the spine remains in a low-loss region and quantify its functional drift. We then measure eigenspace rotation and test its predictive effect through comparisons with a fixed MAP basis, a single-checkpoint tube, and a fresh-refresh variant.

The diagnostics show substantial rotation of the local stiff subspace, but the predictive behavior in this benchmark is captured mainly by the MAP-identified low-rank transverse subspace and tube-scale calibration. FixBN supplies the BatchNorm consistency correction required for sampled-weight prediction. The spine remains in a low-loss corridor with limited but nonzero functional variation, and its incremental contribution on clean CIFAR-100 is modest.

The experiments below also quantify the fidelity--cost trade-off behind the practical choices in Section~\ref{sec:trl-practical}. The ideal tube suggests exact low-curvature tangents and local transverse covariance factors, but for the main ResNet/CIFAR-100 regime, the predictive payoff of making these objects more exact is limited compared with the cost. The important mechanism is not exact tracking of a continuous geodesic; it is the combination of a MAP-identified stiff subspace, a low-loss complement trajectory, validation-calibrated transverse scale, and posterior averaging, together with the BatchNorm recalibration needed for consistent sampled-network evaluation.

\subsection{Low-loss spine diagnostic}
\label{app:spine_loss_diagnostic}
The construction aims to keep discrete spine anchors near the MAP loss level. We assess this directly by evaluating validation cross-entropy at every stored anchor after the same FixBN recalibration used for posterior prediction. This anchor-wise check does not certify a continuous connecting path, basin identity, or every transverse draw. Let $\ell_{\mathrm{val}}(\theta)$ denote validation mean cross-entropy after FixBN recalibration. For a spine anchor $\gamma_t$, we report the absolute loss change
\begin{align}
\Delta_{\mathrm{loss}}(t)=\ell_{\mathrm{val}}(\gamma_t)-\ell_{\mathrm{val}}(\gamma_0),
\label{eq:spine-loss-change}
\end{align}
In Eq.~\eqref{eq:spine-loss-change}, $\gamma_0$ denotes the MAP anchor of the discrete spine. This diagnostic is separate from hyperparameter selection: it is not used to choose $\beta_\perp$, rank, or stored-anchor count. Its purpose is to check whether the predictor--corrector trajectory remains within a low-loss corridor without relying on posterior averaging to recover from high-loss anchors.

\begin{table}[H]
\centering
\caption{Low-loss spine diagnostic on CIFAR-100 with ResNet-18. We evaluate validation cross-entropy at all discrete spine anchors after FixBN recalibration and report changes relative to $\gamma_0$. Values are mean $\pm$ standard deviation over five independently trained checkpoint-specific runs, one per MAP checkpoint.}
\label{tab:app_spine_loss_profile}
\footnotesize
\begin{adjustbox}{max width=0.78\textwidth}
\begin{tabular}{lc}
\toprule
Quantity & Value \\
\midrule
Mean $\Delta_{\mathrm{loss}}$ along spine & \meanstd{0.0030}{0.0031} \\
Mean $|\Delta_{\mathrm{loss}}|$ along spine & \meanstd{0.0045}{0.0015} \\
Maximum $\Delta_{\mathrm{loss}}$ along spine & \meanstd{0.0131}{0.0039} \\
Endpoint $\Delta_{\mathrm{loss}}$ & \meanstd{0.0098}{0.0040} \\
Validation CE at $\gamma_0$ & \meanstd{1.0298}{0.0384} \\
Validation CE at endpoint & \meanstd{1.0396}{0.0410} \\
\bottomrule
\end{tabular}
\end{adjustbox}
\end{table}

Across the five checkpoint-specific runs in Table~\ref{tab:app_spine_loss_profile}, validation loss remains close to the MAP anchor throughout the constructed path. The mean loss change along the spine is $0.0030\pm0.0031$, and the endpoint increase is $0.0098\pm0.0040$ cross-entropy units relative to a base validation loss of $1.0298\pm0.0384$. Some intermediate points decrease the validation loss for individual seeds, and the maximum increase along each spine remains small. After BatchNorm recalibration, the stored trajectory therefore stays within a narrow low-loss corridor.

\subsection{Functional disagreement along the spine}
\label{app:spine_functional_diagnostic}
Low validation loss alone does not determine whether the spine changes the represented predictor or merely follows near-exact functional invariances. We therefore compare predictions at each spine anchor $\gamma_t$ with the MAP anchor $\gamma_0$ on the validation split after the same FixBN recalibration used for posterior prediction. We report top-1 disagreement, Jensen--Shannon (JS) divergence, cosine distance between predictive probability vectors, and the corresponding validation cross-entropy change, denoted $\Delta_{\mathrm{CE}}(t)=\ell_{\mathrm{val}}(\gamma_t)-\ell_{\mathrm{val}}(\gamma_0)$ within this diagnostic run.

\begin{table}[H]
\centering
\caption{Functional disagreement along the CIFAR-100 ResNet-18 spine. We compare predictions at each spine anchor $\gamma_t$ with the MAP anchor $\gamma_0$ after FixBN recalibration. Values are mean $\pm$ standard deviation over five independently trained checkpoint-specific runs, one per MAP checkpoint.}
\label{tab:app_spine_functional_disagreement}
\footnotesize
\begin{adjustbox}{max width=0.82\textwidth}
\begin{tabular}{lc}
\toprule
Quantity & Value \\
\midrule
Mean top-1 disagreement along spine & \meanstd{0.0316}{0.0028} \\
Maximum top-1 disagreement along spine & \meanstd{0.0523}{0.0050} \\
Endpoint top-1 disagreement & \meanstd{0.0508}{0.0054} \\
Mean JS divergence along spine & \meanstd{0.0028}{0.0004} \\
Endpoint JS divergence & \meanstd{0.0063}{0.0009} \\
Mean cosine distance along spine & \meanstd{0.0053}{0.0008} \\
Endpoint cosine distance & \meanstd{0.0117}{0.0017} \\
Endpoint $\Delta_{\mathrm{CE}}$ & \meanstd{0.0104}{0.0028} \\
\bottomrule
\end{tabular}
\end{adjustbox}
\end{table}

At the endpoint in Table~\ref{tab:app_spine_functional_disagreement}, about $5.1\%$ of validation examples change top-1 prediction relative to $\gamma_0$, with JS divergence $0.0063\pm0.0009$ and a validation cross-entropy increase of $0.0104\pm0.0028$. The spine is therefore not functionally identical to the MAP anchor, even though it remains low-loss. The low-loss profile in Table~\ref{tab:app_spine_loss_profile} and the functional-disagreement summary here come from separate five-checkpoint diagnostic executions, each applying FixBN within its own run. Their endpoint cross-entropy aggregates differ slightly for that reason; both place the endpoint approximately $0.01$ cross-entropy units above $\gamma_0$.

Although this longitudinal variation is measurable, averaging over it does not measurably improve predictions beyond the single-checkpoint sampler in Table~\ref{tab:app_fresh_refresh_single}. At this magnitude, the induced predictive change is comparable to seed-level variability and is largely redundant with local transverse averaging. On CIFAR-100/ResNet-18, most of TRL's predictive gain is already captured by the curvature-selected transverse subspace identified at the MAP; the spine supplies low-loss longitudinal support but less additional predictive diversity in this regime.

\subsection{Subspace similarity and eigenvalue drift}
The practical TRL implementation transports a low-rank transverse structure without recomputing a fresh local eigendecomposition at every spine anchor. This section quantifies the geometric drift introduced by that approximation. Let $Q_t^{\mathrm{tr}}\in\mathbb{R}^{K\times k}$ denote the transported transverse basis at spine location $t$, let $Q_t^{\mathrm{fr}}$ denote a freshly recomputed local top-$k$ eigenspace at the same location, and let $Q_0$ denote the original MAP basis held fixed along the spine. Here $Q_t^{\mathrm{tr}}$ is the diagnostic notation for the same transported basis denoted by $N_t$ in the sampler of Eq.~\eqref{eq:practical-trl-sample}. For two orthonormal bases $Q_a,Q_b\in\mathbb{R}^{K\times k}$, we measure subspace similarity by
\begin{align}
\mathcal{O}(Q_a,Q_b)=\frac{1}{k}\|Q_a^\top Q_b\|_F^2,
\label{eq:subspace-overlap}
\end{align}
Here $\|\cdot\|_F$ denotes the Frobenius norm. The overlap in Eq.~\eqref{eq:subspace-overlap} is invariant to rotations and permutations within the two $k$-dimensional subspaces. This is important because individual eigenvectors inside a nearly degenerate stiff subspace are not identifiable. We report both $\mathcal{O}(Q_t^{\mathrm{tr}},Q_t^{\mathrm{fr}})$ and $\mathcal{O}(Q_0,Q_t^{\mathrm{fr}})$, together with principal-angle and eigenvalue-drift summaries. The fresh stiff subspace rotates substantially along the spine: in Table~\ref{tab:app_stale_subspace}, transported-fresh overlap falls to approximately $0.20$ at the endpoint.

\begin{table}[H]
\centering
\caption{Stale-eigenspace mechanism diagnostic along the TRL spine. Mean $\pm$ standard deviation over three independently trained checkpoint-specific runs (training seeds 0--2). Transported-fresh overlap denotes $\mathcal{O}(Q_t^{\mathrm{tr}},Q_t^{\mathrm{fr}})$, and MAP-fresh overlap denotes $\mathcal{O}(Q_0,Q_t^{\mathrm{fr}})$.}
\label{tab:app_stale_subspace}
\footnotesize
\begin{adjustbox}{max width=\textwidth}
\begin{tabular}{ccccc}
\toprule
Spine fraction & Transported-fresh overlap $\uparrow$ & MAP-fresh overlap $\uparrow$ & Mean angle (deg) $\downarrow$ & Mean eigenvalue drift $\downarrow$ \\
\midrule
0.00 & \meanstd{1.000}{0.000} & \meanstd{1.000}{0.000} & \meanstd{0.0}{0.0} & \meanstd{0.000}{0.000} \\
0.26 & \meanstd{0.705}{0.043} & \meanstd{0.710}{0.043} & \meanstd{31.6}{3.3} & \meanstd{0.080}{0.044} \\
0.51 & \meanstd{0.418}{0.035} & \meanstd{0.422}{0.036} & \meanstd{53.1}{3.0} & \meanstd{0.192}{0.038} \\
0.74 & \meanstd{0.250}{0.045} & \meanstd{0.252}{0.046} & \meanstd{66.3}{3.8} & \meanstd{0.266}{0.070} \\
1.00 & \meanstd{0.198}{0.021} & \meanstd{0.198}{0.021} & \meanstd{70.8}{1.8} & \meanstd{0.311}{0.076} \\
\bottomrule
\end{tabular}
\end{adjustbox}
\end{table}

\begin{figure}[H]
\centering
\includegraphics[width=0.70\textwidth]{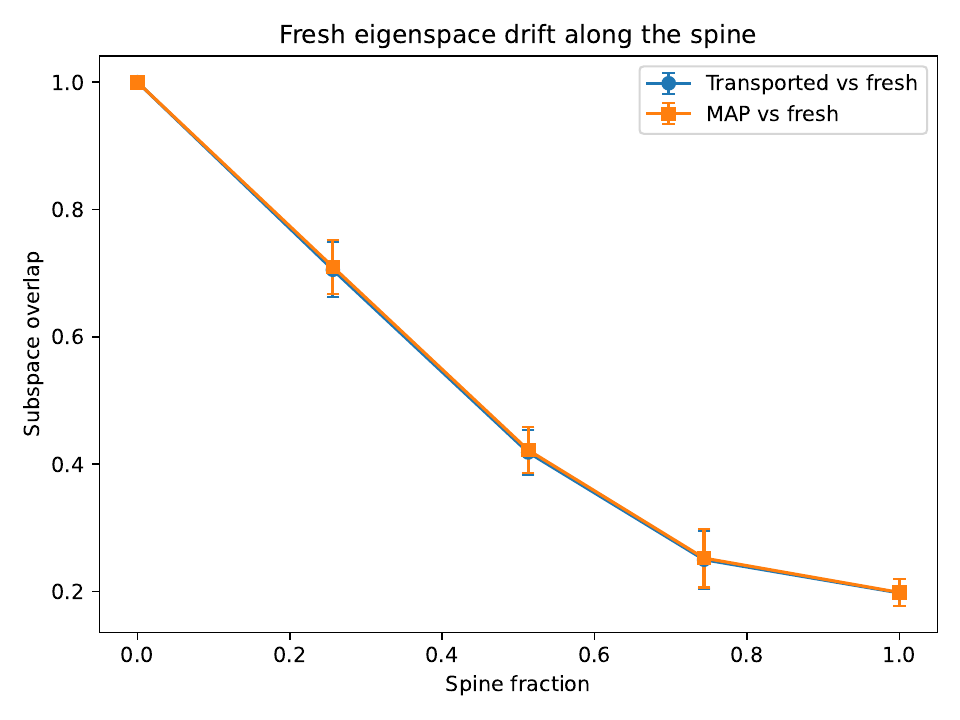}
\caption{Subspace overlap along the TRL spine. Transported and fixed-MAP bases both drift from the fresh local eigenspace as the spine moves away from the MAP.}
\label{fig:app_stale_subspace}
\end{figure}

Transported--fresh and MAP--fresh overlap are nearly identical throughout the path in Figure~\ref{fig:app_stale_subspace}; the transported basis is no closer to the fresh local eigenspace than the fixed MAP basis. The principal angles and eigenvalue drift likewise show the local stiff subspace rotating away from the MAP. Projection--QR is a transport heuristic rather than a local eigenspace tracker, so the observed gains cannot be attributed to accurate local curvature tracking. Section~\ref{app:fresh_refresh_single} tests the more expensive alternative of recomputing $Q_t^{\mathrm{fr}}$ at every anchor.

\subsection{Predictive effect of transported versus fixed-MAP basis}
To test whether the observed eigenspace rotation matters predictively, we compare two samplers on the same constructed spine: one uses the transported basis $Q_t^{\mathrm{tr}}$, and the other reuses the fixed MAP basis $Q_0$ at all anchors. They are tied within seed variability in Table~\ref{tab:app_basis_ablation}, while both variants substantially improve over deterministic MAP in the main CIFAR-100 results. Transporting the MAP basis provides no measurable predictive advantage over keeping it fixed in this benchmark, so the CIFAR-100 gains do not depend on accurate local curvature tracking. The next subsection tests fresh local recomputation at every spine anchor.

\begin{table}[H]
\centering
\caption{Mechanism diagnostic: predictive effect of transported transverse basis versus a fixed MAP basis. Mean $\pm$ standard deviation over three independently trained checkpoint-specific runs (training seeds 0--2). No unique winner is highlighted because the variants are tied within checkpoint-level variability.}
\label{tab:app_basis_ablation}
\footnotesize
\begin{adjustbox}{max width=0.78\textwidth}
\begin{tabular}{lcccc}
\toprule
Variant & Acc $\uparrow$ & NLL $\downarrow$ & ECE $\downarrow$ & Brier $\downarrow$ \\
\midrule
TRL-fixed-MAP-basis & \meanstd{0.743}{0.004} & \meanstd{0.955}{0.004} & \meanstd{0.015}{0.004} & \meanstd{0.356}{0.002} \\
TRL-transported-basis & \meanstd{0.745}{0.004} & \meanstd{0.956}{0.004} & \meanstd{0.016}{0.003} & \meanstd{0.355}{0.001} \\
\bottomrule
\end{tabular}
\end{adjustbox}

\end{table}

\subsection{Fresh-refresh and single-checkpoint checks}
\label{app:fresh_refresh_single}
The preceding ablation compares two inexpensive choices of transverse basis along the same constructed spine. We additionally test two endpoint alternatives. The first removes longitudinal mixing by using only the MAP checkpoint with the same transverse sampling recipe (TRL-single-checkpoint). The second recomputes a fresh local leading-curvature eigenspace $Q_t^{\mathrm{fr}}$ at every spine anchor and samples from those refreshed bases (TRL-fresh-refresh-basis). Neither this fresh control nor the fixed-MAP basis is reprojected onto the later tangent complement, whereas the transported frame is; these are sampling-frame sensitivity checks rather than clean one-factor curvature-refresh ablations. Appendix~\ref{app:additional_orthogonality} quantifies the distinction. This fresh-refresh variant is not part of the default TRL algorithm; it is a diagnostic upper-cost variant that tests whether local curvature refresh gives a qualitatively different posterior predictor.

\begin{table}[H]
\centering
\caption{Mechanism diagnostic: additional spine and curvature-refresh checks on CIFAR-100 with ResNet-18. Values are mean $\pm$ standard deviation over three independently trained checkpoint-specific runs (training seeds 0--2). The fresh-refresh variant recomputes a rank-30 eigenspace at every spine anchor before posterior prediction; the construction time is extra offline cost and is not part of the default TRL pipeline.}
\label{tab:app_fresh_refresh_single}
\footnotesize
\begin{adjustbox}{max width=\textwidth}
\begin{tabular}{lccccc}
\toprule
Variant & Acc $\uparrow$ & NLL $\downarrow$ & ECE $\downarrow$ & Brier $\downarrow$ & Extra refresh construction \\
\midrule
TRL-single-checkpoint & \meanstd{0.741}{0.004} & \meanstd{0.958}{0.005} & \meanstd{0.015}{0.003} & \meanstd{0.357}{0.002} & -- \\
TRL-fresh-refresh-basis & \meanstd{0.743}{0.003} & \meanstd{0.955}{0.008} & \meanstd{0.013}{0.003} & \meanstd{0.356}{0.002} & \meanstd{89.2}{7.7} min \\
\bottomrule
\end{tabular}
\end{adjustbox}
\end{table}

The single-checkpoint tube in Table~\ref{tab:app_fresh_refresh_single} already captures most of the clean CIFAR-100 improvement over deterministic MAP: its NLL and ECE lie in the same range as the full-spine transported and fixed-MAP variants in Table~\ref{tab:app_basis_ablation}. The fresh-basis replacement brings only a small numerical improvement under its stated sampling convention, while recomputing eigenspaces at all 40 stored spine locations, including the MAP anchor, costs $89.2\pm7.7$ minutes per seed as an additional offline step. Together with the modest functional drift in Table~\ref{tab:app_spine_functional_disagreement}, these results favor computing the transverse structure once at the MAP for this benchmark. Curvature refresh or longer spines may still be useful in other regimes.

\section{Small-data Fine-tuning Diagnostic}
\label{app:finetune_spine_diagnostic}

The main CIFAR-100 diagnostics above show that, for a large network trained from scratch, most of TRL's predictive gain is already captured by the MAP-centered transverse tube. This does not imply that the longitudinal component is always negligible. To test a more favorable non-toy regime, we run an exploratory small-data fine-tuning diagnostic: a ResNet-18-CIFAR checkpoint trained on CIFAR-100 is adapted to CIFAR-10 using 100 training examples per class and a disjoint 100-example-per-class validation split. The official CIFAR-10 test set is used only for final reporting. The configuration is selected as an existence diagnostic and is not a benchmark protocol: we fine-tune for 50 epochs with learning rate $0.003$, then run TRL with $T=20$ stored anchors, $k_\perp=30$, tube scale $0.1$, prior base $5$, non-head boost $c=50$, 25 posterior samples, and 25 FixBN batches per draw. The diagnostic fixes $\lambda_{\mathrm{base}}=5$, $c=50$, and $\beta_\perp=0.1$ as operating values; unlike the main CIFAR benchmark, they are not inherited from its marginal-likelihood fit or clean-validation tube-scale selection. Because the target dataset, fine-tuned checkpoint, active curvature, and projected prior differ, these numerical scales should not be compared directly with the main CIFAR values. Head precision is $5$ and non-head precision is $\max(50\cdot5,5)=250$, so the minimum-precision floor is inactive. Final test evaluation and functional-variance estimation for every stochastic posterior variant use $S=25$. In the full-spine variant, each draw samples one of all 20 stored anchors uniformly and then samples a transverse perturbation; in fixed-center variants, the anchor index is fixed.

We compare deterministic MAP fine-tuning, a MAP-centered single-checkpoint tube, the full uniform spine mixture, the endpoint-centered tube, and a validation-selected single spine anchor. The last control is a targeted center-selection diagnostic, not an exhaustive best-anchor oracle: validation selection evaluates a six-anchor subset of the 20 stored anchors using $S=8$ predictive samples per candidate, after which the selected anchor is frozen and reevaluated on the test set with $S=25$. The restricted candidate set and smaller selection-time Monte Carlo budget make this a weaker control than an exhaustive 20-anchor search. Even so, it separates two effects: whether the spine finds a better local center and whether uniform averaging over all stored anchors adds useful predictive dispersion.

\begin{table}[H]
\centering
\caption{Small-data CIFAR-100 $\rightarrow$ CIFAR-10 fine-tuning diagnostic. A ResNet-18-CIFAR checkpoint trained on CIFAR-100 is fine-tuned on 100 CIFAR-10 examples per class. Values are mean $\pm$ standard deviation over ten seeds. The single-anchor control selects from six of the 20 stored anchors on a disjoint validation subset using $S=8$, then freezes the selected anchor and evaluates it on the official test set using $S=25$; all reported stochastic metrics use $S=25$. Avg. func. var. measures variance across posterior predictive class-probability functions; it is a dispersion diagnostic for which larger values are not intrinsically better, and it is not applicable to deterministic MAP.}
\label{tab:app_finetune_spine_diagnostic}
\footnotesize
\begin{adjustbox}{max width=\textwidth}
\begin{tabular}{lccccc}
\toprule
Method & Acc $\uparrow$ & NLL $\downarrow$ & ECE $\downarrow$ & Brier $\downarrow$ & Avg. func. var. \\
\midrule
MAP-finetuned & \meanstd{0.777}{0.005} & \meanstd{0.649}{0.014} & \meanstd{0.018}{0.003} & \meanstd{0.315}{0.005} & -- \\
TRL-single-checkpoint & \meanstd{0.776}{0.005} & \meanstd{0.649}{0.014} & \meanstd{0.019}{0.003} & \meanstd{0.315}{0.005} & \meanstd{0.00078}{0.00010} \\
TRL-best-single-val & \meanstd{0.777}{0.005} & \meanstd{0.647}{0.015} & \meanstd{0.019}{0.004} & \meanstd{0.314}{0.006} & \meanstd{0.00094}{0.00008} \\
TRL-endpoint-single & \meanstd{0.778}{0.005} & \meanstd{0.647}{0.014} & \meanstd{0.019}{0.004} & \meanstd{0.314}{0.006} & \meanstd{0.00096}{0.00011} \\
TRL-full-spine & \meanstd{0.778}{0.005} & \meanstd{0.647}{0.015} & \meanstd{0.017}{0.003} & \meanstd{0.314}{0.006} & \meanstd{0.00230}{0.00043} \\
\bottomrule
\end{tabular}
\end{adjustbox}
\end{table}

Table~\ref{tab:app_finetune_spine_diagnostic} shows that this fine-tuning regime is qualitatively different from the from-scratch CIFAR-100 benchmark. Relative to the MAP-centered tube, full-spine mixing improves NLL and Brier in all ten seeds, ECE in eight, and average posterior functional variance in all ten. It does not consistently outperform the validation-selected anchor; seed 0 has slightly higher NLL and Brier under full-spine mixing. The NLL and Brier gains are small but consistently signed. Posterior functional variance rises from $7.76{\times}10^{-4}$ to $2.30{\times}10^{-3}$ on average, about a three-fold increase, showing that longitudinal mixing injects measurable functional variation in this small-data adaptation setting.

The single-anchor controls indicate that center relocation explains most of the NLL/Brier change: the endpoint and validation-selected tubes nearly match the full-spine mixture. Against the validation-selected single-anchor control, full-spine has lower NLL in seven of ten seeds and lower ECE in eight of ten, while Brier is essentially tied. The endpoint requires no search but is fixed by path length rather than by a predictive criterion, while validation selection requires an additional anchor search. Uniform mixing avoids either discrete choice and increases average functional variance to $2.30{\times}10^{-3}$, compared with roughly $0.9{\times}10^{-3}$ for the single-anchor controls, while retaining comparable NLL and Brier. Thus, in this regime, relocation explains most of the mean-score change, whereas mixing primarily preserves longitudinal predictive diversity. The contrast with from-scratch CIFAR-100 is clear: the MAP-identified transverse subspace dominates there, while small-data fine-tuning leaves stronger functional variation along the spine.

For this diagnostic, top-1 disagreement is the fraction of validation examples whose predicted class differs from the initial fine-tuned point, JS divergence is the Jensen--Shannon divergence between predictive probability vectors, and $\Delta_{\mathrm{CE}}$ is the validation cross-entropy change relative to the initial point. \emph{Mean along spine} averages this quantity over discrete spine anchors, whereas \emph{endpoint} reports the final spine anchor.

\begin{table}[H]
\centering
\caption{Longitudinal spine-signal diagnostic in the CIFAR-10 small-data fine-tuning regime. We compare deterministic predictions at discrete spine anchors with the initial fine-tuned point on the disjoint validation split, without transverse posterior samples. Values are mean $\pm$ standard deviation over ten seeds.}
\label{tab:app_finetune_spine_signal}
\footnotesize
\begin{adjustbox}{max width=0.82\textwidth}
\begin{tabular}{lc}
\toprule
Quantity & Value \\
\midrule
Mean top-1 disagreement along spine & \meanstd{0.0519}{0.0104} \\
Endpoint top-1 disagreement & \meanstd{0.0995}{0.0180} \\
Mean JS divergence along spine & \meanstd{0.0065}{0.0015} \\
Endpoint JS divergence & \meanstd{0.0175}{0.0036} \\
Mean $\Delta_{\mathrm{CE}}$ along spine & \meanstd{0.0191}{0.0087} \\
Endpoint $\Delta_{\mathrm{CE}}$ & \meanstd{0.0594}{0.0239} \\
\bottomrule
\end{tabular}
\end{adjustbox}
\end{table}

Figure~\ref{fig:regime_spine_dependence} in the main text summarizes this regime contrast. Tables~\ref{tab:app_spine_functional_disagreement}, \ref{tab:app_finetune_spine_diagnostic}, and~\ref{tab:app_finetune_spine_signal} provide the underlying values and separate the functional-drift diagnostic from posterior predictive performance.

We next ask whether the spine geometry is itself informative about when longitudinal mixing should matter. Across the two evaluated regimes, the functional-disagreement diagnostic from Appendix~\ref{app:transport} co-varies with the additional dispersion induced by full-spine mixing. We use it as a regime-level characterization of effective longitudinal variation, not as a scalar predictor of the performance gain from mixing. On CIFAR-100 from scratch (Table~\ref{tab:app_spine_functional_disagreement}), the transported spine stays functionally close to the MAP: at the endpoint it changes only $5.08\%\pm0.54\%$ of top-1 predictions, with endpoint JS $0.0063\pm0.0009$ and endpoint $\Delta_{\mathrm{CE}}$ $0.0104\pm0.0028$ over five seeds. In the CIFAR-10 small-data fine-tuning regime (Table~\ref{tab:app_finetune_spine_signal}), the corresponding endpoint signal is markedly larger: $9.95\%\pm1.80\%$ top-1 disagreement, endpoint JS $0.0175\pm0.0036$, and endpoint $\Delta_{\mathrm{CE}}$ $0.0594\pm0.0239$ over ten seeds. The same ordering holds when averaged over the spine: mean JS rises from $0.0028\pm0.0004$ to $0.0065\pm0.0015$. In the measured seeds, the seed-level ranges are separated on endpoint disagreement, endpoint JS, and endpoint $\Delta_{\mathrm{CE}}$; endpoint disagreement and JS are therefore roughly $2\times$ and $2.8\times$ larger under fine-tuning. In this terminology, the from-scratch CIFAR-100 corridor has nearly null effective longitudinal variation for prediction, whereas the fine-tuning corridor has a moderate one. This is also the regime in which full-spine mixing increases posterior functional variance and improves over the single-checkpoint posterior; on CIFAR-100 from scratch, the weak longitudinal signal is consistent with the near-redundancy of full-spine mixing and local transverse sampling.

The larger longitudinal signal does not by itself make the spine better: the fine-tuning trajectory also drifts more in validation loss. These are separate questions---whether the path stays in a low-loss corridor and whether movement along it changes predictions---so we report both quantities.

As an auxiliary diagnostic, we repeated the spine-signal measurement on the same $n=100$ fine-tuning split while varying the trainable subspace (three seeds). Endpoint JS divergence is near zero for head-only fine-tuning ($0.00020\pm0.00004$), modest when the last residual block and head are trained ($0.0045\pm0.0003$, despite freeing $\approx75\%$ of the parameters), and largest under full fine-tuning on the same three-seed subset ($0.0198\pm0.0028$; the corresponding ten-seed estimate is reported in Table~\ref{tab:app_finetune_spine_signal}). The longitudinal signal is not explained by trainable-parameter count alone; it depends on which parts of the representation are allowed to move. This auxiliary sweep characterizes spine geometry; predictive performance is not evaluated. Endpoint $\Delta_{\mathrm{CE}}$ rises together with JS across the three modes ($-0.0045$, $0.0013$, and $0.0685$), so the stronger longitudinal signal also comes with larger validation-loss drift. Because we did not run the full posterior comparison of Table~\ref{tab:app_finetune_spine_diagnostic} in the restricted-subspace modes, their predictive effect remains unresolved.

\section{ImageNet Scale-check}
\label{app:imagenet_scale_check}

To test scalability, we evaluate a pretrained ResNet-50 on ImageNet under three questions: whether the transverse tube calibrates a large model, whether the spine adds functional diversity, and whether rank $k_\perp=30$ remains sufficient. We treat this experiment as a scale check rather than a standalone benchmark.

\paragraph{Protocol.}
We use the torchvision ResNet-50 \texttt{IMAGENET1K\_V1} checkpoint as a fixed pretrained deterministic predictor, not as an exact MAP solution under TRL's two-block prior. Because the checkpoint was pretrained on the official ImageNet training set, a split carved from that set would not be out-of-sample with respect to the checkpoint. We follow the standard post-hoc calibration convention and split the official ImageNet validation set into a 25k tuning half and a 25k held-out evaluation half~\citep{guo2017calibration}. Hyperparameters are selected by NLL on the tuning half and reported only on the held-out half. The MAP gate gives full-validation accuracy $0.7615$, matching the canonical torchvision accuracy $0.7613$. The MAP accuracy $0.7611$ in Table~\ref{tab:imagenet_rank} is measured on the held-out half used by the paired rank diagnostic and is therefore not directly comparable to those full-validation values.

Unless otherwise stated, we use $S=25$ posterior predictive samples, FixBN with 25 training batches per sample, a rank-$30$ transverse subspace, and the same two-block prior used in the CIFAR experiments: the classifier head uses the fitted base precision $\lambda_{\rm base}$, while the backbone uses $\max(c\lambda_{\rm base},5)$. For seed 0, $\lambda_{\rm base}=12.5651$. The rank-$30$ basis is obtained by Lanczos with 90 iterations and 5 HVP batches; its leading-to-trailing Ritz spectrum is $348.6$ to $96.2$. For the single-checkpoint experiment in Table~\ref{tab:imagenet_single_checkpoint}, hyperparameter selection uses the grid $c\in\{50,150,450\}$ and $\beta_\perp\in\{0.5,1,1.5,2,3,4\}$. The rank ablation below fixes $c=50$ and sweeps $\beta_\perp\in\{2,3,4\}$. This narrower grid is anchored at the selected rank-$30$ operating point: for seed 0, the broader single-checkpoint grid already selected $\beta_\perp=2$, so the rank sweep tests whether adding more transverse directions helps at the calibrated scale without a separate low-temperature re-optimization for each rank.

\paragraph{Single-checkpoint transverse tube.}
Table~\ref{tab:imagenet_single_checkpoint} reports the single-checkpoint transverse variant on ten random splits, using only the tube around the fixed checkpoint and no longitudinal spine. TRL reduces ECE in all ten held-out halves of the official validation set, while NLL and accuracy incur small, consistent costs.

\begin{table}[H]
\centering
\caption{ImageNet/ResNet-50 single-checkpoint transverse TRL over ten random splits. Hyperparameters are selected by NLL on the tuning half of the official validation set and reported on the held-out half. $\Delta=\mathrm{TRL}-\mathrm{checkpoint}$. Seeds 0--2 are the original scale-check; seeds 3--9 are the extension under the same grid and protocol. Relative ECE reduction is computed per split as $(\mathrm{ECE}_{\mathrm{ckpt}}-\mathrm{ECE}_{\mathrm{TRL}})/\mathrm{ECE}_{\mathrm{ckpt}}$ and then averaged over the ten splits; the reported $32.7\%\pm9.8\%$ is the mean $\pm$ sample standard deviation of those per-split ratios, computed from unrounded per-split ECE values, not a ratio of averaged ECE values.}
\label{tab:imagenet_single_checkpoint}
\scriptsize
\begin{adjustbox}{max width=\textwidth}
\begin{tabular}{ccccccc}
\toprule
Seed & Selected $(c,\beta_\perp)$ & TRL ECE & Checkpoint ECE & $\Delta$ECE & $\Delta$NLL & $\Delta$Acc \\
\midrule
0 & $(50,2)$   & 0.0207 & 0.0383 & $-0.0176$ & $+0.0066$ & $-0.0043$ \\
1 & $(50,2)$   & 0.0218 & 0.0377 & $-0.0159$ & $+0.0070$ & $-0.0040$ \\
2 & $(50,1.5)$ & 0.0258 & 0.0361 & $-0.0102$ & $+0.0011$ & $-0.0007$ \\
3 & $(150,3)$  & 0.0252 & 0.0408 & $-0.0156$ & $+0.0107$ & $-0.0028$ \\
4 & $(450,4)$  & 0.0250 & 0.0382 & $-0.0132$ & $+0.0050$ & $-0.0044$ \\
5 & $(50,1.5)$ & 0.0303 & 0.0370 & $-0.0067$ & $+0.0074$ & $-0.0028$ \\
6 & $(150,2)$  & 0.0287 & 0.0367 & $-0.0080$ & $+0.0050$ & $-0.0008$ \\
7 & $(150,3)$  & 0.0234 & 0.0361 & $-0.0127$ & $+0.0027$ & $-0.0026$ \\
8 & $(150,2)$  & 0.0316 & 0.0400 & $-0.0084$ & $+0.0060$ & $-0.0013$ \\
9 & $(150,3)$  & 0.0210 & 0.0359 & $-0.0148$ & $+0.0055$ & $-0.0043$ \\
\midrule
Mean & -- & 0.0254 & 0.0377 & $-0.0123$ & $+0.0057$ & $-0.0028$ \\
Std. & -- & 0.0038 & 0.0017 & 0.0038 & 0.0026 & 0.0015 \\
\bottomrule
\end{tabular}
\end{adjustbox}
\end{table}

The mean relative ECE reduction, computed per split and then averaged, is $32.7\%\pm9.8\%$ (mean $\pm$ sample standard deviation across the ten per-split ratios). Because NLL and accuracy are slightly worse in every aggregate comparison, the pattern is a calibration--accuracy trade-off rather than a global predictive improvement. These splits quantify sensitivity to the calibration/evaluation partition; they are not independent model or dataset replicates. The selected configurations remain on the low-NLL plateau of the tuning grid; the larger values of $c$ selected in several extension splits reflect split-specific posterior scaling, with no change to the structural protocol.

\paragraph{Temperature-scaling control.}
We additionally fit one scalar temperature per split by minimizing NLL on the same tuning half and evaluate it once on the held-out half. Table~\ref{tab:imagenet_map_ts} reports the ten-split result. Temperature scaling preserves the checkpoint argmax and improves all three probabilistic metrics: NLL falls by $0.0131$, ECE by $0.0164$, and Brier by $0.0011$. Its mean temperature $\tau=1.1297>1$ indicates moderate checkpoint overconfidence.

\begin{table}[H]
\centering
\caption{ImageNet/ResNet-50 temperature scaling on the same ten random tuning/evaluation splits. Temperature is selected by tuning-half NLL and frozen before held-out evaluation.}
\label{tab:imagenet_map_ts}
\footnotesize
\begin{adjustbox}{max width=\textwidth}
\begin{tabular}{lccccc}
\toprule
Method & Acc $\uparrow$ & NLL $\downarrow$ & ECE $\downarrow$ & Brier $\downarrow$ & Temperature \\
\midrule
Checkpoint & \meanstd{0.760592}{0.001540} & \meanstd{0.964229}{0.006882} & \meanstd{0.037676}{0.001690} & \meanstd{0.336113}{0.001582} & -- \\
Checkpoint+TS & \meanstd{0.760592}{0.001540} & \meanstd{0.951139}{0.006371} & \meanstd{0.021269}{0.001705} & \meanstd{0.335003}{0.001472} & \meanstd{1.129749}{0.004098} \\
\bottomrule
\end{tabular}
\end{adjustbox}
\end{table}

MAP+TS is the stronger scalar calibrator in this pretrained-checkpoint regime: it preserves accuracy, improves NLL, and attains lower mean ECE than TRL ($0.0213$ versus $0.0254$). The outputs are nevertheless different. Temperature scaling returns one deterministic rescaled predictive distribution; TRL returns parameter-space posterior samples, nonzero functional variance, and the transverse/longitudinal support diagnostics studied below. ImageNet serves here as a scale and mechanism check; when scalar held-out calibration is the only objective, MAP+TS is the better choice.

\paragraph{Seed-count note.}
The single-checkpoint and temperature-scaling results above use ten random splits. The predictor--corrector-step and rank sweeps below remain seed-0 diagnostics: they are paired dose-response tests within a fixed split and are not new multi-seed benchmarks. We interpret them through their within-run trends while using the ten-split spread in Table~\ref{tab:imagenet_single_checkpoint} as context for the scale of split-to-split variation.

\paragraph{Longitudinal spine dose-response.}
We next test whether the constructed longitudinal spine contributes additional functional diversity around this pretrained checkpoint. We construct the spine using the same mechanics as in the CIFAR experiments: a combined predictor--corrector step $\Delta s\,v_\parallel-\eta_{\rm corr}(I-v_\parallel v_\parallel^\top)\nabla_\theta\ell_{\mathcal B}$, norm control on the combined displacement, tangent update from the realized displacement, rotating correction batches, and transported transverse bases. This ImageNet diagnostic is parameterized by the number $n_{\mathrm{step}}$ of predictor--corrector transitions, not by the CIFAR convention in which $T$ counts stored anchors. We sweep the nominal path budget $n_{\mathrm{step}}\Delta s$ from 0 to 0.48 while keeping $c=50$, $\beta_\perp=2$, the split, the basis, and $S=25$ fixed. With every transition retained, $n_{\mathrm{step}}=0$ yields only the MAP anchor and $n_{\mathrm{step}}>0$ yields $n_{\mathrm{step}}+1$ stored anchors.

Across the predictor--corrector sweep in Table~\ref{tab:imagenet_spine_length}, the longitudinal component is effectively inert: posterior functional variance, NLL, ECE, and Brier remain flat across step count and nominal path budget and are indistinguishable from the single-checkpoint tube. The sweep spans nominal path lengths from $0$ to $0.48$, bracketing the CIFAR retained-anchor length $(T-1)\Delta s=0.39$. The matched-step-size setting $\Delta s=0.01$ reaches $0.24$, whereas the $0.48$ endpoint uses $\Delta s=0.02$; the diagnostic therefore matches the path-length magnitude but not the full CIFAR discretization. The construction is numerically healthy as well: held-out diagnostic loss remains flat along the spine, and the norm-control threshold is not activated.

\begin{table}[H]
\centering
\caption{ImageNet/ResNet-50 predictor--corrector-step sweep, seed 0. The tuning half is used with $c=50$, $\beta_\perp=2$, and $S=25$. Here $n_{\mathrm{step}}$ counts transitions: $n_{\mathrm{step}}=0$ is the single-checkpoint tube with only the MAP anchor, while retaining every transition gives $n_{\mathrm{step}}+1$ stored anchors.}
\label{tab:imagenet_spine_length}
\footnotesize
\begin{adjustbox}{max width=\textwidth}
\begin{tabular}{ccccccc}
\toprule
Nominal length $n_{\mathrm{step}}\Delta s$ & Transitions $n_{\mathrm{step}}$ & $\Delta s$ & Func. var. trace & NLL & ECE & Brier \\
\midrule
0.00 & 0  & --   & 0.01380 & 0.9635 & 0.0200 & 0.3372 \\
0.08 & 8  & 0.01 & 0.01373 & 0.9635 & 0.0204 & 0.3372 \\
0.24 & 24 & 0.01 & 0.01362 & 0.9638 & 0.0210 & 0.3372 \\
0.48 & 24 & 0.02 & 0.01372 & 0.9647 & 0.0212 & 0.3374 \\
\bottomrule
\end{tabular}
\end{adjustbox}
\end{table}

The held-out half gives the same conclusion: nominal lengths $0.08$, $0.24$, and $0.48$ yield NLL $0.9705$, $0.9709$, and $0.9705$, respectively, with ECE in the range $0.0195$--$0.0207$. Flatness is assessed within this realization. The spine-free reference is the $n_{\mathrm{step}}=0$ row of Table~\ref{tab:imagenet_spine_length}, evaluated on the tuning half within that table's own realization. We do not compare the held-out sweep against the separately constructed rank-$30$ record in Table~\ref{tab:imagenet_rank}. The same near-redundancy of longitudinal mixing appears in CIFAR-100 from-scratch training. In both cases, the deeply trained checkpoint sits in a corridor that behaves mostly as a near-invariance and adds little functional diversity; the fine-tuning regimes of Appendix~\ref{app:finetune_spine_diagnostic} show the contrasting case in which nearby anchors are more functionally distinct.

\paragraph{Transverse rank.}
Finally, Table~\ref{tab:imagenet_rank} tests whether the default $k_\perp=30$ is underdimensioned for ImageNet/ResNet-50. We compare ranks $30$, $50$, and $100$ on seed 0, keeping $c=50$ fixed and selecting $\beta_\perp\in\{2,3,4\}$ by tuning NLL. Each rank uses a fresh Lanczos basis, with iterations scaled as $3\times k_\perp$. The rank-$100$ spectrum remains nontrivial, with trailing Ritz value $38.9$, so the higher-rank test is not limited by a collapsed Lanczos tail. The rank-$30$ functional-variance trace ($0.0160$) is not a duplicate of the $n_{\mathrm{step}}=0$ value in Table~\ref{tab:imagenet_spine_length} ($0.01380$): the rank table evaluates the held-out half with a freshly computed Lanczos basis and independent Monte Carlo draws, whereas the spine-length table reports the tuning-half diagnostic with its fixed basis. The numerical difference therefore reflects distinct diagnostic realizations, not inconsistent evaluation of one stored posterior.

\begin{table}[H]
\centering
\caption{ImageNet/ResNet-50 transverse rank ablation, seed 0. Hyperparameters are selected by NLL on the tuning half and evaluated on the held-out half. The higher-rank rows report larger functional variance without a material held-out improvement in this split; the rank comparison is not replicated across splits.}
\label{tab:imagenet_rank}
\footnotesize
\begin{adjustbox}{max width=0.86\textwidth}
\begin{tabular}{cccccc}
\toprule
Variant & Selected $\beta_\perp$ & NLL & ECE & Acc. & Func. var. trace \\
\midrule
MAP      & -- & 0.9643 & 0.0383 & 0.7611 & -- \\
Rank 30  & 2  & 0.9710 & 0.0207 & 0.7568 & 0.0160 \\
Rank 50  & 2  & 0.9731 & 0.0183 & 0.7554 & 0.0203 \\
Rank 100 & 2  & 0.9721 & 0.0210 & 0.7552 & 0.0292 \\
\bottomrule
\end{tabular}
\end{adjustbox}
\end{table}

Ranks $50$ and $100$ do not materially improve the held-out metrics relative to rank $30$. Rank $50$ lowers ECE by $0.0024$ in this seed, but this within-split change is not replicated across splits, while NLL and accuracy worsen. Rank $100$ likewise leaves NLL/ECE/accuracy effectively unchanged. The higher-rank rows also report larger functional variance, but this single-split pattern is descriptive and does not establish a general rank--variance law. In this split, adding directions beyond $k_\perp=30$ brings no robust predictive gain, consistent with the rank saturation observed on CIFAR-100.

Taken together, the ImageNet experiments show that TRL remains operational at ResNet-50 scale: the transverse tube reduces ECE in all ten splits, the longitudinal spine is effectively redundant around this checkpoint, and increasing the transverse rank beyond $30$ gives no robust gain. The comparison with temperature scaling also separates posterior functionality from calibration alone. ImageNet therefore serves primarily as a scale and mechanism check; the corresponding storage and construction costs are discussed in Appendix~\ref{app:complexity_discussion}.

\section{Full-network support selection versus last-layer and all-weight low-rank Laplace baselines}
\label{app:trl_vs_ela_support}

\subsection{What is being compared to last-layer and all-weight low-rank Laplace?}

At ResNet scale, all compared posterior approximations make a tractability cut. Last-layer ELA/LLA restrict uncertainty to the classifier; FM-LR-LLA ValScale uses a rank-$30$ all-weight subspace chosen to preserve a specified linearized predictive covariance; TRL instead uses stochastic HVPs to select a rank-$30$ full-network transverse support, couples it to a discrete predictor--corrector spine, and evaluates nonlinear sampled networks. The FM comparison is checkpoint-, architecture-, split-, and rank-matched but not prior-matched, so it tests alternative support-allocation strategies; it is not a one-factor subspace ablation. Appendix~\ref{app:faller_martin_comparison} gives the full protocol.

\subsection{Support-selection diagnostics}

TRL uses curvature primarily to choose where low-rank transverse uncertainty is placed, while longitudinal support is restricted to stored spine anchors. The sine corrector does not enforce a loss-acceptance test (Appendix~\ref{app:toy_corrector_audit}). The remaining parameter directions stay at their MAP values, so the method is a support-restricted full-network approximation and does not represent a complete posterior over all flat directions. Appendices~\ref{app:transport} and~\ref{app:finetune_spine_diagnostic} separately test when longitudinal mixing contributes; the control below asks whether full-network low-rank access alone explains the CIFAR gain.

As an additional full-network control, we evaluated random rank-$30$ posteriors using the same two-block prior family and validation-scale grid, $S=25$ predictive samples, and the same FixBN protocol as TRL. On a fixed MAP checkpoint, five independent random bases produced nearly identical, MAP-like predictions (Acc $0.7413 \pm 0.0001$, NLL $1.0378 \pm 0.0002$, ECE $0.0951 \pm 0.0001$, versus the same checkpoint's MAP values of $0.7399$, $1.0498$, and $0.0996$). One-basis controls on two additional MAP checkpoints showed the same MAP-like behavior, and all seven random controls selected the largest value in the original scale grid, $\beta_\perp = 20$ (overall Acc $0.7418 \pm 0.0008$, NLL $1.0331 \pm 0.0076$, ECE $0.0925 \pm 0.0041$, AUROC $0.8647 \pm 0.0128$). The predictive-entropy AUROC range overlaps the MAP and TRL ranges in Table~\ref{tab:cifar100_clean_5seed}, consistent with AUROC being less sensitive than NLL/ECE to the calibration mechanism isolated by this control.

Extending one basis to $\beta_\perp = 40$ improved validation NLL only modestly (from $1.016$ at $\beta_\perp = 2$ to $1.000$ at $\beta_\perp = 40$, a gain of $0.016$ across a $20\times$ scale range) and left the predictor MAP-like on test data (Acc $0.742$, NLL $1.023$, ECE $0.089$). In contrast, the curvature-selected TRL subspace reaches its calibrated operating regime at $\beta_\perp = 4$ and collapses at $\beta_\perp = 20$ (Table~\ref{tab:app_ablation_tube}), showing that TRL is sensitive to scale because its selected directions are functionally loaded. These historical controls do not isolate orientation: the original grid ended at 20, and a later cache audit found a different prior magnitude for the checkpoint-0 random controls. They share the prior family, not its numerical precision. Appendix~\ref{app:orientation_checkpoint} instead matches the projected prior and extends the common scale grid before comparing K with RB.

\begin{table}[H]
\centering
\caption{Support-selection scale diagnostic. The seven random controls comprise five independent random bases on one fixed MAP checkpoint plus one basis on each of two additional checkpoints; their aggregate is descriptive and mixes basis and checkpoint variation. The curvature-selected TRL row reports three posterior-sampling seeds on one fixed MAP checkpoint, so its uncertainty is not directly comparable to the seven-control aggregate. The representative extension is one random basis evaluated beyond the original scale grid. These historical variants use the same two-block prior family, validation-scale selection, $S=25$, and FixBN protocol; their numerical priors are not fully matched. See Appendix~\ref{app:orientation_checkpoint} for the separate matched-prior comparison.}
\label{tab:main_random_scale_diagnostic}
\scriptsize
\setlength{\tabcolsep}{4pt}
\renewcommand{\arraystretch}{1.08}
\begin{adjustbox}{max width=\textwidth}
\begin{tabular}{p{0.19\textwidth}p{0.18\textwidth}p{0.28\textwidth}p{0.28\textwidth}}
\toprule
Support & Scale protocol & Validation behavior & Test behavior \\
\midrule
Random rank-$30$, seven controls & Original grid $\{2,3,4,6,10,20\}$ & All seven select the upper boundary, $\beta_\perp=20$ & MAP-like overall: Acc $0.7418\pm0.0008$, NLL $1.0331\pm0.0076$, ECE $0.0925\pm0.0041$ \\
Random rank-$30$, representative extension & $\beta_\perp:2\rightarrow40$ & Validation NLL changes only $1.016\rightarrow1.000$ across a $20\times$ scale range & At $\beta_\perp=40$: Acc $0.742$, NLL $1.023$, ECE $0.089$ \\
Curvature-selected TRL & Original grid $\{2,3,4,6,10,20\}$ & Interior optimum at $\beta_\perp=4$; high scales are rejected & Three sampling seeds on one fixed checkpoint: at $4$, Acc $0.743\pm0.003$, NLL $0.955\pm0.006$, ECE $0.014\pm0.002$; at $20$, Acc $0.387\pm0.022$, NLL $2.551\pm0.056$, ECE $0.195\pm0.016$ \\
\bottomrule
\end{tabular}
\end{adjustbox}
\end{table}

Table~\ref{tab:main_random_scale_diagnostic} describes the historical scale ranges shown, not the best achievable random-support scores. The matched-prior experiment in Table~\ref{tab:orientation_checkpoint} gives substantially improved random-support NLL at larger selected scales and a heterogeneous ECE advantage for K. The historical results alone do not establish a one-factor orientation effect.

For the five-basis random-subspace control and the seed-0 FM-LR-LLA diagnostic, both pipelines load the same byte-identical CIFAR-100 MAP seed-0 checkpoint, as verified by the release manifest. The comparison is therefore checkpoint-matched. On this checkpoint, MAP gives Acc/NLL/ECE $0.7399/1.0498/0.0996$; the five-basis random rank-$30$ control gives $0.7413/1.0378/0.0951$ with Brier $0.3694$; FM-LR-LLA gives $0.7410/0.9913/0.0463$ with Brier $0.3582$; and the seed-0 TRL result reported in Table~\ref{tab:cifar100_clean_5seed} gives $0.743200/0.959857/0.014815$ with Brier $0.355263$. The random support remains close to MAP, whereas the informed low-rank constructions produce substantially larger probabilistic corrections. Within this checkpoint-matched comparison, full-network access and rank $30$ are not sufficient to reproduce the TRL correction. It is not, however, a one-factor ablation: FM-LR-LLA uses an all-weight Kronecker Laplace reference, a marginal-likelihood prior, and MC-linearized predictions, whereas the random and TRL controls use TRL's two-block prior, FixBN, and nonlinear network evaluation. Accordingly, the remaining FM--TRL gap should not be attributed solely to subspace selection.

Table~\ref{tab:diagnostic_controls} summarizes the corresponding diagnostic controls and the failure modes they isolate.

\begin{table}[H]
\centering
\caption{Diagnostic controls of the practical TRL approximation. Random-support statements below concern the historical restricted-grid runs; Appendix~\ref{app:orientation_checkpoint} gives the subsequent matched-prior, wider-grid results.}
\label{tab:diagnostic_controls}
\small
\renewcommand{\arraystretch}{1.14}
\setlength{\tabcolsep}{4pt}
\begin{adjustbox}{max width=\textwidth}
\begin{tabular}{p{0.31\columnwidth}p{0.62\columnwidth}}
\toprule
Question & Control \\
\midrule
Full-network access? & Random rank-$30$ full-network bases remain MAP-like; curvature selection matters (Appendix~\ref{app:trl_vs_ela_support}). \\
Spine as main CIFAR mechanism? & TRL-single-checkpoint nearly matches full TRL on CIFAR-100, isolating the MAP-identified transverse subspace (Table~\ref{tab:app_fresh_refresh_single}). \\
Transport fidelity? & Fresh-refresh diagnostics show eigenspace drift, but small predictive gains relative to cost (Appendix~\ref{app:transport}). \\
Curvature-operator confound? & A seed-0 replacement of the positive-spectrum Hessian selector by exact GGN selection yields nearly identical downstream metrics; this checks predictive robustness, not eigenspace identity (Appendix~\ref{app:practical_algorithm}). \\
BatchNorm-statistics confound? & The rolling-mode sweep (Table~\ref{tab:app_ablation_fixbn}, a joint batch-count/carry-over sensitivity) shows that probabilistic metrics change as recalibration budget and carry-over jointly vary; random full-network controls evaluated with the same FixBN protocol remain MAP-like, showing that FixBN alone does not explain TRL's calibration gain (Appendix~\ref{app:trl_vs_ela_support}). \\
CIFAR-100-C tuning? & Tube scale is selected on clean validation data and reused unchanged for corrupted-test evaluation. \\
\bottomrule
\end{tabular}
\end{adjustbox}
\end{table}

\section{Computational Complexity and Sampling Efficiency}
\label{app:complexity_discussion}

Geodesic-based Riemannian Laplace methods and Laplace diffusion generate samples by numerically integrating geometric dynamics. With $T_{\rm int}$ integration steps per draw and per-step cost $C_{\rm step}$, generating $S$ samples costs $O(S\,T_{\rm int}\,C_{\rm step})$. The step count and cost depend on the method, metric representation, and integrator; this expression describes the sampling procedure and is not a measured runtime comparison.

TRL constructs its finite spine and low-rank transverse structure once per MAP checkpoint. Posterior prediction then samples a spine anchor and applies a low-rank map $\theta^{(r)}=\gamma_t+N_t(\beta_\perp L_{\perp,t}z_\perp^{(r)})$, followed by ordinary forward passes and, for BatchNorm networks, FixBN recalibration. Thus TRL trades additional offline storage and construction time for avoiding a fresh geodesic or curvature solve for every posterior sample.

For ResNet-18, an fp32 rank-30 frame occupies approximately 1.35 GB, so caching 40 transported frames takes about 54 GB of host RAM, excluding anchor vectors and workspaces. CPU staging lowers simultaneous GPU memory demand rather than this host-memory requirement. A single fixed-MAP frame can reduce storage, but changes later tangent orthogonality; its historical predictive comparison is Table~\ref{tab:app_basis_ablation}. The 89.2-minute fresh-curvature cost is the additional diagnostic with 40 decompositions, whereas standard TRL estimates retained curvature once at the MAP.

Because the logged timers cover different scopes and the TRL total contains overlapping nested FixBN timers, Table~\ref{tab:app_runtime_memory} cannot support an end-to-end wall-clock ranking. The robust comparison is computational structure: TRL uses one trained backbone plus post-hoc geometric construction, validation-scale selection, and posterior prediction, whereas the reported Deep Ensemble uses $M=5$ independently optimized backbones. The demonstrated advantage here is fewer independent optimization runs; runtime superiority over Deep Ensembles is not established.

\paragraph{ImageNet-scale cost.}
The ImageNet/ResNet-50 scale-check in Appendix~\ref{app:imagenet_scale_check} follows the same amortized cost profile at larger scale. The MAP checkpoint is fixed; the additional offline cost comes from HVP/Lanczos construction of the transverse basis and, for spine diagnostics, from storing or staging transported bases. For ResNet-50, a rank-$30$ transverse basis has roughly $30\times25.6$M floating-point entries, about $3.1$GB in fp32, while the rank-$100$ ablation increases this to about $10.2$GB. At high rank, the Lanczos workspace, not the final basis alone, is the main memory bottleneck, so the rank-$100$ diagnostic stages the basis on CPU. For full-spine diagnostics, storing a transported basis at every anchor would multiply this cost by the number of anchors; we therefore stage anchor bases on CPU and move the selected basis to GPU only when drawing a posterior sample. Thus the ImageNet experiment preserves the intended TRL trade-off: one offline geometric construction around a fixed MAP checkpoint, followed by ordinary posterior forward passes and FixBN recalibration, with no training of multiple independent models.

\section{Limitations and Additional Discussion}
\label{app:limitations}

TRL is a post-hoc, single-MAP approximation. Uniform anchor weights and validation tempering do not recover exact posterior mass, and the anchors are not stationary points of a common full-data posterior objective. It does not capture the cross-basin diversity of Deep Ensembles; its OOD AUROC can therefore be lower than that of methods with more diffuse function-space uncertainty. It also has offline and prediction-time overhead: spine construction requires HVP/Lanczos operations, and BatchNorm posterior prediction requires FixBN unless an alternative normalization or cached-statistics strategy is used. For architectures without running normalization statistics, including LayerNorm-based Vision Transformers, this FixBN step is unnecessary, although curvature construction and nonlinear posterior-sample evaluation remain. The current implementation uses a one-dimensional spine, a low-rank transverse subspace, and a validation-selected tube scale $\beta_\perp$ as an explicit calibration/tempering parameter. Transported low-rank structure can drift from freshly recomputed local eigenspaces, so exact local geometric fidelity should not be assumed. We leave higher-dimensional spines, periodic curvature refresh, normalization without per-sample recalibration, and cross-basin mixtures to future work. VaLLA \citep{ortega2024valla}, accelerated LLA (ELLA) \citep{deng2022ella} and fixed-mean Gaussian processes (FMGP) \citep{ortega2024fmgp} remain relevant untested alternatives; the evaluated all-weight rank-matched reference does not substitute for these comparisons. Curvature-based transverse selection prioritizes directions that are data-sensitive under the training curvature proxy. Directions that are weakly identified by the training data but affect predictions away from them are not explicitly targeted by the present low-rank construction.

\input{appendix_impact_statement}

\section{Extended Related Work and Positioning}
\label{app:extended_related}

Riemannian Laplace maps local Gaussians along geodesics \citep{bergamin2023riemannian,yu2024fisherlaplace}, whereas Laplace diffusion uses metric-dependent random walks \citep{roy2024reparameterization}. TRL constructs its posterior approximation in two stages: minibatch gradients guide the predictor--corrector spine, defining its anchors, and posterior samples are then generated by selecting an anchor and drawing a transverse Gaussian perturbation. Reparameterization invariance is not enforced; Appendix~\ref{app:transport} measures local eigenspace drift.

The continuous idealization is related to the geometric-Gaussian design space of \citet{dacosta2025geometric}, but the implemented piecewise spine is not a global smooth diffeomorphic pushforward and inherits no ReparamGA universality guarantee. Function-space methods such as FSP-Laplace \citep{cinquin2024fsp} and functional variational BNNs \citep{sun2019functional} instead impose function-space prior structure during training; TRL is post hoc and does not provide their explicit far-from-data prior-reversion guarantees.

Low-rank subspace methods differ mainly in what the retained rank is asked to preserve. \citet{faller2026lowrank} target the predictive covariance of a specified full linearized GGN--Laplace reference; TRL chooses parameter-space support and evaluates nonlinear sampled networks, with the matched FM-LR-LLA ValScale comparison in Appendix~\ref{app:faller_martin_comparison}. \citet{izmailov2020subspace} compare random, SGD-PCA, and curve subspaces; our random rank-$30$ basis is only a control, while TRL uses HVP-selected transverse support plus discrete anchor mixing. Mode-connectivity methods connect independently trained minima \citep{garipov2018loss,draxler2018essentially}, whereas TRL starts from one MAP with no prescribed endpoint. Loss-tunnel MCMC \citep{dold2024embedding} also targets posterior connectivity but requires a learned tunnel and a prediction-time chain; TRL builds a finite spine with minibatch gradients and then samples from its Gaussian fibers.

\input{appendix_additional_diagnostics}

\begin{samepage}
\section{AI Use Statement}
\label{app:llm_usage}

Generative AI tools were used for manuscript drafting and language editing, consistency checking, conceptual and methodological feedback, experimental-design feedback, interpretation of experimental results, assistance in implementing research methods and editing/reviewing associated software code, and creation or modification of scientific figures and release artifacts. All AI-assisted text, code, figures, analyses, and artifacts were reviewed and validated by the author against the manuscript, source code, experiment logs, and released artifacts. The author takes responsibility for the final content of this work, including text, claims, code, figures, and artifacts produced with AI assistance.

\end{samepage}
\end{document}

%% file: appendix_geometry_idealization.tex
\section{Geometric Motivation and Continuous TRL Idealization}
\label{app:geometry_idealization}

In deep networks, the geometry of $\mathcal L(\theta)$ is characterised by two key features:

\begin{enumerate}[leftmargin=*, nosep]
    \item \textbf{Low-loss valleys.} Deep networks can exhibit several sources of weakly constrained motion. Continuous functional symmetries, such as positive-homogeneous rescalings, can generate locally flat tangent directions along which the represented function changes little. Discrete symmetries such as neuron permutations instead produce functionally equivalent parameter configurations but do not, by themselves, define a continuous tangent direction between them \citep{kristiadi2023geometry}. Approximate mode connectivity is distinct: low-loss paths can connect independently trained solutions that are functionally close but not identical, so motion along them can correspond to genuinely different predictors of comparable loss \citep{garipov2018loss,draxler2018essentially}. A low-loss valley can mix continuous near-invariances with connectivity directions, with relative importance depending on architecture and dataset.
    
    \item \textbf{Extreme anisotropy of curvature.} The loss is extremely flat along approximate invariance and connectivity directions and steep along data-sensitive directions. The Hessian spectrum thus exhibits a wide range of eigenvalues, with a bulk of near-zero values \citep{sagun2017empirical,maddox2019simple}.
\end{enumerate}

This distinction matters for what a longitudinal posterior construction can be expected to contribute. TRL does not assume that every longitudinal direction produces predictive diversity. It builds a discrete predictor--corrector spine to cover the low-loss corridor around the MAP. Appendix~\ref{app:transport} measures anchor loss for the CIFAR constructions studied there. This does not extend to every implementation: the released sine paths violate near-MAP loss preservation (Appendix~\ref{app:toy_corrector_audit}). Transverse Gaussian perturbations are then sampled around those anchors using a curvature-selected subspace; the low-loss anchor diagnostic does not certify every posterior draw. In the practical large-network implementation, the selector is the positive spectrum of the cross-entropy Hessian, used as a Fisher/GGN-motivated proxy. The longitudinal spine and the transverse subspace therefore play distinct, non-redundant roles: the spine provides stored centers around which the approximation places mass, and the transverse sampling determines how it is dispersed. How much predictive diversity the longitudinal component adds depends on how much functional connectivity, rather than near-exact invariance, the traversed corridor actually contains.

\begin{figure}[t]
\centering
\includegraphics[width=\textwidth]{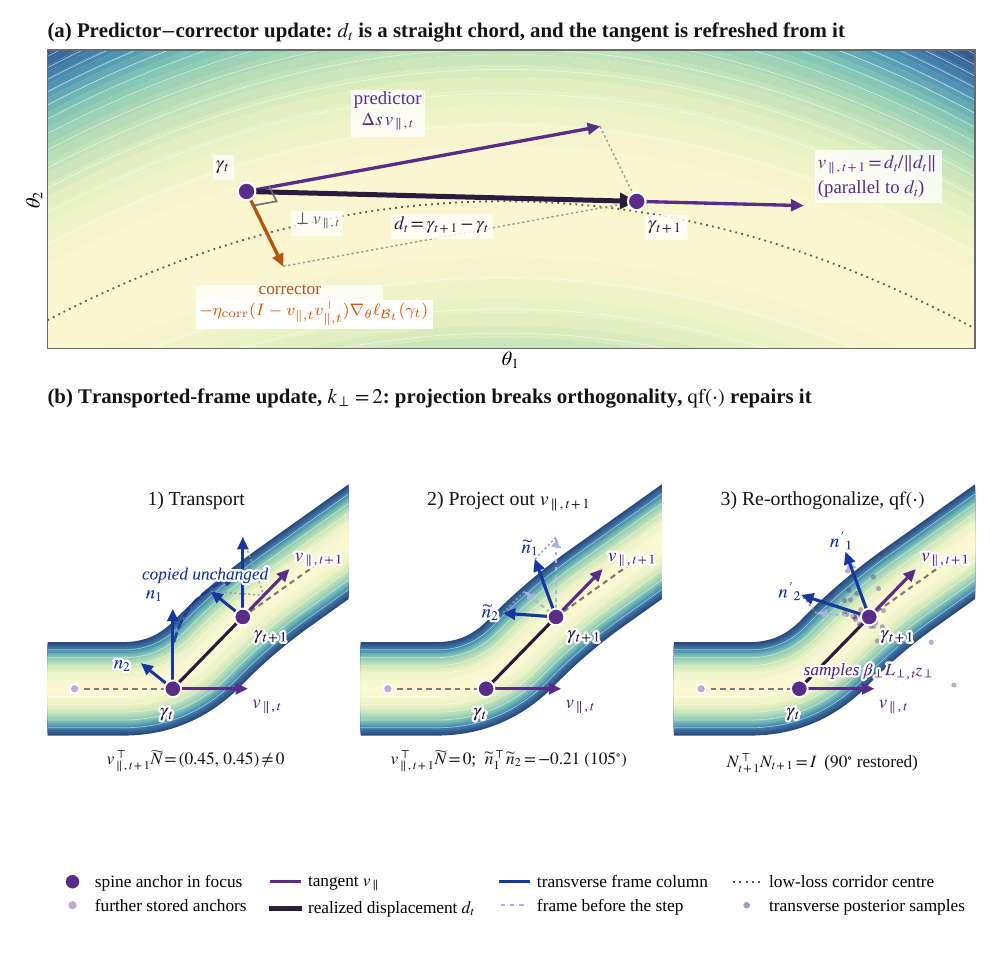}
\caption{\textbf{Update mechanics of the discrete spine.}
\textbf{(a)} One predictor--corrector step, Eq.~\eqref{eq:spine-step}. Both terms
are evaluated at $\gamma_t$, so $\gamma_{t+1}$ is computed rather than chosen: the
predictor advances along the current tangent, which is only an approximate
corridor direction, and the corrector projects out the component of
$\nabla_\theta\ell_{\mathcal B_t}(\gamma_t)$ along that tangent and pulls back toward the
low-loss region. The realized displacement $d_t=\gamma_{t+1}-\gamma_t$ therefore
departs from $v_{\parallel,t}$, and Eq.~\eqref{eq:tangent-refresh} refreshes the
tangent from $d_t$. The corrector-to-predictor ratio, the resulting tangent
rotation, and the offset of $\gamma_t$ from the corridor centre are all
exaggerated for legibility; Appendix~\ref{app:transport} reports the offsets
actually observed along the spine.
\textbf{(b)} The transported-frame update, Eq.~\eqref{eq:frame-transport}, drawn
with $k_\perp=2$ because re-orthogonalisation is vacuous for a single column.
Projecting the transported frame onto the complement of $v_{\parallel,t+1}$
removes the tangent component but leaves the columns mutually oblique
($\widetilde{n}_1^\top\widetilde{n}_2=-0.21$, $105^\circ$); $\mathrm{qf}(\cdot)$
restores $N_{t+1}^\top N_{t+1}=I$ while preserving the direction of the first
column. The frame is transported, not recomputed from a fresh local curvature
decomposition at $\gamma_{t+1}$. Shading is the low-loss corridor, drawn from the
same quadratic loss field in both panels.
Figure~\ref{fig:support} shows the resulting global support; this figure shows
only the local update rule.}
\label{fig:frame-transport}
\end{figure}

Figure~\ref{fig:support} in the main text illustrates the discrete support used by practical TRL. Figure~\ref{fig:frame-transport} below focuses on the complementary frame-update view: updated tangents, transported transverse frames, and local perturbations around discrete spine anchors. This geometric mismatch with a single Euclidean ellipsoid is the starting point for TRL. The scalable method is a discrete transported low-rank approximation, not a fully smooth tubular pushforward.

\subsection{Fisher-motivated tangent/normal decomposition}
The tangent/transverse interpretation follows the Fisher/GGN sensitivity picture in Appendix~\ref{app:background}: low-curvature directions approximate functionally weak or invariant motion, while leading positive-curvature directions define a data-sensitive transverse subspace. Practical TRL retains one constructed longitudinal direction and a rank-$k_\perp$ transverse basis rather than a full spectral decomposition. Figure~\ref{fig:frame-transport} shows how that basis is transported and re-orthogonalized along the discrete spine.

\subsection{Prior- vs data-dominated variance}

Let $F_0$ denote the likelihood-side Fisher/GGN curvature at the MAP, with eigenpairs $(u_i,\lambda_i(F_0))$, and let $\lambda I$ denote an isotropic prior precision. Under the idealized Gaussian precision $H\approx F_0+\lambda I$, the variance along $u_i$ is approximately $1/(\lambda_i(F_0)+\lambda)$.
Thus:
\begin{itemize}[leftmargin=*]
\item For tangential directions with $\lambda_i(F_0) \approx 0$, the variance is dominated by the prior, $\mathrm{Var}_\parallel \approx 1/\lambda$.
\item For transverse directions with $\lambda_i(F_0) \gg \lambda$, the variance is dominated by the data, $\mathrm{Var}_\perp \approx 1/\lambda_i(F_0)$.
\end{itemize}

The tangential/transverse split above is most literal when transverse curvature dominates prior precision. In the low-rank subspace retained by the practical method, that condition holds only partially. The retained directions identify data-sensitive structure, but the HVP eigenvalues are computed from mini-batch-mean cross-entropy, whereas the inherited prior base precision comes from a Laplace marginal-likelihood fit that accumulates likelihood curvature over a 5,000-example subset. These terms are therefore not a matched dataset-sum Laplace precision, and the transverse Gaussian can be only weakly anisotropic and substantially prior-dominated. This separates support selection from scale: the Fisher/GGN picture motivates where transverse uncertainty should live, while the implementation selects that support with the positive spectrum of the cross-entropy Hessian and sets its magnitude largely through projected prior precision and validation tempering. The large-network sampler uses a two-block, block-isotropic approximation: classifier-head parameters use the base precision inherited from the last-layer Laplace fit, while non-head parameters use a boosted backbone precision. Appendix~\ref{sec:boost-ablation} shows that this two-block structure is important for calibrated samples. Accordingly, $\mathrm{Var}_\perp \approx 1/\lambda_i(F_0)$ is a subspace-selection idealization rather than a literal description of the transverse covariance used by the large-network sampler.

A single Euclidean ellipsoid can become wide in poorly identified directions, but those directions need not remain aligned with a curved functional valley and may leave the low-loss corridor. Weight-space sampling can then place mass in directions that look flat under a local quadratic approximation but perform poorly under the nonlinear network. TRL instead allocates uncertainty to a low-rank tube: a Fisher/GGN-motivated positive-curvature proxy selects the transverse support, while the longitudinal component is anchored to a constructed discrete spine. Appendix~\ref{app:trl_vs_ela_support} makes this distinction explicit. Relative to Euclidean and linearized Laplace baselines, TRL changes where posterior mass is allocated---low-rank Gaussian uncertainty across selected data-sensitive directions plus discrete coverage along stored anchors---rather than changing the within-direction Laplace variance law.

\subsection{Continuous geometric idealization}
\label{sec:trl-ideal}

\begin{figure}[H]
    \centering
    \includegraphics[width=\linewidth]{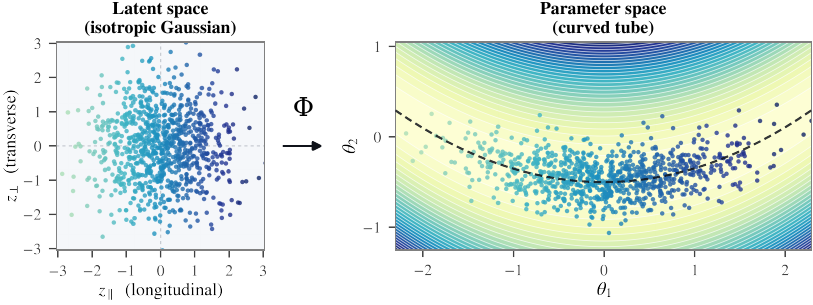}
    {\footnotesize
    \caption{\textbf{Continuous tubular idealization.}
    An isotropic Gaussian in latent longitudinal/transverse coordinates is mapped by $\Phi$ to a curved tube in a schematic 2D parameter-space projection; $\theta_1$ and $\theta_2$ are projected coordinates, not individual weights. The dashed curve is the centerline and the spread represents local fibers. Practical TRL uses a finite predictor--corrector spine and transported low-rank transverse factors rather than a global smooth pushforward or repeated curvature decomposition.}
    \label{fig:trl-map}
    }
\end{figure}

Figure~\ref{fig:trl-map} illustrates the continuous idealization: a latent coordinate controls displacement along a low-loss valley, while transverse coordinates control perturbations across it. The implementation discretizes this picture: samples are drawn around stored spine anchors and transverse perturbations are expressed in a transported low-rank basis. The idealization is explanatory rather than literal; it identifies the geometric objects of interest---a low-loss centerline, a transverse frame, and a local transverse scale---while the scalable algorithm specifies how they are approximated from one trained checkpoint.

Let $\theta_0=\theta_{\mathrm{MAP}}$ and let $G(\theta)$ denote a positive semi-definite curvature metric, such as a Fisher or GGN approximation plus prior precision. The idealized TRL construction assumes a smooth curve
\begin{align}
\gamma : \mathbb{R} \to \mathbb{R}^K, \qquad t \mapsto \gamma(t),
\label{eq:ideal-spine-curve}
\end{align}
The curve in Eq.~\eqref{eq:ideal-spine-curve} satisfies $\gamma(0)=\theta_0$ and has approximately constant loss over the range of interest. Its tangent $\dot\gamma(t)$ represents a low-Fisher-curvature direction, while the transverse space captures directions in which the likelihood is data-sensitive.

At each point on the ideal curve, suppose we have a transverse frame $N(t)\in\mathbb{R}^{K\times k}$ and a transverse covariance square-root factor $L_\perp(t)\in\mathbb{R}^{k\times k}$. The idealized tubular map is
\begin{align}
\Phi(z_\parallel,z_\perp)
=
\gamma(\alpha_\parallel z_\parallel)+N(\alpha_\parallel z_\parallel)L_\perp(\alpha_\parallel z_\parallel)z_\perp,
\label{eq:trl-general}
\end{align}
where $z=(z_\parallel,z_\perp)\sim\mathcal{N}(0,I_{1+k})$ and $\alpha_\parallel$ sets the longitudinal scale; the local transverse covariance in this idealization is $L_\perp L_\perp^\top$. This construction motivates TRL: posterior mass should be allowed to occupy flat, low-loss directions while remaining more cautious across data-sensitive transverse directions. Whether longitudinal averaging yields additional predictive diversity depends on whether the traversed corridor contains functional connectivity beyond near-exact invariances.

The map in Eq.~\eqref{eq:trl-general} is a valid tubular parametrization only locally: its transverse radius must remain below the normal injectivity radius, or reach, of the curve $\gamma$. Beyond this radius, normal fibers can intersect and the coordinates $(z_\parallel,z_\perp)$ need not be unique. We use this condition only as part of the continuous geometric idealization. The practical sampler operates with a finite discrete spine and validation-selected transverse scales, but we do not explicitly certify the reach of the learned trajectory.

Equation~\eqref{eq:trl-general} is a geometric model rather than the literal high-dimensional algorithm. A fully faithful implementation would require a smooth curve, continuous transport of frames, exact low-curvature tangents, and fresh local curvature factors along the entire path. That is computationally prohibitive in modern deep networks and is not the approximation used in our experiments. The implementation instead asks a more practical question: whether an amortized discrete approximation retains the parts of the geometry that matter empirically.

%% file: appendix_impact_statement.tex
\section{Impact Statement}
\label{app:impact_statement}

This work studies a post-hoc uncertainty approximation that may support reliability analysis, but it does not establish safety in deployed or high-stakes systems. Improved calibration does not correct dataset bias, distributional mismatch, or failures outside the evaluated settings; deployment therefore requires independent validation and domain-specific safeguards.

%% file: appendix_additional_diagnostics.tex
\section{Additional Predictive and Geometric Diagnostics}
\label{app:additional_diagnostics}

These analyses were added after inspection of the original experiments. Toy scale selection uses separate development seeds. Predictive contrasts are paired at the seed/checkpoint level; separate historical geometric diagnostics are labeled below. The 100 sine reporting seeds had already been examined; the two-moons extension uses 100 newly evaluated reporting seeds. The three CIFAR diagnostics below are new realizations from saved MAP checkpoints, separate from the historical three-checkpoint ablation and the five-checkpoint headline benchmark. They do not replace those benchmark rows.

\subsection{Development-selected toy scales}
\label{app:toy_scale_controls}
For each method we minimize mean predictive NLL on development seeds $50$--$69$ over the same ascending grid
\begin{equation}
\label{eq:additional_toy_scale_grid}
\mathcal A=\{0,0.001,0.002,0.005,0.01,0.02,0.05,0.1,0.2,0.5,1,2,4\}.
\end{equation}
Ties choose the first ascending value. ELA scales weight standard deviations; LLA scales logit/function perturbations before the likelihood transformation, preserving the analytic regression mean at MAP. TRL scales are absolute transverse $\beta_\perp$ values. ELA and LLA keep the marginal-likelihood-fitted prior precision fixed: $\alpha$ is a standard-deviation tempering parameter, so their Gaussian covariance is multiplied by $\alpha^2$. It is selected by development NLL, rather than by refitting the prior; MAP+TS instead tempers output logits. Each method retains its original rank and predictive construction; the spine prefix/length is fixed rather than reselected. Common random draws are used across candidate scales within each method. We freeze selected scales before evaluating the reporting outputs of that extension. This is post-hoc experimental design with development-only scale selection, not a preregistered confirmatory study.

\paragraph{Sine.}
ELA/LLA/TRL-single select $0.05/0.2/0.1$, respectively; TRL-full selects the boundary $\beta=0$, retaining uniform anchor mixing without transverse perturbations. The full-minus-original-full mixture-NLL difference is $-0.000269$ (95\% paired t interval $[-0.000916,0.000378]$), with 56/100 seeds favoring zero. Against tuned single, zero-scale full has difference $-0.242198$ ($[-0.294687,-0.189708]$), winning 84/100. Thus the original full-spine gain over the MAP-centered tube is not explained solely by single-anchor scale. This contrast concerns the MAP-centered tube; the additional relocated-center controls below address the choice of center. The zero-scale result does not establish a need for transverse perturbations in this sine configuration.

Tuned LLA is better than both full configurations: zero-scale full minus LLA NLL is $+0.250391$ ($[0.214779,0.286004]$), with LLA winning 98/100 seeds. LLA also has lower CRPS. Tuned ELA is closest to nominal 68\%/95\% coverage among these three methods, demonstrating that one likelihood score does not establish best calibration under every criterion. Table~\ref{tab:additional_sine_scales} reports all cited mixture scores.

\begin{table}[H]
\centering
\caption{Sine scale controls: mean $\pm$ sample SD over the same 100 reporting seeds, with 200 predictive samples. Central coverage uses predictive mixture quantiles. The known observation-noise standard deviation remains 0.15 for every method and scale. Single-center indices and their own scales are fixed on development seeds; $t=0$ denotes the MAP.}
\label{tab:additional_sine_scales}
\footnotesize
\begin{adjustbox}{max width=\textwidth}
\begin{tabular}{lcccc}
\toprule
Method & NLL$_{\rm mix}$ $\downarrow$ & CRPS $\downarrow$ & Cov. 68\% & Cov. 95\% \\
\midrule
TRL-single, original $\beta=0.005$ & \meanstd{1.1293}{0.6509} & \meanstd{0.1724}{0.0219} & \meanstd{0.478}{0.050} & \meanstd{0.741}{0.052} \\
ELA, selected $\alpha=0.05$ & \meanstd{0.4347}{0.1284} & \meanstd{0.2165}{0.0314} & \meanstd{0.715}{0.075} & \meanstd{0.960}{0.040} \\
LLA, selected $\alpha=0.2$ & \meanstd{0.0622}{0.1821} & \meanstd{0.1569}{0.0176} & \meanstd{0.626}{0.068} & \meanstd{0.914}{0.052} \\
TRL-single, selected $\beta=0.1$ & \meanstd{0.5548}{0.4521} & \meanstd{0.1820}{0.0273} & \meanstd{0.501}{0.065} & \meanstd{0.780}{0.068} \\
TRL-shifted, $t=2$, $\beta=0.1$ & \meanstd{0.5517}{0.4390} & \meanstd{0.1819}{0.0267} & \meanstd{0.502}{0.063} & \meanstd{0.780}{0.068} \\
TRL-endpoint, $t=14$, $\beta=0.5$ & \meanstd{1.2849}{0.1123} & \meanstd{0.4984}{0.0757} & \meanstd{0.743}{0.129} & \meanstd{0.998}{0.012} \\
TRL-full, original $\beta=0.005$ & \meanstd{0.3129}{0.3176} & \meanstd{0.1686}{0.0212} & \meanstd{0.605}{0.092} & \meanstd{0.859}{0.077} \\
TRL-full, selected $\beta=0$ & \meanstd{0.3126}{0.3165} & \meanstd{0.1686}{0.0212} & \meanstd{0.605}{0.092} & \meanstd{0.859}{0.077} \\
\bottomrule
\end{tabular}
\end{adjustbox}
\end{table}

\paragraph{Relocated single-center controls.}
We reuse the saved fits and the same 15-anchor prefix, rank, scale grid and sample budget. Mean development NLL on seeds $50$--$69$ jointly selects a global index $t\in\{0,\ldots,14\}$ and its own $\beta$; ties choose ascending $t$, then $\beta$. The unrestricted optimum is the MAP ($t=0$, $\beta=0.1$). Because that choice is not displaced, a companion search restricted to $t>0$ uses the same development grid and selects $t=2$, $\beta=0.1$. The endpoint is the last anchor of the mixed prefix, $t=14$, with independently selected $\beta=0.5$. Each choice uses development scores only and is fixed across the reporting seeds; the strict-displacement companion is selected before its own reporting evaluation. This further analysis is post hoc, on the previously examined reporting tasks, with no new fitting or path construction.

Table~\ref{tab:additional_sine_scales} reports displaced/endpoint NLLs $0.551656/1.284910$. Tuned full minus displaced is $-0.239040$ (95\% paired t interval $[-0.290268,-0.187813]$), with 86/100 wins; original full minus displaced is $-0.238771$ ($[-0.289909,-0.187633]$), also with 86/100 wins. These differ from the earlier 84/100 comparisons against the tuned MAP-centered tube. Tuned full also has lower CRPS than displaced single: difference $-0.013362$ ($[-0.015786,-0.010938]$), with 85/100 wins. The gain persists after allowing a relocated center and its own scale within this globally selected family. It does not rule out per-task anchor selection, establish exact posterior fidelity, or reverse the tuned LLA ranking.

For LLA the Gaussian observation likelihood and linearized function give an analytic Gaussian predictive. At unit scale its analytic NLL is $0.391421\pm0.078089$, versus $0.405237\pm0.090020$ for the 200-sample mixture estimate. At selected scale 0.2, the respective values are $0.056960\pm0.178207$ and $0.062224\pm0.182149$; analytic CRPS is $0.156424$. Negative logs of unbiased Monte Carlo density estimates have upward bias in expectation by Jensen's inequality; a particular estimate need not exceed the analytic score. This finite-sample issue applies to sampled mixture likelihoods generally, including TRL, and does not invalidate the tuned LLA advantage. Reported closed-form scores use the float32 eigendecomposition covariance factor of the predictive sampler, evaluated on all 100 saved fits; this holds the numerical covariance construction fixed when comparing analytic and sampled scores. A float64 Cholesky reconstruction was checked at selected scale 0.2 on five prespecified seeds, with maximum absolute NLL difference 0.0056. This is a limited numerical sensitivity check, not a replacement calculation for all 100 seeds.

The standard-deviation multiplier 0.2 reduces LLA functional covariance to $0.04$ of its original value, leaving observation noise fixed and the analytic predictive mean unchanged. The regional summaries below are descriptive: each entry is a mean over the 100 reporting seeds; $V_f$ denotes empirical functional variance of the 200 predictive draws, excluding observation noise. ``In'' and ``out'' refer to $[-3,3]$ and its complement in the evaluation grid. NLL uses the same finite-mixture estimate as Table~\ref{tab:additional_sine_scales}, rather than the analytic LLA score. These summaries do not identify the cause of the cross-method differences.

\begin{center}
\small
\begin{tabular}{lrrrr}
\toprule
Method & NLL in & NLL out & $V_f$ in & $V_f$ out \\
\midrule
MAP & -0.424902 & 2.700125 & 0.000000 & 0.000000 \\
LLA, $\alpha=1$ & -0.426604 & 1.237077 & 0.002965 & 3.402111 \\
LLA, $\alpha=0.2$ & -0.425146 & 0.549594 & 0.000119 & 0.136084 \\
TRL-full, $\beta=0.005$ & -0.373274 & 0.999044 & 0.015177 & 0.059365 \\
\bottomrule
\end{tabular}
\end{center}
Development-only interior and exterior criteria each select scale 0.2, although the interior score differences are small. Appendix~\ref{app:sine_linearization} separates linearized retained-subspace and complementary predictive variance with the prior precision held fixed.

\paragraph{Two-moons.}
The same grid selects ELA/LLA multipliers $0.02/0.05$ and TRL-single/full transverse scales $0.10/0.05$. All 250 predictive draws and the original training protocol are retained; only scale is selected. Table~\ref{tab:toy_moons} reports the 100-seed extension. Both NLL and ECE comparisons between tuned LLA and TRL-full have paired intervals containing zero; no formal equivalence test was performed. Full-minus-tuned-single and LLA-minus-full NLL differences are $0.000486164$ and $0.000486556$, respectively: their identical five-decimal rounded values are not identical unrounded estimates. ECE is evaluated at NLL-selected settings, without a separate calibration-metric search.

\subsection{Paired full projected precision and rotations}
\label{app:additional_precision}
We evaluate 40-anchor, rank-30 spines rebuilt with the submitted configuration from three saved CIFAR-100/ResNet-18 MAP checkpoints, with fixed $\beta_\perp=4$, 25 common predictive draws, and identical augmentation streams and rolling 25-batch FixBN across variants. No new validation-scale tuning is performed. The same posterior draw is used for ID and OOD evaluation. These new realizations explain why the diagonal reference here differs slightly from the historical three-checkpoint benchmark. Define
\begin{align}
P_t & =\operatorname{diag}(\widehat\lambda)+N_t^\top\Lambda_{\rm prior}N_t,\label{eq:additional_projected_precision}\\
D_t & =\operatorname{diag}(P_t),\label{eq:additional_diagonal_precision}\\
\rho_t & =\lVert D_t^{-1/2}(P_t-D_t)D_t^{-1/2}\rVert_2.\label{eq:additional_precision_correlation}
\end{align}
``Full'' retains the complete projected prior, while the curvature values $\widehat\lambda$ remain frozen at the MAP; this is not a freshly evaluated local Hessian. Across the 120 anchors, maximum $\rho_t$ is 0.008938 and maximum $\lVert D_t^{-1}-P_t^{-1}\rVert_F/\lVert P_t^{-1}\rVert_F$ is 0.357\%. Since $(1-\rho_t)D_t\preceq P_t\preceq(1+\rho_t)D_t$, directional standard-deviation ratios lie between $(1+\rho_t)^{-1/2}$ and $(1-\rho_t)^{-1/2}$, approximately $-0.444\%$ to $+0.450\%$ at the observed maximum, for common mean and scale. This is a bound for linear projections of the weight Gaussian, not for nonlinear predictive NLL or arbitrary network reparameterizations.

The two-block prior dominates scale in these three diagnostics: $\widehat\lambda_{\max}/\lambda_{\rm back}$ is $0.0894/0.0981/0.0823$, and the retained-curvature contribution is at most 9\% of any diagonal precision entry. The largest full-precision eigenvalue ratios within checkpoints are $1.0567/1.0691/1.0542$. Curvature therefore selects the transverse subspace while prior precision and global tempering dominate its scale in this setting; a small Frobenius-norm rotation discrepancy alone would not establish a uniform spectral bound.

For an orthogonal transverse coordinate change, we rotate the entire precision, $P'_t=R^\top P_tR$, including its curvature block. The full covariance mapped back to the original coordinates is $R(P'_t)^{-1}R^\top$; in the tested $30\times30$ float64 matrices, its relative error is below $5.8\times10^{-16}$. This numerical identity is not a test of general network-reparameterization invariance. General rotations use QR of Gaussian random matrices with no angular restriction. The rotated diagonal rule is $R\operatorname{diag}(P'_t)^{-1}R^\top$; its maximum covariance change is 1.333\%. For rotations restricted to consecutive MAP-eigenvalue blocks whose maximum-minus-minimum span is at most 5\% of the largest eigenvalue in that block, the maximum is 0.155\%. The block-restricted tests are matrix diagnostics only; predictive evaluation of the general-rotation diagonal variant is included in Table~\ref{tab:additional_cifar_predictive}.

\begin{table}[H]
\centering
\caption{New paired CIFAR diagnostics: maxima over 40 anchors per checkpoint. Covariance discrepancies are relative Frobenius norms, shown as percentages; the last column is the largest diagonal precision fraction contributed by retained curvature. Three independently trained checkpoints are the replicates, not 120 anchors.}
\label{tab:additional_precision_geometry}
\footnotesize
\begin{adjustbox}{max width=\textwidth}
\begin{tabular}{lccccc}
\toprule
Checkpoint & $\rho_{\max}$ & Diag./full $C$ (\%) & General rotation (\%) & 5\%-block rotation (\%) & Curvature fraction (\%) \\
\midrule
0 & 0.007823 & 0.294 & 1.204 & 0.117 & 8.23 \\
1 & 0.008938 & 0.357 & 1.333 & 0.143 & 8.97 \\
2 & 0.007630 & 0.325 & 1.100 & 0.155 & 7.63 \\
\bottomrule
\end{tabular}
\end{adjustbox}
\end{table}

\begin{table}[H]
\centering
\caption{Matched CIFAR predictive comparisons: mean $\pm$ sample SD over three checkpoints. All variants use the same newly constructed anchors, sample budget and FixBN protocol. Recentering is the auxiliary raw mean-objective step defined below, without damping or line search.}
\label{tab:additional_cifar_predictive}
\footnotesize
\begin{adjustbox}{max width=\textwidth}
\begin{tabular}{lccccc}
\toprule
Variant & NLL $\downarrow$ & ECE $\downarrow$ & Brier $\downarrow$ & Acc. $\uparrow$ & SVHN AUROC $\uparrow$ \\
\midrule
Diagonal & \meanstd{0.954754}{0.001951} & \meanstd{0.01426}{0.00184} & \meanstd{0.35576}{0.00226} & \meanstd{0.74510}{0.00173} & \meanstd{0.84749}{0.02430} \\
Full precision & \meanstd{0.954766}{0.001955} & \meanstd{0.01419}{0.00184} & \meanstd{0.35576}{0.00226} & \meanstd{0.74517}{0.00169} & \meanstd{0.84750}{0.02428} \\
General-rotation diagonal & \meanstd{0.954781}{0.001923} & \meanstd{0.01376}{0.00160} & \meanstd{0.35576}{0.00227} & \meanstd{0.74527}{0.00167} & \meanstd{0.84751}{0.02434} \\
Full + raw recentering & \meanstd{0.960225}{0.004621} & \meanstd{0.01473}{0.00310} & \meanstd{0.35751}{0.00365} & \meanstd{0.74153}{0.00211} & \meanstd{0.84936}{0.02612} \\
\bottomrule
\end{tabular}
\end{adjustbox}
\end{table}
Full minus diagonal mean NLL is $+0.0000116$; the general-rotation diagonal mean is 0.954781. The discarded precision terms and tested rotations have negligible effects on these aggregate predictive scores in this setting. This observation does not establish small effects for other priors, checkpoints or parameterizations.

\subsection{Auxiliary transverse gradients and raw recentering}
\label{app:additional_centering}
The implemented anchors follow a data-loss spine and need not be stationary for a posterior objective. To quantify one centering diagnostic, we fix the first 5,000 clean training examples in the training split and use train-mode BatchNorm with fixed nonshuffled batches of 128. The objective is
\begin{align}
U_b(\theta) & =b\,\overline{\mathrm{CE}}_{\mathcal B}(\theta)+\tfrac12\theta^\top\Lambda_{\rm prior}\theta,\label{eq:additional_auxiliary_objective}\\
P_{b,t} & =b\operatorname{diag}(\widehat\lambda)+N_t^\top\Lambda_{\rm prior}N_t,\quad b\in\{1,5000\}.\label{eq:additional_auxiliary_precision}
\end{align}
The batch partition is part of this auxiliary train-BatchNorm objective. It differs from the evaluation-mode minibatch objective of the large-network spine and from FixBN posterior prediction. The curvature remains the transported MAP proxy obtained from five clean HVP batches, not the local Hessian of $U_b$ and not a matched full-data Laplace construction. Setting $b=5000$ converts the 5,000-example mean loss to a sum while retaining the same prior; it changes the likelihood-to-prior balance rather than merely changing global units.

Writing $g_{b,t}=N_t^\top\nabla U_b(\gamma_t)$, the candidate coordinate shift and its sampling-standardized magnitude are
\begin{align}
\delta_{b,t} & =-P_{b,t}^{-1}g_{b,t},\label{eq:additional_raw_recentering}\\
s_{b,t} & =\beta_\perp^{-1}\sqrt{g_{b,t}^\top P_{b,t}^{-1}g_{b,t}},\qquad \beta_\perp=4.\label{eq:additional_standardized_recentering}
\end{align}
This magnitude is the Mahalanobis norm of the candidate mean shift under that auxiliary Gaussian covariance; it need not be negligible. Table~\ref{tab:additional_centering} reports matched median/max summaries for both targets. Only the raw $b=1$ step was evaluated predictively, starting from the full-precision sampler, and worsened NLL in every checkpoint (mean 0.954766 to 0.960225). No damping, line search or predictive evaluation of the $b=5000$ step was performed. Worse predictive NLL neither resolves non-stationarity nor validates a posterior interpretation; the test characterizes a surrogate preconditioned correction.

\begin{table}[H]
\centering
\caption{Auxiliary standardized centering shifts: median and maximum across 40 anchors for each checkpoint and target. The scales change both data gradient and the MAP-curvature proxy consistently, while keeping the prior fixed.}
\label{tab:additional_centering}
\footnotesize
\begin{tabular}{lcccc}
\toprule
Checkpoint & $s_1$ median & $s_1$ max & $s_{5000}$ median & $s_{5000}$ max \\
\midrule
0 & 0.836 & 0.852 & 1.454 & 2.356 \\
1 & 1.059 & 1.086 & 1.355 & 2.223 \\
2 & 1.122 & 1.167 & 1.390 & 2.417 \\
\bottomrule
\end{tabular}
\end{table}

\subsection{Fresh/fixed bases and tangent orthogonality}
\label{app:additional_orthogonality}
The default transported frame enforces approximate tangent orthogonality; in the new controlled spines, maximum $\lVert N_t^\top v_t\rVert_2$ is below $1.8\times10^{-7}$. Historical fresh leading eigenspaces and fixed-MAP bases do not enforce the same constraint at later anchors. For the historical fresh cache we reconstruct unit incoming displacement tangents at the 39 noninitial anchors per checkpoint; the initial tangent was not stored and is omitted. The fixed measurements below instead use the actual initial frame of each new controlled spine and its stored tangents. They are distinct realizations and should not be read as paired fresh-versus-fixed predictive comparisons.

For an orthonormal frame $Q$ and a unit tangent $v$, overlap is $\lVert Q^\top v\rVert_2$. The tangential fraction of weight-sampling variance, for coordinate covariance $A$, is $v^\top QAQ^\top v/\operatorname{tr}(A)$; global transverse scale cancels. Table~\ref{tab:additional_orthogonality} gives both statistics. Fresh mean overlaps are $0.421/0.388/0.402$, with maximum 0.559; mean tangential variance fractions are 0.52\%--0.61\%, with maxima below 1\%. These are weight-space measures, not bounds on predictive effects. They establish neither flatness of realized tangents nor the effect of projecting the fresh bases. A projected-fresh predictive control was not performed.

\begin{table}[H]
\centering
\caption{Tangent diagnostics, mean / maximum within each checkpoint. Fresh uses 39 historical noninitial anchors; fixed uses 40 anchors of the new controlled spine. Overlap is dimensionless; variance fractions are percentages. These are separate geometric diagnostics rather than a matched predictive ablation.}
\label{tab:additional_orthogonality}
\footnotesize
\begin{adjustbox}{max width=\textwidth}
\begin{tabular}{lcccc}
\toprule
Checkpoint & Fresh overlap & New fixed overlap & Fresh tangent var. (\%) & New fixed tangent var. (\%) \\
\midrule
0 & 0.421 / 0.559 & 0.229 / 0.296 & 0.608 / 0.982 & 0.181 / 0.292 \\
1 & 0.388 / 0.498 & 0.320 / 0.368 & 0.522 / 0.799 & 0.350 / 0.447 \\
2 & 0.402 / 0.534 & 0.285 / 0.331 & 0.562 / 0.882 & 0.279 / 0.363 \\
\bottomrule
\end{tabular}
\end{adjustbox}
\end{table}
The results support interpreting the historical fresh/fixed tables as sampling-frame sensitivity checks. Close aggregate scores under these alternatives do not isolate curvature refresh while preserving the transverse constraint, or prove that non-orthogonality has no predictive cost.

\subsection{Sine linearization and retained-support controls}
\label{app:sine_linearization}
We reuse the 20 development and 100 reporting fits, without training, prior optimization or new path construction. The marginal-likelihood-fitted prior precision remains frozen. Each linearized family selects its own standard-deviation tempering multiplier $\alpha$ by mean development NLL on the grid in Appendix~\ref{app:toy_scale_controls}; the choices and source hashes are frozen before its reporting evaluation. All selected values are interior to the grid. These are post-hoc controls on previously examined reporting tasks. Every family shares the MAP mean, known observation noise 0.15 and Jacobian $J(x)$; Gaussian predictive NLL, CRPS and quantiles are evaluated analytically.

Let $N_0,L_{\perp,0}$ be the saved MAP transverse frame and factor, with original scale $\beta_0=0.005$. The retained linearized TRL law uses $B_0=N_0L_{\perp,0}/\beta_0$ and variance $\alpha^2\lVert J(x)B_0\rVert_2^2$. Thus its support and relative coordinate scales are those actually used by the nonlinear MAP-centered sampler. Let $\Sigma=LL^\top$ be the full LLA covariance reconstructed by its original float32 eigendecomposition helper, and let $P=QQ^\top$, where $Q$ orthonormalizes the saved $N_0$. The projected retained and complementary laws use $P\Sigma P$ and $(I-P)\Sigma(I-P)$, respectively; these are projections of full LLA perturbations, not inverse restricted precision matrices. Table~\ref{tab:sine_linearized_controls} reports the controls.

\begin{table}[H]
\centering
\caption{Analytic Gaussian sine controls: mean $\pm$ sample SD over 100 paired reporting seeds. Each family has its own development-selected standard-deviation tempering multiplier; the prior precision and predictive mean stay fixed. The bottom two rows use a newly recomputed leading-precision-eigenvector split rather than substituting that basis for the saved TRL frame.}
\label{tab:sine_linearized_controls}
\small
\begin{tabular}{lccc}
\toprule
Linearized family & $\alpha$ & NLL $\downarrow$ & CRPS $\downarrow$ \\
\midrule
Full LLA & 0.2 & \meanstd{0.0570}{0.1782} & \meanstd{0.1564}{0.0175} \\
Saved TRL rank-30 law & 0.2 & \meanstd{0.0680}{0.1898} & \meanstd{0.1574}{0.0183} \\
Full LLA projected to saved rank-30 span & 0.2 & \meanstd{0.0679}{0.1898} & \meanstd{0.1574}{0.0183} \\
Full LLA projected to its complement & 0.5 & \meanstd{0.1102}{0.1887} & \meanstd{0.1599}{0.0162} \\
Recomputed leading-precision rank 30 & 0.2 & \meanstd{0.0678}{0.1888} & \meanstd{0.1574}{0.0183} \\
Recomputed spectral complement & 0.5 & \meanstd{0.1114}{0.1932} & \meanstd{0.1600}{0.0163} \\
\bottomrule
\end{tabular}
\end{table}

The saved rank-30 linearized law has NLL $0.067987$, versus $0.056960$ for full LLA; its paired difference is $0.011027$ (95\% t interval $[0.007029,0.015025]$). The complementary law gives $0.110201$; retained minus complement is $-0.042214$ ($[-0.051265,-0.033164]$), favoring retained in 83/100 seeds. Full LLA therefore retains an advantage, but the retained directions alone already support a linearized predictive close in NLL to the full reference. This weakens an explanation of the nonlinear sampler's larger NLL based only on discarded directions.

At the previously selected nonlinear MAP-tube scale $\beta=0.1$, linearizing the same saved factor gives analytic NLL $0.304261$, compared with the existing finite-mixture nonlinear score $0.554813$. The paired difference is $-0.250552$ ($[-0.305941,-0.195163]$), with 81/100 seeds favoring the linearization. This fixed-scale comparison changes the predictive map and the analytic-versus-finite-sample scoring construction; it does not quantify a pure causal fraction of the original cross-method gap. The independently tuned linearized law selects $\alpha=0.2$. Neither result establishes fidelity to an exact Bayesian posterior.

\paragraph{Regional variance decomposition.}
For the saved span, the full untempered functional variance decomposes as
\begin{equation}
\label{eq:additional_variance_decomposition}
V(x)=\lVert J(x)PL\rVert_2^2+\lVert J(x)(I-P)L\rVert_2^2
+2\langle J(x)PL,J(x)(I-P)L\rangle.
\end{equation}
The cross term is retained rather than assuming that the saved frame diagonalizes the replayed full covariance exactly. Table~\ref{tab:sine_variance_split} reports mean regional shares, formed within each seed before averaging. The retained projection accounts for 99.80\% of interior variance and 68.12\% of exterior variance; the complement contributes 0.20\% and 31.96\%, respectively. The complement's exterior variance is non-negligible, but its presence alone does not imply that it is needed to approach the full reference's likelihood score.

\begin{table}[H]
\centering
\caption{Untempered full LLA variance and decomposition shares: mean $\pm$ sample SD across 100 reporting seeds. Shares are ratios of regional mean variances within each seed; their means, including the signed cross term, sum to 100\%. Observation noise is excluded. These analytic variances differ from empirical 200-draw variances in the descriptive regional summary of Appendix~\ref{app:toy_scale_controls}.}
\label{tab:sine_variance_split}
\small
\begin{tabular}{lrrrr}
\toprule
Region & Full variance & Retained (\%) & Complement (\%) & Cross (\%) \\
\midrule
Inside $[-3,3]$ & \meanstd{0.002969}{0.000196} & \meanstd{99.802}{0.147} & \meanstd{0.204}{0.152} & \meanstd{-0.006}{0.006} \\
Outside $[-3,3]$ & \meanstd{3.432400}{1.266128} & \meanstd{68.118}{8.234} & \meanstd{31.956}{8.229} & \meanstd{-0.073}{0.084} \\
\bottomrule
\end{tabular}
\end{table}

\paragraph{Numerical frame check.}
A new eigendecomposition does not reproduce the saved rank-30 frame exactly: the mean squared singular overlap is 0.547, while the adjacent relative precision gap at rank 30 averages $4.15\times10^{-4}$, close to the prior-dominated part of the spectrum. We preserve the saved frame for the primary control. Across all reporting grid points, the absolute cross term is at most 0.739\% of full variance; replacing the saved TRL factor by the projected-full factor changes variance by at most 1.285\% of full variance. Their mean tuned NLLs differ by only 0.000127. The separate recomputed spectral split is additive by construction and gives retained/complement NLLs $0.067792/0.111426$, supporting the same bounded interpretation. Replay of the previous full LLA analytic scores agrees to $1.7\times10^{-16}$; double-precision matrix multiplication changes the selected-scale NLL by at most $1.1\times10^{-7}$.

\subsection{Matched-prior orientation control: per-checkpoint results}
\label{app:orientation_checkpoint}
We compare curvature-selected and random transverse subspaces around the same five CIFAR-100/ResNet-18 MAP checkpoints. All Gaussians are centered at the MAP. The question is whether subspace orientation matters when the projected prior is matched and each configuration can select its own scale. There is no longitudinal mixing in this experiment.

We call the curvature-selected configuration K; here K names a configuration rather than a parameter count. Its frame $U_{\rm K}$ comes from the audited reconstruction of the canonical rank-30 MAP frame. At each checkpoint, we also generate five RB frames, where RB means \emph{blockwise-random, prior-matched subspace}. Each frame has 30 orthonormal columns in parameter space. The RB construction randomizes the head and backbone blocks while preserving the head Gram matrix $G_h=U_{{\rm K},h}^{\top}U_{{\rm K},h}$. Since the columns are orthonormal, the backbone Gram matrix is then $I_{30}-G_h$.

The parameter-space prior precision $\Lambda$ is isotropic within each block, with precision $\lambda_h$ in the classifier head and $\lambda_b$ in the backbone. For either K or RB,
\begin{equation}
\label{eq:orientation_projected_prior}
U^\top\Lambda U=\lambda_b I_{30}+(\lambda_h-\lambda_b)G_h.
\end{equation}
Matching $G_h$ therefore matches the full projected prior precision. We keep the inherited prior values: the seed-3 precision was previously reconstructed with the historical fitting procedure, and seed 4 uses its logged four-decimal value. No prior is refitted for this comparison.

For each frame $U$, we use the full projected precision and sample weights as
\begin{align}
P_U & =\operatorname{PSD}\!\left(\operatorname{sym}(U^\top H_0U)\right)+U^\top\Lambda U,\label{eq:orientation_precision}\\
\theta & =\theta_{\rm MAP}+\beta_\perp U P_U^{-1/2}\xi,
\quad \xi\sim\mathcal N(0,I_{30}).\label{eq:orientation_sampling}
\end{align}
Here $\operatorname{sym}(A)=(A+A^\top)/2$, and $\operatorname{PSD}$ replaces negative eigenvalues by zero. The scale $\beta_\perp$ multiplies the standard deviation. The operator $H_0$ is the canonical Hessian of mean cross-entropy over five fixed clean batches of 128, with BatchNorm in training mode. It is not a full-training-set GGN or a conventional full-data Laplace fit.

The K scores should not be read as a spine ablation of the main TRL result in Table~\ref{tab:cifar100_clean_5seed}. Besides removing longitudinal mixing, this experiment uses reconstructed MAP frames, full projected precision, independent scale selection and separate sampling streams. A difference between those scores cannot be attributed solely to the spine.

K and each RB basis select their scale independently by the lowest validation NLL on the grid $\{0,1,2,3,4,6,10,20,40,60,100,150,200,300\}$. Each candidate uses $S_{\rm sel}=8$ Monte Carlo samples; no test labels enter this choice. Final evaluation uses five independent replicas of $S=25$. Within a replica, configurations share the Gaussian draws and the same 25 training mini-batches for rolling BatchNorm recalibration per posterior draw.

We score each predictive distribution before averaging its metrics. K averages five replicas per checkpoint; RB averages five replicas for each of the five bases at that checkpoint. We neither pool their probabilities into a larger ensemble nor select a best random basis. Table~\ref{tab:orientation_checkpoint} reports these checkpoint means and the selected scales. The five checkpoints are the statistical units; bases and replicas are not additional checkpoints.

\begin{table}[H]
\centering
\caption{Matched-prior orientation diagnostic. Panel (a) reports each configuration's validation-selected scale and mean test scores; lower NLL and ECE15 are better. Every RB basis in a checkpoint selected the displayed RB scale. Panel (b) gives the paired differences, K minus RB, calculated before rounding.}
\label{tab:orientation_checkpoint}
\small
\renewcommand{\arraystretch}{1.10}
\setlength{\tabcolsep}{10pt}
\textbf{(a) Selected scale and test scores}\par\smallskip
\begin{tabular}{llrrr}
\toprule
Checkpoint & Method & Selected $\beta_\perp$ & NLL $\downarrow$ & ECE15 $\downarrow$ \\
\midrule
0 & K & 4 & 0.960693 & 0.015600 \\
0 & RB & 60 & 0.958802 & 0.046337 \\
\addlinespace[2pt]
1 & K & 4 & 0.958837 & 0.014812 \\
1 & RB & 100 & 0.958708 & 0.052905 \\
\addlinespace[2pt]
2 & K & 4 & 0.956926 & 0.010967 \\
2 & RB & 60 & 0.956446 & 0.043966 \\
\addlinespace[2pt]
3 & K & 3 & 0.955072 & 0.036553 \\
3 & RB & 60 & 0.949253 & 0.040491 \\
\addlinespace[2pt]
4 & K & 3 & 0.959452 & 0.036084 \\
4 & RB & 60 & 0.941530 & 0.040776 \\
\bottomrule
\end{tabular}

\medskip
\textbf{(b) Paired differences: K minus RB}\par\smallskip
\begin{tabular}{lrr}
\toprule
Checkpoint & $\Delta$NLL (K$-$RB) & $\Delta$ECE15 (K$-$RB) \\
\midrule
0 & $+0.001891$ & $-0.030737$ \\
1 & $+0.000129$ & $-0.038093$ \\
2 & $+0.000480$ & $-0.032999$ \\
3 & $+0.005819$ & $-0.003938$ \\
4 & $+0.017923$ & $-0.004692$ \\
\bottomrule
\end{tabular}
\end{table}

K has lower ECE in all five checkpoints, but the size of the difference varies. The reductions are about $0.031$, $0.038$ and $0.033$ in checkpoints 0--2, compared with $0.004$ and $0.005$ in checkpoints 3--4. K selected $\beta_\perp=4$ in the first three and 3 in the last two. This association is descriptive: the grid is coarse near 3--6, and scale selection minimizes NLL rather than ECE. RB has numerically lower NLL in all five checkpoints.

The mean paired difference, K minus RB, is $-0.022092$ for ECE15, with a 95\% t interval of $[-0.042514,-0.001670]$. For NLL it is $+0.005249$, with interval $[-0.003984,+0.014482]$. Both intervals use the five checkpoint contrasts and four degrees of freedom. The supplement retains the separate Monte Carlo and between-basis variation.

The ECE advantage concerns transverse orientation, not a gain from mixing spine anchors. The reporting test set had already been examined, so this comparison remains exploratory. The fine-tuning results in Appendix~\ref{app:finetune_spine_diagnostic} address longitudinal mixing separately: they show additional functional diversity and modest calibration changes, with NLL and Brier close to the endpoint-centered Gaussian.

We also checked five unconstrained random frames R per checkpoint; unlike RB, these do not match K's projected prior. The numerical checks thus cover 55 frames: five K, 25 R and 25 RB. An initial random-basis HVP-linearity check failed. Before any predictive selection, we amended the check to compare projected Hessians in float32 and float64, using a 2\% threshold chosen after inspecting two checkpoint-0 frames. All 55 passed, with discrepancies of 0.18\%--0.72\%. This checks the implemented five-batch operator, not its agreement with full-data curvature. The original protocol, amendments, scale selections and metric records are included in the supplement.